\ifdefined\pdfminorversion\pdfminorversion=7\fi
\RequirePackage{fix-cm}
\PassOptionsToPackage{table,dvipsnames}{xcolor}
\documentclass{yresearch}
\usepackage{amsmath,amssymb,amsthm}
\usepackage{algorithm,algpseudocode}
\usepackage{array,tabularx,adjustbox,longtable,makecell,arydshln}
\usepackage{enumitem,wrapfig,nicefrac,needspace}
\usepackage[normalem]{ulem}
\usepackage{pifont}
\usetikzlibrary{calc}
\graphicspath{{figures/}}

\newcommand{\titlefont}{\fontsize{19}{20}\selectfont}
\newcommand{\myparagraph}[1]{\vspace{1pt}\noindent\textbf{#1}~~}
\AtBeginDocument{\urlstyle{sf}}

\newlength{\BenchmarkNameWidth}
\newlength{\BenchmarkMetricWidth}
\newenvironment{BenchmarkTable}{%
  \small
  \renewcommand{\arraystretch}{1.0}%
  \setlength{\tabcolsep}{3.2pt}%
  \setlength{\BenchmarkNameWidth}{0.30\linewidth}%
  \setlength{\BenchmarkMetricWidth}{%
    \dimexpr(\linewidth-\BenchmarkNameWidth-13\tabcolsep-2\arrayrulewidth)/6\relax}%
  \begin{tabular}{@{}W{l}{\BenchmarkNameWidth}|*{3}{W{c}{\BenchmarkMetricWidth}}|*{3}{W{c}{\BenchmarkMetricWidth}}}%
}{\end{tabular}}
\definecolor{AlgorithmGreen}{RGB}{0,128,0}
\colorlet{NVCodeRowTint}{MidnightBlue!5}
\algrenewcommand\algorithmicrequire{\textbf{Input:}}
\algrenewcommand\algorithmicensure{\textbf{Output:}}
\newcommand{\ResultTableSetup}{%
  \def\arraystretch{1.0}%
}
\newcommand{\ResultStd}[1]{{\scriptsize{$\pm$#1}}}
\newcommand{\ResultMethod}[1]{\scalebox{1.1}{\textbf{#1}}}
\newcommand{\TightUnderline}[1]{%
  \begingroup
  \setbox0=\hbox{#1}%
  \leavevmode\rlap{\raisebox{-0.18ex}{\rule{\wd0}{0.35pt}}}\box0%
  \endgroup
}
\newcommand{\MeanStd}[2]{\mbox{#1\ResultStd{#2}}}
\newcommand{\BestMeanStd}[2]{\mbox{\bfseries #1\ResultStd{#2}}}
\newcommand{\SecondMeanStd}[2]{\TightUnderline{\mbox{#1\ResultStd{#2}}}}
\newcommand{\MeanOnly}[2]{\mbox{#1}}
\newcommand{\BestMeanOnly}[2]{\mbox{\bfseries #1}}
\newcommand{\SecondMeanOnly}[2]{\TightUnderline{\mbox{#1}}}
\DeclareRobustCommand{\BestCaptionMark}{\textbf{bold}}
\DeclareRobustCommand{\SecondCaptionMark}{\underline{underlined}}
\newcommand{\RankingCaption}{Best values are \BestCaptionMark; second-best values are \SecondCaptionMark.}
\newcommand{\ResultTableHeader}[1]{%
  \toprule
  \multicolumn{1}{c}{} &
  \multicolumn{1}{c}{\multirow{2}{*}{\scalebox{1.1}{\textbf{#1}}}} &
  \multicolumn{2}{c}{\scalebox{1.1}{\textbf{Primary Sensory}}} &
  \multicolumn{2}{c}{\scalebox{1.1}{\textbf{High level Cognitive}}} &
  \multicolumn{2}{c}{\scalebox{1.1}{\textbf{Subcortical}}} &
  \multicolumn{1}{c}{\multirow{2}{*}{\scalebox{1.1}{\textbf{Global Signal}}}} &
  \multicolumn{1}{c}{\multirow{2}{*}{\scalebox{1.1}{\textbf{Avg. R$\uparrow$}}}} &
  \multicolumn{1}{c}{\multirow{2}{*}{\scalebox{1.1}{\textbf{CRPS$\downarrow$}}}} &
  \multicolumn{1}{c}{\multirow{2}{*}{\scalebox{1.1}{\textbf{FC-MAE$\downarrow$}}}} \\
  \cmidrule(lr){3-4}\cmidrule(lr){5-6}\cmidrule(lr){7-8}
  \multicolumn{1}{c}{} &
  \multicolumn{1}{c}{} &
  \scalebox{1.1}{\textbf{Cuneus}} & \scalebox{1.1}{\textbf{Heschl's Gyrus}} &
  \scalebox{1.1}{\textbf{Middle Frontal}} & \scalebox{1.1}{\textbf{Precuneus}} &
  \scalebox{1.1}{\textbf{Putamen}} & \scalebox{1.1}{\textbf{Thalamus}} &
  \multicolumn{1}{c}{} &
  \multicolumn{1}{c}{} &
  \multicolumn{1}{c}{} &
  \multicolumn{1}{c}{} \\
  \midrule
}
\newcommand{\ResultTableSubHeader}[1]{%
  \specialrule{\lightrulewidth}{0pt}{\dimexpr\aboverulesep+\belowrulesep\relax}
  \multicolumn{1}{c}{} &
  \multicolumn{1}{c}{\multirow{2}{*}{\scalebox{1.1}{\textbf{#1}}}} &
  \multicolumn{2}{c}{\scalebox{1.1}{\textbf{Primary Sensory}}} &
  \multicolumn{2}{c}{\scalebox{1.1}{\textbf{High level Cognitive}}} &
  \multicolumn{2}{c}{\scalebox{1.1}{\textbf{Subcortical}}} &
  \multicolumn{1}{c}{\multirow{2}{*}{\scalebox{1.1}{\textbf{Global Signal}}}} &
  \multicolumn{1}{c}{\multirow{2}{*}{\scalebox{1.1}{\textbf{Avg. R$\uparrow$}}}} &
  \multicolumn{1}{c}{\multirow{2}{*}{\scalebox{1.1}{\textbf{CRPS$\downarrow$}}}} &
  \multicolumn{1}{c}{\multirow{2}{*}{\scalebox{1.1}{\textbf{FC-MAE$\downarrow$}}}} \\
  \cmidrule(lr){3-4}\cmidrule(lr){5-6}\cmidrule(lr){7-8}
  \multicolumn{1}{c}{} &
  \multicolumn{1}{c}{} &
  \scalebox{1.1}{\textbf{Cuneus}} & \scalebox{1.1}{\textbf{Heschl's Gyrus}} &
  \scalebox{1.1}{\textbf{Middle Frontal}} & \scalebox{1.1}{\textbf{Precuneus}} &
  \scalebox{1.1}{\textbf{Putamen}} & \scalebox{1.1}{\textbf{Thalamus}} &
  \multicolumn{1}{c}{} &
  \multicolumn{1}{c}{} &
  \multicolumn{1}{c}{} &
  \multicolumn{1}{c}{} \\
  \midrule
}
\newcommand{\IntraBlock}[1]{\multirow{#1}{*}{\rotatebox[origin=c]{90}{\scalebox{1.1}{\textbf{Intra subject}}}}}
\newcommand{\InterBlock}[1]{\multirow{#1}{*}{\rotatebox[origin=c]{90}{\scalebox{1.1}{\textbf{Inter subject}}}}}
\newcommand{\DatasetBlock}[2]{\multirow{#1}{*}{\rotatebox[origin=c]{90}{\footnotesize #2}}}
\newcommand{\SleepEEGfMRI}{Sleep EEG--fMRI}
\newcommand{\MultiCLARID}{MULTI-CLARID}
\newcommand{\GradCPT}{gradCPT}
\newcommand{\XPone}{XP1}
\newcommand{\AffectiveMusic}{Affective Music}
\newcommand{\ExperienceSampling}{Experience Sampling}
\newcommand{\Best}[1]{\textbf{\strut #1}}
\newcommand{\Second}[1]{\TightUnderline{\strut #1}}
\newcommand{\BestRightEdge}[1]{\Best{#1}}
\newcommand{\SecondRightEdge}[1]{\Second{#1}}
\definecolor{BestPerfPurple}{RGB}{221,204,246}
\definecolor{SecondPerfPurple}{RGB}{241,233,251}
\DeclareRobustCommand{\CVPRBestCaptionMark}{%
  \begingroup
  \setlength{\fboxsep}{1pt}%
  \colorbox{BestPerfPurple}{\strut\textbf{darker purple, bold}}%
  \endgroup
}
\DeclareRobustCommand{\CVPRSecondCaptionMark}{%
  \begingroup
  \setlength{\fboxsep}{1pt}%
  \colorbox{SecondPerfPurple}{\strut\underline{light purple, underlined}}%
  \endgroup
}
\newcommand{\UseCVPRTablePalette}{%
  \renewcommand{\Best}[1]{\cellcolor{BestPerfPurple}{\bfseries\strut ##1}}%
  \renewcommand{\Second}[1]{\cellcolor{SecondPerfPurple}{\TightUnderline{\strut ##1}}}%
  \renewcommand{\BestRightEdge}[1]{\multicolumn{1}{>{\columncolor{BestPerfPurple}[\tabcolsep][0pt]}c@{}}{\bfseries\strut ##1}}%
  \renewcommand{\SecondRightEdge}[1]{\multicolumn{1}{>{\columncolor{SecondPerfPurple}[\tabcolsep][0pt]}c@{}}{\TightUnderline{\strut ##1}}}%
  \renewcommand{\BestMeanStd}[2]{\cellcolor{BestPerfPurple}{\mbox{\bfseries ##1\ResultStd{##2}}}}%
  \renewcommand{\SecondMeanStd}[2]{\cellcolor{SecondPerfPurple}{\TightUnderline{\mbox{##1\ResultStd{##2}}}}}%
  \renewcommand{\BestMeanOnly}[2]{\cellcolor{BestPerfPurple}{\mbox{\bfseries ##1}}}%
  \renewcommand{\SecondMeanOnly}[2]{\cellcolor{SecondPerfPurple}{\TightUnderline{\mbox{##1}}}}%
  \let\BestCaptionMark\CVPRBestCaptionMark
  \let\SecondCaptionMark\CVPRSecondCaptionMark
}
\newcommand{\ResultBlockRule}{\specialrule{0.6pt}{0pt}{6pt}}

\newcommand{\ResultBottomRule}{\specialrule{\heavyrulewidth}{0pt}{0pt}}

\newcommand{\DownstreamTableBegin}[1]{%
  \begin{tabular}{@{} c c c c c c c c c c @{}}%
  \toprule
  \multicolumn{1}{c}{\multirow{2}{*}{\textbf{#1}}} & \multicolumn{8}{c}{\textbf{Dataset accuracy / $\Delta$}} & \multirow{2}{*}{\textbf{Mean}} \\
  & \SleepEEGfMRI & \MultiCLARID & \GradCPT & NeuroBOLT & \XPone & \AffectiveMusic & NatView & \ExperienceSampling & \\
  \midrule
}
\newcommand{\DownstreamTableEnd}{\end{tabular}}

\newenvironment{MainAlgorithm}[1][!htbp]{%
  \begin{algorithm}[#1]\small
}{\end{algorithm}}

\theoremstyle{definition}

\theoremstyle{plain}
\newtheorem{proposition}{Proposition}
\newtheorem{theorem}{Theorem}
\newtheorem{lemma}{Lemma}
\newenvironment{shadedeq}{\begin{equation*}}{\end{equation*}}
\newenvironment{shadedeqstar}{\begin{equation*}}{\end{equation*}}
\newcommand{\MainPaperEqref}[1]{\eqref{#1}}

\definecolor{marineblue}{RGB}{3,78,162}


\researchlogo{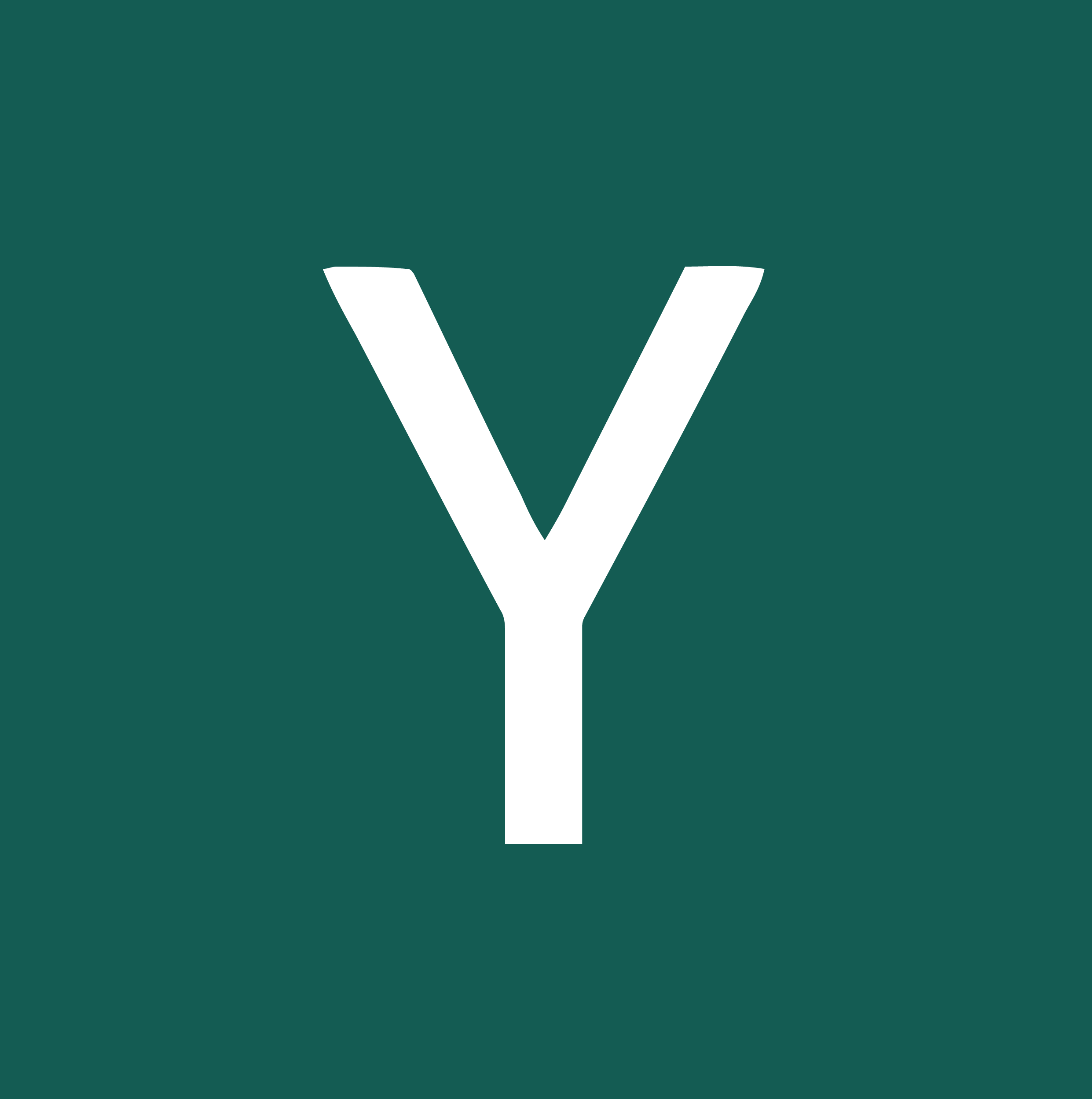}
\title{Learning Neurovascular Codes for EEG-to-fMRI Synthesis}
\DeclareRobustCommand{\NVCodeAuthorProfile}[2]{%
  \href{https://openreview.net/profile?id=~#1}{\textcolor{black}{#2}}%
}
\author[1,2]{\NVCodeAuthorProfile{Yiming_Sun10}{Yiming Sun$^{\circ}$}}
\author[1]{\NVCodeAuthorProfile{Yifan_Wang23}{Yifan Wang$^{\circ}$}}
\author[3]{\NVCodeAuthorProfile{Aosong_Feng1}{Aosong Feng}}
\author[4]{\NVCodeAuthorProfile{Yijia_Ma5}{Yijia Ma}}
\author[2]{\NVCodeAuthorProfile{Ben_Cheng2}{Ben Cheng}}
\author[1]{\NVCodeAuthorProfile{Xurui_Wang2}{Xurui Wang}}
\author[5]{\NVCodeAuthorProfile{Yufan_He2}{Yufan He}}
\author[5]{\NVCodeAuthorProfile{Yucheng_Tang1}{Yucheng Tang}}
\author[3]{\NVCodeAuthorProfile{Qing_Lyu6}{Qing Lyu}}
\author[5]{\NVCodeAuthorProfile{Daguang_Xu2}{Daguang Xu}}
\author[1]{\NVCodeAuthorProfile{Pei-Fen_Kuan1}{Pei-Fen Kuan}}
\author[6]{\NVCodeAuthorProfile{Mingzhou_Ding1}{Mingzhou Ding}}
\author[4]{\NVCodeAuthorProfile{Wen_Li25}{Wen Li}}
\author[1]{\NVCodeAuthorProfile{Chenyu_You1}{Chenyu You}}

\affiliation[1]{Stony Brook University}
\affiliation[2]{University of Toronto}
\affiliation[3]{Yale University}
\affiliation[4]{UTHealth Houston}
\affiliation[5]{NVIDIA}
\affiliation[6]{University of Florida}
\contribution{$^\circ$Equal contribution}

\metadata[GithHub]{\sffamily \url{https://github.com/Y-Research-SBU/NVCode}}
\correspondence{\sffamily Chenyu You: \href{mailto:chenyu.you@stonybrook.edu}{\nolinkurl{chenyu.you@stonybrook.edu}}}
\date{\sffamily\today}
\hypersetup{
  pdftitle={Learning Neurovascular Codes for EEG-to-fMRI Synthesis},
  pdfauthor={Yiming Sun; Yifan Wang; Aosong Feng; Yijia Ma; Ben Cheng; Xurui Wang; Yufan He; Yucheng Tang; Qing Lyu; Daguang Xu; Pei-Fen Kuan; Mingzhou Ding; Wen Li; Chenyu You}
}
\abstract{
In this paper, we introduce \textbf{Neurovascular Codes} as shared EEG representations for fMRI generation. We first train a separate EEG encoder to predict each fMRI signal, learning target-specific temporal and spectral features from the same EEG window. After training, we freeze the encoders and project their concatenated hidden features into a shared {neurovascular code}. Conditioned on this code, \textit{stochastic interpolant} transforms Gaussian noise into a vector of fMRI signals, allowing different responses to be sampled for the same EEG window. We evaluate our approach on eight public simultaneous EEG-fMRI datasets, including four resting-state and four task-based datasets with 61,646 volume-level pairs. Under within-subject and held-out-subject evaluation, our model improves average Pearson correlation over the strongest encoder baselines by 19.1\% and 19.8\%, respectively. The corresponding improvements over the strongest generator baselines are 18.8\% and 47.9\%. Beyond reconstruction accuracy, we also evaluate the generated signals using subject-identification classifiers trained on real fMRI. Our method achieves the highest mean accuracy among the evaluated generators. We hope that our work opens up new opportunities for cross-modal brain modeling.

}

\begin{document}
\setcounter{topnumber}{1}
\maketitle
\section{Introduction}
\label{sec:intro}

% Align each illustration and its caption with the adjacent text block.
\newsavebox{\NVIntroTextBox}
\newlength{\NVIntroTextHeight}
\begin{lrbox}{\NVIntroTextBox}
\begin{minipage}[t]{0.58\textwidth}
\vspace{0pt}
\setlength{\parskip}{6pt}
Learning representations that transfer across tasks is a central aim of foundation models~\citep{bommasani2021opportunities}. This approach has been extended to brain signals, including electroencephalography (EEG) and functional magnetic resonance imaging (fMRI). EEG models~\citep{kostas2021bendr,jiang2024large,wang2024eegpt,ouahidi2025reve} learn across participants and recording settings. BrainLM~\citep{caro2023brainlm} learns spatiotemporal fMRI patterns for clinical prediction. Unlike studies that train on one modality alone, we use fMRI prediction to supervise EEG representations for generation.

EEG captures rapid electrical fluctuations in the brain and supports brain-computer interfaces~\citep{mcfarland2011brain,lotte2018review}. fMRI measures blood-oxygen-level-dependent (BOLD) signals~\citep{ogawa1990brain,logothetis2008we} and maps brain regions and networks~\citep{smith2013resting,glasser2016multi}. Simultaneous EEG-fMRI recordings~\citep{huster2012methods,murta2015electrophysiological,bullock2021artifact} pair these complementary measurements.
\end{minipage}
\end{lrbox}
\setlength{\NVIntroTextHeight}{\dimexpr\ht\NVIntroTextBox+\dp\NVIntroTextBox\relax}
\noindent\usebox{\NVIntroTextBox}\hfill
\begin{minipage}[t][\NVIntroTextHeight][s]{0.38\textwidth}
\vspace{0pt}
\centering
\captionsetup{type=figure,position=bottom,skip=3pt}
\includegraphics[width=.945\linewidth]{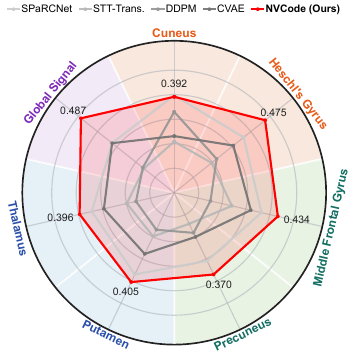}\par
\vfill
\caption{\textbf{EEG-to-fMRI synthesis on held-out subjects.} Target-level Pearson correlation for NVCode and representative baselines, averaged equally across eight datasets.}
\label{fig:intro-regional-correlation}
\end{minipage}\par

Paired EEG-fMRI requires specialized hardware~\citep{allen2000method} and artifact suppression~\citep{niazy2005removal,huster2012methods,bullock2021artifact}, limiting scale. We build on EEG prediction of individual fMRI signals~\citep{kovalev2022fmri,li2024leveraging,li2024neurobolt} to learn fMRI synthesis from EEG without a concurrent scan.

In this paper, we introduce \textsc{Neurovascular Codes}, termed \textbf{NVCode}, for EEG-to-fMRI synthesis. We first train a separate EEG encoder to predict each fMRI signal from the same EEG window. Through these prediction tasks, the encoders learn target-specific spatiotemporal~\citep{vaswani2017attention,dosovitskiy2020image} and spectral~\citep{schirrmeister2017deep,banville2021uncovering} features. After training, we discard the prediction heads and freeze the encoders, retaining their hidden features for generation. A learned projection maps the concatenated features into a shared neurovascular code that conditions a stochastic fMRI generator.

\begin{lrbox}{\NVIntroTextBox}
\begin{minipage}[t]{0.552\textwidth}
\vspace{0pt}
An EEG window may correspond to more than one plausible fMRI response (Figure~\ref{fig:intro-one-to-many}). Scalp EEG has spatial ambiguity~\citep{huster2012methods}, and neurovascular coupling varies across regions and individuals~\citep{logothetis2008we}. We therefore condition a \textit{stochastic interpolant} on the neurovascular code, using conditional velocity matching~\citep{lipman2022flow,tong2023improving}. The model learns a velocity field that transforms Gaussian noise into a vector of fMRI signals. Each velocity prediction depends on the current values of all target signals, allowing the field to model dependencies among them. Different noise samples yield different response vectors for the same EEG window. Averaging these samples gives a predictive mean for each fMRI target. The individual draws retain the joint variation across targets, which is not visible in the mean alone.
\end{minipage}
\end{lrbox}
\setlength{\NVIntroTextHeight}{\dimexpr\ht\NVIntroTextBox+\dp\NVIntroTextBox\relax}
\noindent\usebox{\NVIntroTextBox}\hfill
\begin{minipage}[t][\NVIntroTextHeight][s]{0.4085\textwidth}
\vspace{0pt}
\centering
\captionsetup{type=figure,position=bottom,skip=3pt}
\includegraphics[width=\linewidth]{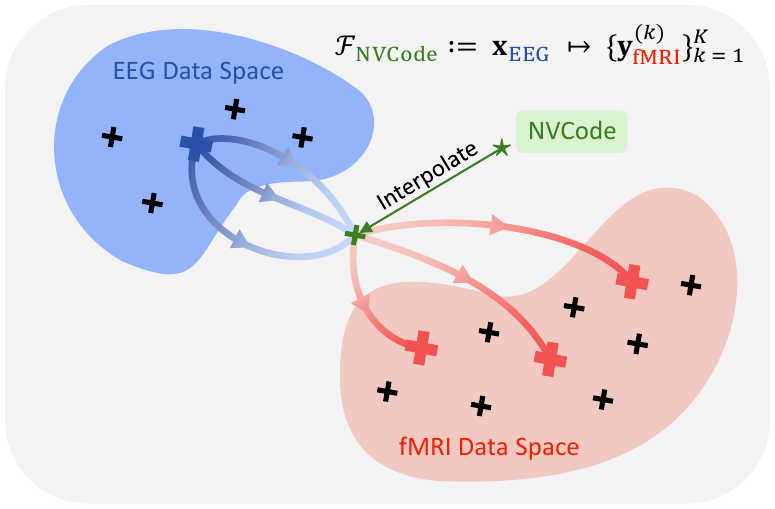}\par
\vfill
\caption{\textbf{One-to-many EEG conditioning.} For a fixed EEG window, different draws from the Gaussian source can produce distinct joint fMRI responses under the same code.}
\label{fig:intro-one-to-many}
\end{minipage}\par

We evaluate our approach on eight public simultaneous EEG-fMRI datasets, including four resting-state and four task-based datasets with $61{,}646$ volume-level pairs. The benchmark uses common target definitions, data splits, and evaluation metrics. We compare EEG encoders using a common generator architecture and compare generation networks using identical cached EEG codes. We also ablate which target-specific representations are supplied to the generator. Under within-subject and held-out-subject evaluation, our method achieves relative gains in average Pearson correlation of $19.1\%$ and $19.8\%$ over the strongest encoder baselines, respectively. The corresponding gains over the strongest generator baselines are $18.8\%$ and $47.9\%$.

Beyond prediction accuracy, we evaluate marginal predictive distributions and functional connectivity computed from predictive-mean time series. Our method achieves the lowest average functional connectivity error in both comparisons and evaluation settings. With classifiers trained on real fMRI and held fixed, our generated signals also achieve the highest mean subject-identification accuracy, exceeding accuracy on real fMRI test signals by $3.1$ percentage points on average. Our contributions are as follows:

\begin{enumerate}[leftmargin=*,label=\textbf{\arabic*.},labelsep=0.55em,topsep=2pt,itemsep=2pt,parsep=0pt]
    \item We introduce \textsc{Neurovascular Codes}, shared EEG representations learned through target-specific fMRI prediction and used to condition stochastic fMRI generation.
    \item We establish a benchmark across eight public simultaneous EEG-fMRI datasets, covering four resting-state and four task-based datasets. It uses common target definitions, data splits, and evaluation metrics for comparisons in both within-subject and held-out-subject settings.
    \item We compare EEG encoders, conditional generators, and target-specific representation sets under controlled experimental settings. Beyond reconstruction accuracy, we evaluate the marginal predictive distributions and functional connectivity computed from predictive-mean time series. We assess subject identification performance using classifiers trained on real fMRI signals.
\end{enumerate}

\section{Related Work}
\label{sec:related-work}

\myparagraph{EEG and fMRI representation learning.}
Representation learning is widely used in brain--computer interfaces~\citep{mcfarland2011brain,lotte2018review}. Large-scale pretraining provides one way to learn features that transfer across tasks~\citep{bommasani2021opportunities}. For EEG, BENDR~\citep{kostas2021bendr} uses masked contrastive learning, while BIOT~\citep{yang2023biot} enables pretraining across heterogeneous biosignal recordings. LaBraM~\citep{jiang2024large} predicts masked neural tokens, EEGPT~\citep{wang2024eegpt} pairs masked reconstruction with representation alignment, and REVE~\citep{ouahidi2025reve} uses masked autoencoding with electrode-coordinate embeddings. This line of work addresses variation in electrode layouts~\citep{yang2023biot,ouahidi2025reve} and supports a range of EEG decoding tasks~\citep{lotte2018review,wang2024eegpt}. For fMRI, BrainLM~\citep{caro2023brainlm} learns spatial and temporal structure through masked prediction and supports clinical prediction~\citep{caro2023brainlm}. During pretraining, these methods construct learning targets from the input signals themselves. Our work instead uses fMRI prediction to supervise EEG representation learning.

\myparagraph{EEG-to-fMRI modeling.}
EEG--fMRI studies relate electrophysiological activity to the BOLD response~\citep{ogawa1990brain,logothetis2008we} to investigate neurovascular coupling~\citep{logothetis2008we,murta2015electrophysiological} and constrain EEG source localization~\citep{huster2012methods}. fMRI also supports the study of functional networks~\citep{biswal1995functional,van2010exploring} through statistical dependencies across brain regions~\citep{biswal1995functional,friston2011functional}.
BEIRA~\citep{kovalev2022fmri} learns interpretable mappings from EEG to BOLD signals, while SIREN-based prediction~\citep{li2024leveraging} uses sinusoidal networks to extract frequency information. NeuroBOLT~\citep{li2024neurobolt} learns spatiotemporal and multiscale spectral features and evaluates prediction on both resting-state and task recordings. UnEBOLT~\citep{li2026unebolt} predicts BOLD time series across brain parcels with a single adaptive decoder and estimates functional connectivity from those predictions. NVCode uses target-specific prediction to learn EEG features that condition stochastic generation of the fMRI signal vector.

\myparagraph{Conditional generative models.}
EEG-to-fMRI ambiguity and variable neurovascular coupling motivate probabilistic modeling~\citep{logothetis2008we,murta2015electrophysiological}. Models use stochastic latents~\citep{kingma2013auto,sohn2015learning}, adversarial training~\citep{mirza2014conditional,isola2017image}, diffusion denoising~\citep{ho2020denoising,song2020score,karras2022elucidating}, or cross-time consistency for few-step sampling~\citep{song2023consistency,yang2024consistency}. Neural ODEs and FFJORD formulate continuous normalizing flows~\citep{chen2018neural,grathwohl2018ffjord}, whereas flow matching, rectified flow, and optimal-transport conditional flow matching learn path velocities~\citep{lipman2022flow,liu2022flow,tong2023improving}. Stochastic interpolants connect distributions through deterministic or stochastic dynamics~\citep{albergo2025stochastic}. NVCode conditions this framework on neurovascular codes and the full fMRI state, allowing interactions among targets.

\section{Neurovascular Codes}
\label{sec:methodology}
NVCode learns EEG representations through individual fMRI prediction tasks and uses these representations to condition fMRI generation. We first train a separate EEG encoder for each fMRI target (Section~\ref{sec:method:representations}). With the encoders frozen, we then train a projector and a conditional velocity field together (Sections~\ref{sec:method:conditioning} and~\ref{sec:method:interpolant-learning}). The projector maps the encoder features to a neurovascular code, which guides generation from Gaussian noise to a vector of fMRI signals. Appendix~\ref{app:method} provides architectural specifications, derivations, and training and sampling details.

\subsection{Preliminaries}
\label{sec:method:prelim}
Each observation pairs the $16$\,s EEG window $\mathbf{x}\in\mathbb{R}^{C\times T}$ immediately preceding an fMRI volume with its BOLD target vector $\mathbf{y}\in\mathbb{R}^{N}$. Here, $C$ and $T$ count EEG channels and time samples. The vector contains six DiFuMo region-of-interest (ROI) signals~\citep{dadi2020fine} and one global signal ($N=7$).

Projecting the target-specific EEG encoders' hidden features yields the neurovascular code $\mathbf{c}_{\psi}(\mathbf{x})\in\mathbb{R}^{d_c}$ (Section~\ref{sec:method:conditioning}). It conditions $p_{\theta}(\mathbf{y}\mid\mathbf{c}_{\psi}(\mathbf{x}))$, with projector parameters $\psi$ and generator parameters $\theta$. Each generated vector contains seven signals at one acquisition time point. Applying the model to successive EEG windows yields time series for evaluation. Let $\boldsymbol{\mu},\boldsymbol{\sigma}\in\mathbb{R}^{N}$ denote the per-target mean and standard deviation estimated only from the training split. The standardized generation endpoint is
\begin{equation}
\mathbf{z}_1
= (\mathbf{y}-\boldsymbol{\mu})\oslash\boldsymbol{\sigma},
\label{eq:nvsi-normalization}
\end{equation}
where $\oslash$ denotes elementwise division. Standardizing each coordinate separately puts the regional and global signals on a common scale for velocity regression. The same training statistics are used in all splits, including testing on held-out subjects. Inverting the transformation puts generated and observed BOLD vectors on the same fMRI target scale for evaluation.

\begin{figure}[t]
\centering
\includegraphics[width=0.931\textwidth]{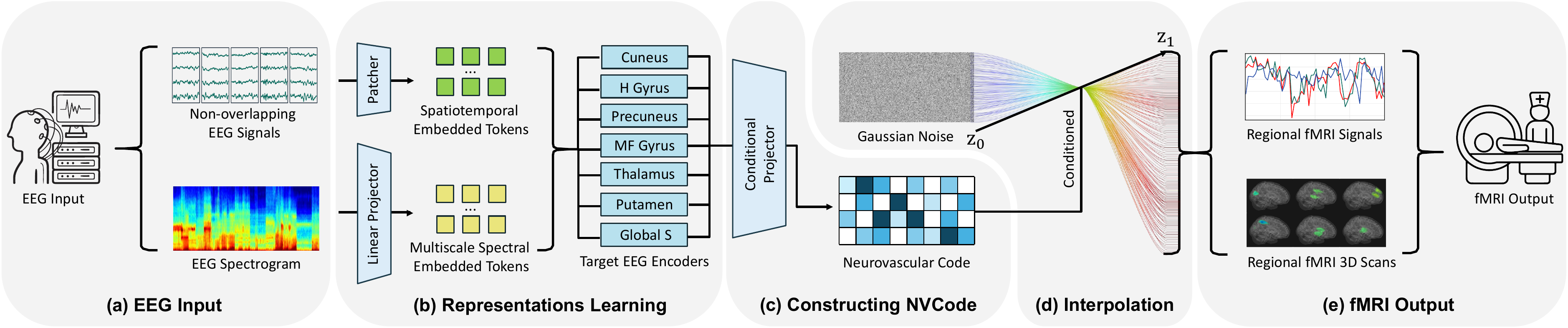}
\caption{\textbf{NVCode pipeline.} \textbf{(a)} Non-overlapping patches and spectrograms describe each EEG window. \textbf{(b)} Target-specific encoders learn spatiotemporal and multiscale spectral features under fMRI supervision. \textbf{(c)} A projector combines the features into a neurovascular code. \textbf{(d)} The code conditions joint generation from Gaussian noise through a learned interpolant. \textbf{(e)} Time series and anatomical projections display the regional and global outputs.}
\label{fig:nvcode-pipeline}
\end{figure}

\subsection{Learning Target-Specific EEG Representations}
\label{sec:method:representations}

Each EEG encoder extracts spatiotemporal and multiscale spectral features to predict one fMRI target from the same window. For target $n$, we attach a linear prediction head to the encoder features and minimize the squared difference between its prediction and the corresponding fMRI target:
\begin{equation}
\begin{gathered}
\mathbf{e}_n=g_{\phi_n}(\mathbf{x})\in\mathbb{R}^{d},\\
\widehat y_n^{\mathrm{aux}}=\mathbf{a}_n^{\top}\mathbf{e}_n+b_n,\\
\mathcal{L}_{\mathrm{rep}}^{(n)}
=\mathbb{E}_{(\mathbf{x},\mathbf{y})\sim\mathcal{D}_{\mathrm{tr}}}
\!\left[(\widehat y_n^{\mathrm{aux}}-y_n)^2\right],
\quad n=1,\ldots,N.
\end{gathered}
\label{eq:enc-loss}
\end{equation}
Here, $\mathcal{D}_{\mathrm{tr}}$ is the training set and $d$ the feature dimension. Encoder--head pairs are fitted independently and selected by validation mean squared error (MSE). We discard the temporary heads and freeze the selected encoders, retaining their hidden features for code construction.

\subsection{Constructing Neurovascular Codes}
\label{sec:method:conditioning}

NVCode concatenates the frozen encoder outputs and applies a learned projector $\Pi_{\psi}$. Its first-layer weight matrix $\mathbf{W}_1=[\mathbf{W}_1^{(1)}\;\cdots\;\mathbf{W}_1^{(N)}]$ is partitioned by encoder output. We write the concatenation, the code, and the first-layer map as:
\begin{equation}
\begin{gathered}
\mathbf{e}(\mathbf{x})=[\mathbf{e}_1;\ldots;\mathbf{e}_N]\in\mathbb{R}^{N\times d},\\
\mathbf{c}_{\psi}(\mathbf{x})=\Pi_{\psi}(\mathbf{e}(\mathbf{x}))\in\mathbb{R}^{d_c},\\
\mathbf{W}_1\mathbf{e}(\mathbf{x})+\mathbf{b}_1
=\sum_{n=1}^{N}\mathbf{W}_1^{(n)}\mathbf{e}_n+\mathbf{b}_1.
\end{gathered}
\label{eq:shared-conditioning}
\end{equation}
Here, $[\cdot;\cdot]$ denotes concatenation, and $\mathbf{W}_1^{(n)}$ linearly transforms target $n$ features. The projector uses two linear layers with layer normalization~\citep{ba2016layer}. With target-supervised encoders frozen, generation jointly trains the projector and velocity field (Section~\ref{sec:method:interpolant-learning}). For each EEG window, the same code conditions all targets at every step of sampling.

\subsection{Learning the Conditional Velocity Field}
\label{sec:method:interpolant-learning}

For each training pair $(\mathbf{x},\mathbf{y})$, we compute the neurovascular code $\mathbf{c}=\mathbf{c}_{\psi}(\mathbf{x})$ and the normalized target $\mathbf{z}_1$ from Eq.~\eqref{eq:nvsi-normalization}. We draw $\mathbf{z}_0\sim\mathcal{N}(\mathbf{0},\mathbf{I}_N)$ and $t\sim\mathcal{U}(0,1)$ independently of the observation and each other. A linear interpolant connects the Gaussian source and normalized fMRI target, with its path velocity given by~\citep{lipman2022flow,tong2023improving,albergo2025stochastic}:
\begin{equation}
\begin{gathered}
\mathbf{z}_t=(1-t)\mathbf{z}_0+t\mathbf{z}_1, \\
\mathbf{u}_t=\frac{d\mathbf{z}_t}{dt}=\mathbf{z}_1-\mathbf{z}_0.
\end{gathered}
\label{eq:interpolation}
\end{equation}
Here, $t$ is interpolation time, not recording time. The state $\mathbf{z}_t$ and its velocity target $\mathbf{u}_t$ are computed directly from the endpoints, without integrating an ODE during training.

We project the full state $\mathbf{z}_t$ through a learned linear map and concatenate it with the code and time embedding $\boldsymbol{\tau}_t=\boldsymbol{\tau}(t)$. A multilayer perceptron (MLP) then maps the concatenated features to one velocity for each fMRI target:
\begin{equation}
\begin{gathered}
\mathbf{r}_t=\mathbf{A}\mathbf{z}_t+\mathbf{b}_A,\\
\mathbf{h}_t=[\mathbf{c};\mathbf{r}_t;\boldsymbol{\tau}_t],\\
v_{\theta}(\mathbf{z}_t,t,\mathbf{c})=\operatorname{MLP}_{\theta}(\mathbf{h}_t),\\
D_{\mathbf{z}_t}v_\theta=\mathbf{J}_{\mathrm{MLP},r}\mathbf{A}.
\end{gathered}
\label{eq:nvsi-field-main}
\end{equation}
The parameters $\theta$ include the state projection $(\mathbf{A},\mathbf{b}_A)$ and the MLP. With the code and time fixed, $\mathbf{J}_{\mathrm{MLP},r}$ is the Jacobian of the MLP output with respect to $\mathbf{r}_t$. The resulting state Jacobian need not be diagonal, allowing the velocity of each target to depend on other entries in $\mathbf{z}_t$.

We train the projector and velocity field by minimizing
\begin{equation}
\mathcal{L}_{\mathrm{NVCode}}(\psi,\theta)
=\mathbb{E}_{(\mathbf{x},\mathbf{y}),\mathbf{z}_0,t}
\!\left[
\left\|v_{\theta}\!\left(\mathbf{z}_t,t,\mathbf{c}_{\psi}(\mathbf{x})\right)
-\mathbf{u}_t\right\|_2^2
\right].
\label{eq:nvsi-loss}
\end{equation}
The expectation averages over training pairs, Gaussian source samples, and interpolation times. The loss sums squared velocity errors across all fMRI targets and jointly updates the projector and velocity field parameters. The EEG encoder parameters remain fixed while the projector and conditional velocity field are fitted.

\begingroup
\setlength{\intextsep}{2pt}
\begin{wrapfigure}[15]{r}{0.51\textwidth}
\begin{minipage}{\linewidth}
\setlength{\intextsep}{0pt}
\captionsetup{hypcap=false}
\begin{MainAlgorithm}[H]
\caption{NVCode fitting}
\label{alg:nvsi-fit-main}
\footnotesize
\begin{algorithmic}[1]
\Require Training data $\mathcal{D}_{\mathrm{tr}}$; validation data $\mathcal{D}_{\mathrm{val}}$; $N$ targets.
\Ensure Frozen encoders $g_{\phi_{1:N}}$; projector $\Pi_\psi$; field $v_\theta$; statistics $(\boldsymbol{\mu},\boldsymbol{\sigma})$.
\State Estimate $(\boldsymbol{\mu},\boldsymbol{\sigma})$ on training targets.
\Statex \textcolor{AlgorithmGreen}{\textbf{Representation fitting}}
\For{$n=1,\ldots,N$}
    \State Fit $g_{\phi_n}$ and its head with Eq.~\eqref{eq:enc-loss}; select by validation MSE.
    \State Freeze $g_{\phi_n}$ and discard its head.
\EndFor
\State Initialize $\Pi_\psi,v_\theta$.
\Statex \textcolor{AlgorithmGreen}{\textbf{Interpolant fitting}}
\Repeat
    \State Draw a training batch; compute $\mathbf{c},\mathbf{z}_1$ and draw $\mathbf{z}_0,t$ independently per example.
    \State Form $(\mathbf{z}_t,\mathbf{u}_t)$; \textcolor{AlgorithmGreen}{update $(\psi,\theta)$ on $\mathcal{L}_{\mathrm{NVCode}}$.}
\Until{the split-specific training schedule ends}
\State Restore the selected projector--field checkpoint.
\end{algorithmic}
\end{MainAlgorithm}
\end{minipage}
\end{wrapfigure}

\myparagraph{Training.}
We fit each EEG encoder and its linear head independently with AdamW~\citep{loshchilov2017decoupled}. The pair minimizes prediction error for its assigned fMRI target:
\begin{equation*}
\begin{gathered}
\min_{\phi_n,\mathbf{a}_n,b_n}\mathcal{L}_{\mathrm{rep}}^{(n)}(\phi_n,\mathbf{a}_n,b_n),\\
n=1,\ldots,N.
\end{gathered}
\end{equation*}
Validation MSE selects each encoder checkpoint $\phi_n^{\star}$. With these encoders fixed, the second stage optimizes
\begin{equation*}
\begin{gathered}
\min_{\psi,\theta}\mathcal{L}_{\mathrm{NVCode}}(\psi,\theta;\phi_1^{\star},\ldots,\phi_N^{\star}),\\
\phi_{1:N}=\phi_{1:N}^{\star}\ \text{fixed}.
\end{gathered}
\end{equation*}
We use AdamW with OneCycle scheduling and resample $(\mathbf{z}_0,t)$ independently for every training example. The projector and velocity field are selected together, giving $(\psi^{\star},\theta^{\star})$. Algorithm~\ref{alg:nvsi-fit-main} summarizes the two stages.
\par\endgroup

\subsection{Generating fMRI Signals}
\label{sec:method:generation}

For an unseen EEG window $\mathbf{x}^{\ast}$, we compute $\mathbf{c}_{\psi^{\star}}(\mathbf{x}^{\ast})$ once from the frozen encoder features. This shared condition remains fixed for all generated samples. We draw $K$ independent Gaussian initial states and integrate each trajectory through the learned velocity field from $t=0$ to $t=1$:
\begin{equation}
\begin{gathered}
\frac{d\mathbf{s}^{(k)}(t)}{dt}
=v_{\theta^{\star}}\!\left(
\mathbf{s}^{(k)}(t),t,\mathbf{c}_{\psi^{\star}}(\mathbf{x}^{\ast})
\right), \\
\mathbf{s}^{(k)}(0)\sim\mathcal{N}(\mathbf{0},\mathbf{I}_N).
\end{gathered}
\label{eq:nvsi-sampling-ode}
\end{equation}
Although the training paths in Eq.~\eqref{eq:interpolation} are linear, the trajectories induced by the learned field need not be. The sampled states evolve in the same standardized coordinates as the training targets. We draw a new initial state for each trajectory while keeping the learned field and its EEG condition fixed. These trajectories generate alternative samples at one fMRI acquisition time, not successive points in the fMRI time series.

\begingroup
\setlength{\intextsep}{2pt}
\begin{wrapfigure}[10]{r}{0.51\textwidth}
\begin{minipage}{\linewidth}
\setlength{\intextsep}{0pt}
\captionsetup{hypcap=false}
\begin{MainAlgorithm}[H]
\caption{NVCode sampling}
\label{alg:nvsi-sample-main}
\footnotesize
\begin{algorithmic}[1]
\Require EEG window $\mathbf{x}^{\ast}$; frozen $g_{\phi_{1:N}},\Pi_\psi,v_\theta$; statistics $(\boldsymbol{\mu},\boldsymbol{\sigma})$; steps $S$; samples $K$.
\Ensure Samples $\{\mathbf{y}^{(k)}\}_{k=1}^{K}$ and predictive mean $\widehat{\mathbf{y}}$.
\State Compute $\mathbf{c}=\Pi_\psi([g_{\phi_1}(\mathbf{x}^{\ast});\cdots;g_{\phi_N}(\mathbf{x}^{\ast})])$ once.
\For{$k=1,\ldots,K$}
    \State Draw $\mathbf{s}^{(k)}_0\sim\mathcal{N}(\mathbf{0},\mathbf{I}_N)$ independently.
    \For{$q=0,\ldots,S-1$}
        \State $\mathbf{s}^{(k)}_{q+1}\gets\mathbf{s}^{(k)}_q+S^{-1}v_\theta(\mathbf{s}^{(k)}_q,q/S,\mathbf{c})$.
    \EndFor
    \State $\mathbf{y}^{(k)}\gets\boldsymbol{\mu}+\boldsymbol{\sigma}\odot\mathbf{s}^{(k)}_S$.
\EndFor
\State \textcolor{AlgorithmGreen}{Compute $\widehat{\mathbf{y}}\gets K^{-1}\sum_{k=1}^{K}\mathbf{y}^{(k)}$.}
\State \textbf{return} $\{\mathbf{y}^{(k)}\}_{k=1}^{K},\widehat{\mathbf{y}}$.
\end{algorithmic}
\end{MainAlgorithm}
\end{minipage}
\end{wrapfigure}

With $S$ forward Euler steps of size $1/S$, the endpoint is $\mathbf{s}^{(k)}_S$. We restore the fMRI target scale and average the $K$ samples to estimate the predictive mean:
\begin{equation}
\begin{gathered}
\mathbf{y}^{(k)}
=\boldsymbol{\mu}+\boldsymbol{\sigma}\odot\mathbf{s}^{(k)}_S, \\
\widehat{\mathbf{y}}
=\frac{1}{K}\sum_{k=1}^{K}\mathbf{y}^{(k)}.
\end{gathered}
\label{eq:nvsi-output}
\end{equation}
Here, $\odot$ denotes elementwise multiplication. Increasing $K$ reduces the Monte Carlo variability of the estimated mean, not the spread of the generated response distribution.

Algorithm~\ref{alg:nvsi-sample-main} returns samples and their predictive mean. Across $M$ aligned EEG windows, these form a $K\times M\times N$ sample tensor and an $M\times N$ predictive-mean time series. Pearson correlation and functional connectivity mean absolute error (FC-MAE) evaluate the time series; the continuous ranked probability score (CRPS) evaluates marginal sample distributions separately for each fMRI target.
\par\endgroup

\begin{table}[t]
\captionsetup{width=.95\linewidth,justification=justified}
\begingroup
\centering
\caption{\textbf{Eight simultaneous EEG--fMRI datasets span resting-state and task-evoked settings.} The source cohorts report 184 participants; after scan-level exclusions, 189 retained scans yield 61,646 paired volumes. Labels follow the source dataset titles, and citations include repository accessions where available.}
\label{tab:datasets}
\label{tab:main-datasets}
\renewcommand{\arraystretch}{0.98}
\setlength{\tabcolsep}{2.8pt}
\resizebox{0.95\linewidth}{!}{%
\begin{tabular}{l|l|*{2}{c}|*{2}{c}|*{3}{c}|l}
\toprule
\multirow{2}{*}{\textbf{Type}} &
\multicolumn{1}{c|}{\multirow{2}{*}{\textbf{Dataset}}} &
\multicolumn{2}{c|}{\textbf{EEG}} &
\multicolumn{2}{c|}{\textbf{fMRI}} &
\multicolumn{3}{c|}{\textbf{Paired Acquisition}} &
\multicolumn{1}{c}{\multirow{2}{*}{\textbf{Paradigm}}} \\
\cmidrule(lr){3-4}\cmidrule(lr){5-6}\cmidrule(lr){7-9}
& & \textbf{Rate (Hz)} & \textbf{Channels} &
\textbf{TR (s)} & \textbf{Pairs} &
\textbf{Duration (s)} & \textbf{Scans} & \textbf{Cohort Size} & \\
\midrule
\DatasetBlock{4}{Rest}
  & \SleepEEGfMRI~\citep{ds003768:1.0.13} & 5000 & 32 & 2.1 & 5,714 & 600 & 20 & 33 & Sleep stage \\
  & \MultiCLARID~\citep{ds005795:1.0.0} & 5000 & 64 & 2.0 & 10,200 & 600 & 34 & 24 & Eyes open rest \\
  & \GradCPT~\citep{ds006040:1.0.2} & 5000 & 64 & 2.0 & 5,040 & 360 & 28 & 28 & Rest subset \\
  & NeuroBOLT~\citep{li2024neurobolt} & 5000 & 26 & 2.1 & 16,571 & 1200 & 29 & 22 & Multi session rest \\
\midrule
\DatasetBlock{4}{Active}
  & \XPone~\citep{ds002336:2.0.2} & 5000 & 64 & 2.0 & 2,400 & 480 & 10 & 10 & Motor imagery feedback \\
  & \AffectiveMusic~\citep{ds002725:1.0.0} & 5000 & 31 & 2.0 & 6,300 & 600 & 21 & 21 & Classical music listening \\
  & \ExperienceSampling~\citep{ds007216:1.0.0} & 5000 & 64 & 2.0 & 14,850 & 660 & 45 & 24 & Sustained attention task \\
  & NatView~\citep{telesford2023open} & 5000 & 61 & 2.1 & 571 & 600 & 2 & 22 & Naturalistic film viewing \\
\bottomrule
\end{tabular}}
\par
\endgroup

\end{table}
\begingroup
\UseCVPRTablePalette
\FloatBarrier
\section{Experiments}
\label{sec:experiments}

% The small results panels share one grid, independent of individual table widths.
\newcommand{\NVResultLeft}[1][12pt]{\par\noindent\begin{minipage}[t]{0.455\linewidth}\vspace{0pt}\setlength{\parskip}{5pt}\setlength{\parindent}{0pt}\setlength{\baselineskip}{#1}}
\newcommand{\NVResultRight}{\par\end{minipage}\hfill\begin{minipage}[t]{0.48925\linewidth}\vspace{0pt}\captionsetup{justification=raggedright,skip=6pt}}
\newcommand{\NVResultTableCaption}{\captionsetup{type=table,width=\linewidth,justification=justified,singlelinecheck=false}}
\newcommand{\NVResultEnd}{\par\vspace{0pt}\end{minipage}\par}
\newcolumntype{U}{>{\scriptsize}W{l}{119pt}|*{8}{W{c}{44pt}}|W{c}{35pt}}
\newcommand{\NVUtilityDatasetHeads}{\scriptsize\shortstack{Sleep\\EEG--fMRI} & \scriptsize\shortstack{MULTI-\\CLARID} & \scriptsize\GradCPT & \scriptsize NeuroBOLT & \scriptsize\XPone & \scriptsize\shortstack{Affective\\Music} & \scriptsize NatView & \scriptsize\shortstack{Experience\\Sampling}}

We evaluate NVCode on eight public simultaneous EEG-fMRI datasets (Table~\ref{tab:main-datasets}) under within-subject and held-out-subject protocols. Section~\ref{sec:exp-main} compares EEG encoders and conditional generators under matched pipelines; Section~\ref{sec:exp-analysis} examines temporal and spectral fidelity in representative scans; Section~\ref{sec:exp-downstream} tests whether generated fMRI retains identity information accessible to a classifier trained only on real data; and Section~\ref{sec:exp-ablation} examines the effects of the main model components. See Appendix~\ref{app:additional} for dataset-level results and scan examples, and Appendix~\ref{app:experiment} for details of the experimental protocols and metrics.

\begin{table}[t]
{
\centering
\captionsetup{width=.931\linewidth,justification=justified}
\caption{\textbf{EEG encoder comparison.} We fit the same generator architecture separately for each frozen encoder. Higher Avg.~$R$ (mean Pearson correlation) and lower marginal CRPS and predictive-mean FC-MAE are better. Best values are \BestCaptionMark; second best values are \SecondCaptionMark.}
\label{tab:main-encoder}
\renewcommand\arraystretch{1.20}
\resizebox{.931\linewidth}{!}{%
\begin{BenchmarkTable}
\toprule
\multirow{2}{*}{\textbf{Encoder}} & \multicolumn{3}{c|}{\textbf{Intra-Subject}} & \multicolumn{3}{c}{\textbf{Inter-Subject}} \\
& \textbf{Avg.~$R\uparrow$} & \textbf{CRPS$\downarrow$} & \textbf{FC-MAE$\downarrow$}
& \textbf{Avg.~$R\uparrow$} & \textbf{CRPS$\downarrow$} & \textbf{FC-MAE$\downarrow$} \\
\midrule
BEIRA~\citep{kovalev2022fmri} & 0.346 & 0.393 & 0.327 & 0.158 & 0.369 & 0.204 \\
BIOT~\citep{yang2023biot} & 0.262 & 0.363 & 0.328 & 0.246 & 0.348 & 0.172 \\
BrainOmni~\citep{xiao2025brainomni} & 0.277 & \Second{0.360} & 0.331 & 0.274 & 0.347 & 0.168 \\
CNN-Trans.~\citep{peh2022transformer} & 0.320 & 0.363 & 0.332 & 0.250 & 0.347 & 0.190 \\
ContraWR~\citep{yang2023self} & 0.329 & 0.390 & 0.348 & 0.238 & 0.358 & 0.221 \\
FFCL~\citep{li2022motor} & 0.315 & 0.389 & 0.351 & 0.264 & 0.363 & 0.197 \\
LaBraM~\citep{jiang2024large} & 0.265 & \Second{0.360} & 0.319 & 0.304 & 0.345 & 0.187 \\
SIREN-based~\citep{li2024leveraging} & \Second{0.403} & 0.383 & \Second{0.315} & 0.301 & 0.354 & 0.189 \\
REVE~\citep{ouahidi2025reve} & 0.220 & 0.365 & 0.340 & 0.174 & 0.351 & 0.200 \\
SPaRCNet~\citep{jing2023development} & 0.330 & 0.379 & 0.330 & \Second{0.353} & \Second{0.344} & \Second{0.163} \\
STT-Trans.~\citep{song2021transformer} & 0.258 & 0.381 & 0.323 & 0.192 & 0.362 & \Second{0.163} \\
\addlinespace[1pt]
\textbf{NVCode (Ours)} & \Best{0.480} & \Best{0.352} & \Best{0.280} & \Best{0.423} & \Best{0.344} & \Best{0.135} \\
\ResultBottomRule
\end{BenchmarkTable}}\par
}

\par
\begin{minipage}{\linewidth}
\centering
\captionsetup{type=figure}
\includegraphics[width=.95\linewidth]{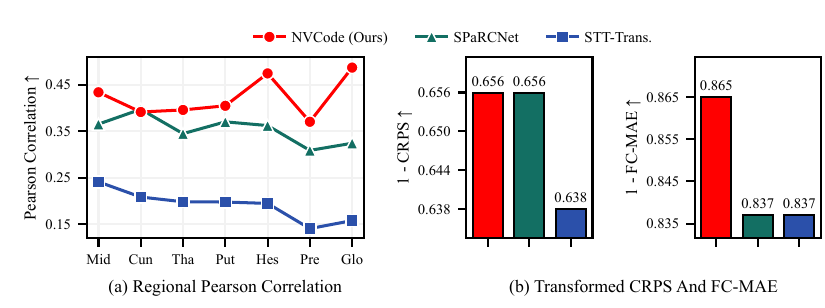}
\caption{\textbf{EEG encoders on held-out subjects.} \textbf{(a)} Pearson $R$ for middle frontal gyrus, cuneus, thalamus, putamen, Heschl's gyrus, precuneus, and global signal, in axis order. \textbf{(b)} One minus each mean error (CRPS and FC-MAE) from the Inter columns of Table~\ref{tab:main-encoder}. Higher bars correspond to smaller marginal and connectivity errors.}
\label{fig:main-encoder-bars}
\end{minipage}
\end{table}

\begin{table}[t]
{
\centering
\captionsetup{width=.931\linewidth,justification=justified}
\caption{\textbf{Conditional generator comparison.} All networks use the same cached EEG codes from seven frozen encoders and the frozen NVCode projector. Each evaluation weights all eight datasets equally.}
\label{tab:main-generator}
\resizebox{.931\linewidth}{!}{%
\begin{BenchmarkTable}
\toprule
\multirow{2}{*}{\textbf{Generator}} & \multicolumn{3}{c|}{\textbf{Intra-Subject}} & \multicolumn{3}{c}{\textbf{Inter-Subject}} \\
& \textbf{Avg.~$R\uparrow$} & \textbf{CRPS$\downarrow$} & \textbf{FC-MAE$\downarrow$}
& \textbf{Avg.~$R\uparrow$} & \textbf{CRPS$\downarrow$} & \textbf{FC-MAE$\downarrow$} \\
\midrule
cGAN~\citep{mirza2014conditional} & 0.310 & 0.462 & 0.505 & 0.211 & 0.368 & 0.297 \\
Consistency~\citep{yang2024consistency} & 0.328 & 0.452 & 0.357 & 0.283 & 0.431 & 0.240 \\
CVAE~\citep{sohn2015learning} & \Second{0.404} & 0.378 & 0.336 & 0.280 & 0.345 & 0.244 \\
DDPM~\citep{ho2020denoising} & 0.302 & \Second{0.358} & \Second{0.309} & 0.197 & 0.346 & \Second{0.170} \\
EDM~\citep{karras2022elucidating} & 0.351 & 0.361 & 0.323 & \Second{0.286} & \Second{0.343} & 0.211 \\
Rectified Flow~\citep{liu2022flow} & 0.302 & 0.364 & 0.328 & 0.234 & 0.350 & 0.242 \\
Score SDE~\citep{song2020score} & 0.343 & 0.361 & 0.346 & 0.280 & \Best{0.339} & 0.202 \\
\addlinespace[1pt]
\textbf{NVCode (Ours)} & \Best{0.480} & \Best{0.352} & \Best{0.280} & \Best{0.423} & 0.344 & \Best{0.135} \\
\ResultBottomRule
\end{BenchmarkTable}}\par
}

\par
\begin{minipage}{\linewidth}
\centering
\captionsetup{type=figure}
\includegraphics[width=.95\linewidth]{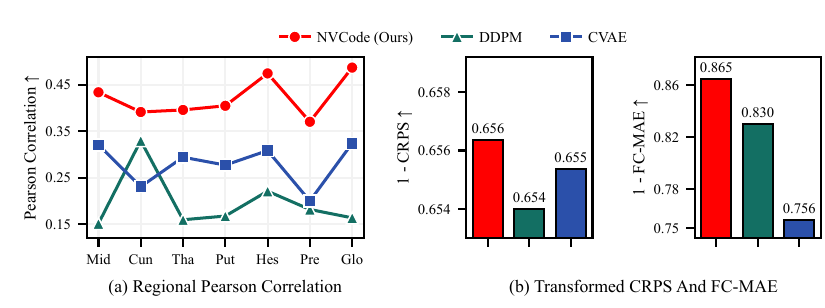}
\caption{\textbf{Generators on held-out subjects.} \textbf{(a)} Pearson $R$ for NVCode, DDPM, and CVAE given the same cached EEG codes. Targets follow Figure~\ref{fig:main-encoder-bars}. \textbf{(b)} One minus each mean error (CRPS and FC-MAE) from Table~\ref{tab:main-generator}, so higher bars indicate lower error. The eight datasets receive equal weight in each held-out-subject summary.}
\label{fig:main-generator-lines}
\end{minipage}
\end{table}
\subsection{Main Experiment}
\label{sec:exp-main}
\label{sec:exp-results}

\myparagraph{Accuracy.}
We first evaluate reconstruction for participants represented during training (Intra). NVCode achieves the highest average Pearson correlation, $0.480$, in both component comparisons. With the same generator architecture fitted separately for each encoder, NVCode exceeds SIREN-based, the best encoder baseline ($0.403$), by $19.1\%$ (Table~\ref{tab:main-encoder}). With identical cached EEG codes, it exceeds CVAE, the best generation baseline ($0.404$), by $18.8\%$ (Table~\ref{tab:main-generator}).

NVCode retains its correlation advantage when evaluated on unseen participants. Under Inter evaluation, it reaches $R=0.423$ across eight equally weighted datasets. This is $19.8\%$ higher than SPaRCNet ($0.353$) in the encoder comparison (Table~\ref{tab:main-encoder}) and $47.9\%$ higher than EDM ($0.286$) with identical cached EEG codes (Table~\ref{tab:main-generator}). The larger generator gap in Inter makes it especially useful to separate the two controls: one changes which EEG features reach the generator, and the other changes how the generator uses a fixed input. Both controls evaluate new participants within each dataset's acquisition setting.

\Needspace{6\baselineskip}
\myparagraph{Encoder comparison.}
We vary the EEG representation while keeping the generator architecture fixed. On held-out participants, NVCode improves on SPaRCNet in six of seven targets (Figure~\ref{fig:main-encoder-bars}a). The largest gain is in the global signal, from $0.324$ to $0.487$; both subcortical targets also improve. The per-target results in Table~\ref{tab:app-main-combined} show how these gains contribute to the average across targets.

The encoder change also improves predictive-mean connectivity. Relative to SPaRCNet, FC-MAE falls from $0.163$ to $0.135$, a $17.2\%$ reduction, while marginal CRPS stays at $0.344$ to the reported precision (Figure~\ref{fig:main-encoder-bars}b). Thus, at fixed generator architecture, gains in how the signals vary together can occur with little change in the marginal scores for the individual targets.

\myparagraph{Generator comparison.}
We next hold the conditioning input fixed and vary the generator. All baselines in Table~\ref{tab:main-generator} receive the same cached codes from the NVCode projector, so differences arise after the EEG representation has already been formed. The leading alternative in Avg.~$R$ changes from CVAE in Intra to EDM in Inter, whereas NVCode retains the highest correlation under both protocols. Figure~\ref{fig:main-generator-lines}a shows that this advantage extends across all seven signals relative to DDPM and CVAE on held-out participants. Heschl's gyrus, for example, reaches $0.475$ versus $0.309$ for CVAE, and the thalamus reaches $0.396$ versus $0.294$. The gains therefore span cortical and subcortical targets as well as the global time series.

The benchmark gives each dataset the same weight regardless of its number of scans. Tables~\ref{tab:app:generator-all-datasets}--\ref{tab:app:generator-all-datasets:natview} separate the dataset means and show seed variation where reported. This breakdown distinguishes the benchmark average from performance on any one dataset.

\NVResultLeft
With those same cached inputs, NVCode attains the lowest FC-MAE, reducing the Inter error by $20.6\%$ relative to DDPM. In Figure~\ref{fig:main-generator-lines}b, its marginal CRPS is also lower than those of DDPM and CVAE, although Score SDE leads on that metric in Table~\ref{tab:main-generator}. These tests separate two uses of a conditional generator: tracking each recorded trajectory and recovering the correlations across the predictive means. NVCode improves both under the same held-out evaluation and fixed EEG code used for the generator comparison.
\NVResultRight
\centering
\NVResultTableCaption
\caption{\textbf{End-to-end comparison.} All methods train on the same pool of eight EEG--fMRI datasets. The scores use matched Intra and Inter splits for all three methods.}
\label{tab:end-to-end}
\small
\renewcommand{\arraystretch}{1.29}
\setlength{\tabcolsep}{5.8pt}
\resizebox{\linewidth}{!}{%
\begin{tabular}{@{}clccc@{}}
\toprule
 & \textbf{Method} & \textbf{Avg.~$R\uparrow$} & \textbf{CRPS$\downarrow$} & \textbf{FC-MAE$\downarrow$} \\
\midrule
\multirow[c]{3}{*}{\rotatebox[origin=c]{90}{\textbf{Intra}}} & NeuroBOLT~\citep{li2024neurobolt} & 0.440 & 0.400 & 0.323 \\
 & UnEBOLT~\citep{li2026unebolt} & \Second{0.475} & \Second{0.361} & \SecondRightEdge{0.287} \\
 & \textbf{NVCode (Ours)} & \Best{0.500} & \Best{0.247} & \BestRightEdge{0.210} \\[\aboverulesep]
\specialrule{\lightrulewidth}{0pt}{\belowrulesep}
\multirow[c]{3}{*}{\rotatebox[origin=c]{90}{\textbf{Inter}}} & NeuroBOLT~\citep{li2024neurobolt} & 0.302 & 0.473 & 0.385 \\
 & UnEBOLT~\citep{li2026unebolt} & \Second{0.403} & \Second{0.384} & \SecondRightEdge{0.262} \\
 & \textbf{NVCode (Ours)} & \Best{0.431} & \Best{0.316} & \BestRightEdge{0.172} \\[\aboverulesep]
\specialrule{\heavyrulewidth}{0pt}{\belowbottomsep}
\end{tabular}%
}

\NVResultEnd

\myparagraph{Pooled training.}
Can a single model preserve the reconstruction advantage across acquisition settings and tasks? Table~\ref{tab:end-to-end} replaces separate dataset fits with a common training pool drawn from all eight datasets. The Inter protocol still holds out participants rather than entire datasets, so this comparison tests generalization to new participants within a common multi-dataset training setup. The seven prediction targets stay the same when training examples come from multiple corpora. NVCode leads on Avg.~$R$, CRPS, and FC-MAE under both splits. Relative to UnEBOLT, Avg.~$R$ rises from $0.475$ to $0.500$ in Intra and from $0.403$ to $0.431$ in Inter. The Inter gains extend to the output distributions and their average trajectories: CRPS decreases from $0.384$ to $0.316$, and FC-MAE from $0.262$ to $0.172$. The pooled model also attains the highest correlation in five of seven signals in either split. On held-out participants, this roughly one-third reduction in FC-MAE accompanies gains in the individual time courses. The benefit thus spans several targets and persists under pooled training. The next control returns to fixed encoder features and gives each generator a separate conditioning map on the same held-out participants.

\begin{figure}[!t]
\centering
\includegraphics[width=.95\linewidth]{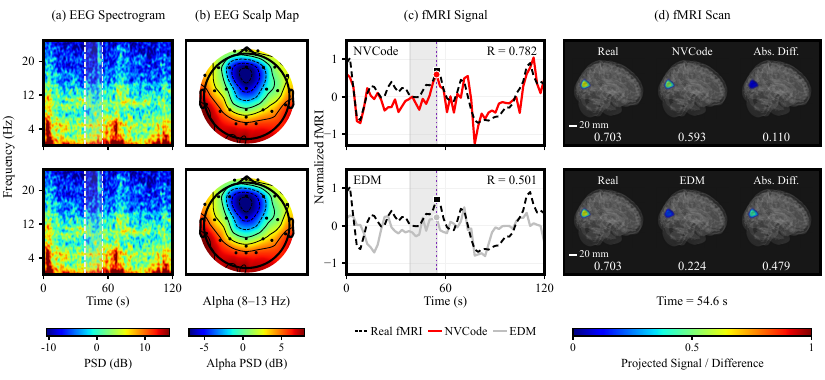}
\caption{\textbf{Matched EEG--fMRI evidence for Cuneus.} \textbf{(a, b):} Identical EEG inputs in both rows. \textbf{(c):} Real fMRI (black dashed), NVCode (top, red), and EDM (bottom, grey), with correlations. \textbf{(d):} Real, predicted, and absolute-error amplitudes projected onto Cuneus at $54.6$\,s (purple guides) with a shared color scale.}
\label{fig:timeseries}

\end{figure}

\NVResultLeft[12.24pt]
\textbf{Generator-specific projectors.}\enspace We fit DDPM, EDM and Score SDE with separate projectors of dimension $512$, retaining the same frozen EEG features (Table~\ref{tab:own-projector-comparison}). This changes the conditioning map but keeps its input and the set of participants used for held-out evaluation fixed.

NVCode retains the highest Avg.~$R$ ($0.423$) and lowest FC-MAE ($0.135$) when the alternatives adapt their own projectors. Score SDE leads on marginal CRPS ($0.310$), so the best generator depends on which property the metric tests. Appendix~\ref{app:exp:projector-control} gives the two training protocols.
\NVResultRight
\NVResultTableCaption
\begingroup
\centering
\caption{\textbf{Generator-specific projectors.} Frozen uses the cached codes; Own jointly fits the projector and generator. Held-out scores give all eight datasets equal weight.}
\label{tab:own-projector-comparison}
\small
\setlength{\tabcolsep}{4pt}
\renewcommand{\arraystretch}{1.12}
% Put separator padding inside the adjoining rows so cell colors reach the rules.
\makeatletter
\newcommand{\ProjectorCellTop}{%
  \rule{0pt}{\dimexpr\ht\@arstrutbox+\belowrulesep\relax}}
\newcommand{\ProjectorCellBottom}[1]{%
  \rule[-\dimexpr\dp\@arstrutbox+#1\relax]{0pt}{\dimexpr\dp\@arstrutbox+#1\relax}}
\makeatother
\resizebox{\linewidth}{!}{%
\begin{tabular}{@{}llccc@{}}
\toprule
\textbf{Generator} & \textbf{Projector} & \textbf{Avg.~$R$}$\uparrow$ & \textbf{CRPS}$\downarrow$ & \textbf{FC-MAE}$\downarrow$ \\
\specialrule{\lightrulewidth}{\aboverulesep}{0pt}
\multirow{2}{*}{DDPM} & Frozen\ProjectorCellTop & 0.197 & 0.346 & \SecondRightEdge{0.170} \\
 & Own\ProjectorCellBottom{1pt} & 0.172 & \Second{0.319} & 0.195 \\
\multirow{2}{*}{EDM} & Frozen & 0.286 & 0.343 & 0.211 \\
 & Own\ProjectorCellBottom{1pt} & \Second{0.299} & 0.381 & 0.207 \\
\multirow{2}{*}{Score SDE} & Frozen & 0.280 & 0.339 & 0.202 \\
 & Own\ProjectorCellBottom{\aboverulesep} & 0.249 & \Best{0.310} & 0.246 \\
\specialrule{\lightrulewidth}{0pt}{0pt}
\textbf{NVCode} & Own\ProjectorCellTop & \Best{0.423} & 0.344 & \BestRightEdge{0.135} \\
\ResultBottomRule
\end{tabular}%
}
\par
\endgroup

\NVResultEnd

The three baselines kept their correlation ranking, but the error metrics responded differently. EDM's Avg.~$R$ increased from $0.286$ to $0.299$ while marginal CRPS rose from $0.343$ to $0.381$. DDPM and Score SDE instead reduced CRPS at lower correlation. Thus a learned conditioning map can improve one aspect of synthesis without improving another. Across these projector settings, the same NVCode configuration led in temporal correlation and connectivity, while the best marginal score came from Score SDE. The fixed-projector result therefore establishes a common comparison point, and the adaptive comparison shows which advantages persist when baselines can change their conditioning map.

\subsection{Analysis}
\label{sec:exp-analysis}

\myparagraph{Temporal fidelity.}
Figure~\ref{fig:timeseries} compares two reconstructions of the same Cuneus signal. Both models receive the same EEG, as reflected by the identical spectrograms and alpha scalp maps. NVCode recovers local reversals and the late positive excursion more closely than EDM, giving $R=0.782$ rather than $0.501$. Correlation measures agreement over the whole segment; the marked time point tests whether that agreement also translates into an accurate amplitude estimate at one time point in the recorded Cuneus signal.

At $54.6$\,s, NVCode predicts $0.593$ and EDM predicts $0.224$ against a measured value of $0.703$. The absolute error drops from $0.479$ to $0.110$ ($77.0\%$). Panel (d) uses the same anatomical weights and color scale for both models to visualize regional amplitude errors at that instant. The shared geometry isolates differences in predicted amplitude from differences in anatomical support.

\begin{table}[!t]
\centering
\captionsetup{width=.95\linewidth,justification=justified}
\caption{\textbf{Synthetic fMRI transfers to a classifier trained only on real fMRI.} The first row gives real input accuracy; synthetic rows report the change $\Delta$ using predictions from matched EEG epochs. Classifier parameters stay fixed, and Mean weights all eight datasets equally in both the real and synthetic test conditions.}
\label{tab:downstream_head}
\renewcommand\arraystretch{1.12}
\setlength{\tabcolsep}{2.8pt}
\resizebox{.95\linewidth}{!}{%
\small
\begin{tabular}{U}
\toprule
\multirow{2}{*}{\textbf{Method}} & \multicolumn{8}{c|}{\textbf{Dataset accuracy / $\Delta$}} & \multirow{2}{*}{\textbf{Mean}} \\
& \NVUtilityDatasetHeads & \\
\midrule
Real test & 0.861 & 0.787 & 0.746 & 0.869 & 0.874 & 0.782 & 0.747 & 0.702 & 0.796 \\
\midrule
cGAN~\citep{mirza2014conditional} & +0.022 & +0.018 & +0.013 & +0.014 & +0.008 & +0.019 & +0.027 & +0.002 & +0.015 \\
Consistency~\citep{yang2024consistency} & +0.017 & \Second{+0.024} & +0.026 & +0.019 & +0.010 & +0.011 & +0.021 & +0.014 & +0.018 \\
CVAE~\citep{sohn2015learning} & +0.011 & +0.016 & \Second{+0.034} & \Second{+0.037} & +0.017 & +0.005 & +0.033 & +0.012 & +0.021 \\
DDPM~\citep{ho2020denoising} & \Second{+0.026} & +0.019 & +0.020 & +0.035 & \Best{+0.024} & +0.023 & +0.032 & +0.014 & \Second{+0.024} \\
EDM~\citep{karras2022elucidating} & +0.007 & +0.021 & +0.026 & +0.022 & +0.009 & \Second{+0.025} & +0.038 & +0.008 & +0.020 \\
Rectified Flow~\citep{liu2022flow} & +0.015 & +0.008 & +0.018 & +0.032 & +0.012 & +0.010 & \Second{+0.041} & +0.011 & +0.018 \\
Score SDE~\citep{song2020score} & +0.021 & +0.016 & +0.027 & +0.029 & +0.015 & +0.006 & +0.026 & \Second{+0.015} & +0.019 \\
\addlinespace[1pt]
\textbf{NVCode (Ours)} & \Best{+0.028} & \Best{+0.027} & \Best{+0.036} & \Best{+0.043} & \Second{+0.019} & \Best{+0.031} & \Best{+0.045} & \Best{+0.020} & \Best{+0.031} \\
\ResultBottomRule
\end{tabular}}
\par

\end{table}

\NVResultLeft
\textbf{Spectral fidelity.}\enspace We next inspect how power is distributed across frequencies in the global signal. Resting BOLD activity is concentrated at low frequencies~\citep{fox2007spontaneous,cordes2001frequencies}. We examine the scan selected for closest alignment with the recorded global signal. NVCode approaches the measured peak and tracks the decay of power (Figure~\ref{fig:spectral-fidelity}), while EDM places its peak at a higher frequency than the peak of the recorded global fMRI signal.

The peak locates the dominant timescale, while the descending branch describes the attenuation of faster fluctuations. These readings complement Figure~\ref{fig:timeseries} by describing global power across frequencies rather than local amplitudes over time.
\NVResultRight
\centering
\captionsetup{type=figure,position=bottom}
\includegraphics[width=0.82\linewidth]{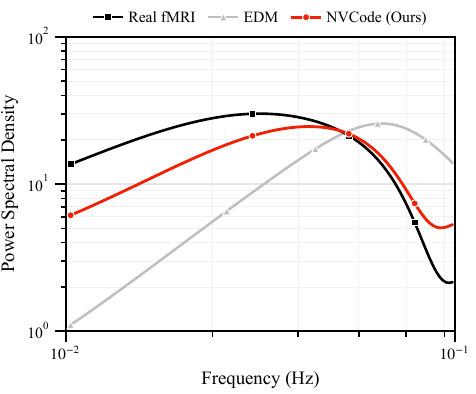}
\caption{\textbf{Global signal spectra.} Power on log axes for real fMRI and the NVCode and EDM predictive means.}
\label{fig:spectral-fidelity}
\NVResultEnd

\subsection{Downstream Utility}
\label{sec:exp-downstream}

\myparagraph{Testing on synthetic fMRI.}
We test whether EEG-conditioned fMRI preserves features that a classifier trained on measured fMRI can use to identify participants. For each dataset, we fit the classifier on real fMRI and keep it fixed for both evaluations. We then replace the held-out real test inputs with predictions from matched EEG epochs while keeping the identity labels and test segments the same. The real-input row in Table~\ref{tab:downstream_head} establishes the reference for each dataset; the remaining rows report accuracy differences relative to that reference. Because the classifier has never seen synthetic fMRI, it cannot adapt its decision rule to the output of any particular generator. The comparison tests transfer across input types within a dataset, using the same participant label set for every generator.

NVCode produces the largest mean gain, $3.1$ percentage points versus $2.4$ for DDPM, and ranks first on seven of the eight datasets. On \XPone, DDPM gains $2.4$ points versus $1.9$ for NVCode. The average gives each dataset equal weight, so the lead reflects gains across acquisition settings rather than the contribution of the largest test set. The table also separates the difficulty of identification on real fMRI from the benefit of replacing those test inputs. For example, NVCode gains $4.3$ points on NeuroBOLT and $4.5$ on NatView, although real-input accuracy differs substantially between the two datasets ($86.9\%$ and $74.7\%$).

All generators raise mean accuracy in this protocol. EEG-conditioned estimates may retain stable features that the classifier uses while attenuating variation in the measured signals that matters less for identity. The experiment measures how readily a fixed classifier can identify participants, whereas the reconstruction analyses measure agreement with the recorded signals. Appendix~\ref{app:downstream-utility} reports the reverse test: classifiers train on synthetic fMRI and are tested on the same held-out real fMRI epochs (Table~\ref{tab:tstr}).

\FloatBarrier
\begin{table}[!t]
\centering
\captionsetup{width=.95\linewidth,justification=justified}
\caption{\textbf{Component ablation on held-out subjects.} Eight variants test all combinations of regional encoders, joint output, and stochastic interpolation. $\checkmark$ and $\times$ indicate whether a component is included or excluded.}
\label{tab:joint-flow-ablation}
\small
\setlength{\tabcolsep}{7pt}
\renewcommand{\arraystretch}{1.10}
\newcommand{\AblationColHead}[1]{\ensuremath{\vcenter{\hbox{\bfseries #1}}}}
\resizebox{.95\linewidth}{!}{%
\begin{tabular}{@{}lccc|ccc@{}}
\toprule
& \multicolumn{3}{c|}{\textbf{Model components}} &
\multicolumn{3}{c}{\textbf{Inter-subject evaluation}} \\
\cmidrule(lr){2-4}\cmidrule(l){5-7}
\AblationColHead{Variant} & \AblationColHead{\shortstack{Regional\\Encoders}} &
\AblationColHead{\shortstack{Joint\\Output}} &
\AblationColHead{\shortstack{Stochastic\\Interpolant}} &
\AblationColHead{Avg.~$R\uparrow$} & \AblationColHead{CRPS$\downarrow$} & \AblationColHead{FC-MAE$\downarrow$} \\
\midrule
Deterministic Joint Regression & $\bm{\checkmark}$ & $\bm{\checkmark}$ & $\bm{\times}$ & 0.317 & 0.368 & 0.250 \\
Independent Interpolants & $\bm{\checkmark}$ & $\bm{\times}$ & $\bm{\checkmark}$ & \Second{0.405} & 0.377 & \SecondRightEdge{0.174} \\
NVCode with a Shared Encoder & $\bm{\times}$ & $\bm{\checkmark}$ & $\bm{\checkmark}$ & 0.349 & 0.362 & 0.182 \\
First-Stage Linear Heads & $\bm{\checkmark}$ & $\bm{\times}$ & $\bm{\times}$ & 0.241 & 0.430 & 0.213 \\
Shared Independent Interpolants & $\bm{\times}$ & $\bm{\times}$ & $\bm{\checkmark}$ & 0.276 & 0.392 & 0.266 \\
Shared Joint Regression & $\bm{\times}$ & $\bm{\checkmark}$ & $\bm{\times}$ & 0.205 & \Second{0.357} & 0.392 \\
Shared Linear Head & $\bm{\times}$ & $\bm{\times}$ & $\bm{\times}$ & 0.119 & 0.428 & 0.485 \\
\specialrule{\lightrulewidth}{\aboverulesep}{0pt}
\textbf{NVCode (Ours)} & $\bm{\checkmark}$ & $\bm{\checkmark}$ & $\bm{\checkmark}$ & \Best{0.423} & \Best{0.344} & \BestRightEdge{0.135} \\
\ResultBottomRule
\end{tabular}}
\par
\end{table}
\FloatBarrier
\Needspace{14\baselineskip}

\subsection{Ablation Study}
\label{sec:exp-ablation}

\NVResultLeft
\textbf{Joint sample quality.}\enspace We shuffle draws by target within each EEG window, preserving the marginal samples and means. Energy score (ES) evaluates the full vector; variogram score (VS) evaluates 21 pairwise contrasts. Original draws reduce ES by $19.3\%$ and VS by $38.3\%$ over shuffling and improve on independent interpolants (Table~\ref{tab:joint-sample-ablation}). Joint coverage is $90.39\%$ at a $90\%$ nominal target for the full response vector.
\NVResultRight
\NVResultTableCaption
\begingroup
\centering
\caption{\textbf{Joint sample quality.} Scores and joint coverage ($90\%$ nominal) on held-out subjects. $\Delta$ is NVCode minus Independent Interpolants; pp denotes percentage points.}
\label{tab:joint-sample-ablation}
\small
\setlength{\tabcolsep}{4pt}
\renewcommand{\arraystretch}{1.03}
\resizebox{\linewidth}{!}{%
\begin{tabular}{@{}lccc@{}}
\toprule
\textbf{Variant} & \textbf{ES}$\downarrow$ & \textbf{VS}$\downarrow$ & \textbf{90\% Joint Cov.} \\
\midrule
Independent Interpolants & 12.63 & 5.39 & 87.41\% \\
NVCode (Shuffled) & \Second{9.36} & \Second{4.72} & \SecondRightEdge{88.36\%} \\
\textbf{NVCode} & \Best{7.55} & \Best{2.91} & \BestRightEdge{90.39\%} \\
\midrule
Paired $\Delta$ & $-5.08$ & $-2.48$ & $+2.98$ pp \\
\ResultBottomRule
\end{tabular}%
}
\par
\endgroup

\NVResultEnd

Predictive-mean FC-MAE uses sequences of means across EEG windows, while joint scores evaluate how targets are paired within a draw. Shuffling preserves the means and thus leaves FC-MAE unchanged. The lower multivariate errors for the original draws reflect the original pairing rather than differences in marginals. Joint coverage requires the full seven-signal observation to lie in the predicted region, rather than testing each target separately (Appendix~\ref{app:ablation:joint-samples}).

Across the four output designs in Table~\ref{tab:joint-flow-ablation}, regional encoders improve Avg.~$R$ and predictive-mean FC-MAE in every comparison, with correlation gains ranging from $0.074$ for joint interpolation to $0.129$ for independent interpolation. Full NVCode leads on all three scores and gives the largest mean identity gain against single-signal representations (Appendix~\ref{app:ablation:single-signal}). Appendix~\ref{app:exp:encoder-capacity} reports the parameter-matched encoder comparison.

\noindent$\vartriangleright$\enspace\textbf{Deterministic Joint Regression.} We replace joint interpolation with a joint MLP, using the same frozen regional features and jointly training the projector (Appendix~\ref{app:ablation:joint-regression}). With interpolation, Avg.~$R$ rises from $0.317$ to $0.423$, FC-MAE falls from $0.250$ to $0.135$, and marginal CRPS falls from $0.368$ to $0.344$. Both designs predict the target vector jointly, so the comparison asks whether learning a conditional distribution improves on a joint point predictor. The gain spans the individual trajectories, their predictive-mean connectivity, and the quality of the predictive distributions for the individual targets.

\Needspace{5\baselineskip}
\noindent$\vartriangleright$\enspace\textbf{Independent Interpolants.} We test whether the interpolant benefits from the other signals' evolving states. The seven scalar interpolants use the same regional-feature condition but evolve each target independently (Appendix~\ref{app:ablation:independent-interpolants}). With joint state access, Avg.~$R$ rises from $0.405$ to $0.423$ and FC-MAE falls from $0.174$ to $0.135$. This contrast changes how the outputs interact while retaining the same EEG representation.

\noindent$\vartriangleright$\enspace\textbf{NVCode with a Shared Encoder.} We use one jointly supervised encoder, retaining the $1400$-dimensional interface and joint interpolant (Appendix~\ref{app:ablation:shared-encoder}). Regional encoders raise Avg.~$R$ from $0.349$ to $0.423$ and lower FC-MAE from $0.182$ to $0.135$. Separately supervised features thus provide more effective conditioning for this output design. The same ordering also holds for three other predictors: direct linear heads, deterministic joint regression and interpolation with independent scalar fields.

\noindent$\vartriangleright$\enspace\textbf{First-Stage Linear Heads.} Without a generator, regional heads reach Avg.~$R=0.241$ versus $0.119$ for the shared head, and FC-MAE falls from $0.485$ to $0.213$ (Table~\ref{tab:joint-flow-ablation}; Appendix~\ref{app:ablation:regional-heads}). With the same frozen regional encoders, the joint interpolant raises Avg.~$R$ to $0.423$.

\noindent$\vartriangleright$\enspace\textbf{Shared Independent Interpolants.} We fit scalar interpolants on frozen shared features, using the shared joint-interpolation encoder (Appendix~\ref{app:ablation:shared-independent}). Joint state access raises Avg.~$R$ from $0.276$ to $0.349$ and lowers FC-MAE from $0.266$ to $0.182$. Marginal CRPS also decreases from $0.392$ to $0.362$, extending the improvement from mean trajectories to marginal probabilistic prediction with one encoder. This output-design benefit therefore holds for both regional and shared EEG representations.

\noindent$\vartriangleright$\enspace\textbf{Shared Joint Regression.} We fit a joint deterministic regressor on the shared encoder features (Appendix~\ref{app:ablation:shared-regression}). Its marginal CRPS of $0.357$ ranks second and equals absolute prediction error for this deterministic control. With the same encoder, joint interpolation raises Avg.~$R$ from $0.205$ to $0.349$ and lowers FC-MAE from $0.392$ to $0.182$. Joint interpolation therefore improves regional trajectories and predictive-mean connectivity; full NVCode achieves the best scores across all three metrics.

\noindent$\vartriangleright$\enspace\textbf{Shared Linear Head.} We retain the shared encoder's linear head (Appendix~\ref{app:ablation:shared-head}). With identical frozen features, Avg.~$R$ rises from $0.119$ to $0.205$ with joint regression, $0.276$ with independent interpolation and $0.349$ with joint interpolation. Joint interpolation reduces FC-MAE from $0.485$ to $0.182$. Among the shared variants, joint interpolation attains the highest temporal correlation and lowest connectivity error. Because all four predictors receive the same frozen EEG features, this comparison isolates the effect of changing the output design while retaining the same learned EEG representation.

\endgroup
\section{Conclusion}
We presented \textsc{Neurovascular Codes} for EEG-to-fMRI synthesis. NVCode learns target-specific EEG features through individual fMRI prediction tasks and projects them into a shared code for stochastic fMRI generation. Across eight public datasets, it achieves the highest average Pearson correlation and the lowest predictive-mean FC-MAE in both encoder and generator comparisons, under within-subject and held-out-subject evaluation. Ablations favor regional conditioning combined with joint interpolation. Using fixed classifiers trained on real fMRI, NVCode signals from matched EEG epochs achieve the highest mean subject-identification accuracy among the generators evaluated under the same classification protocol.

\myparagraph{Discussion.} More broadly, our results suggest that learning to predict fMRI can organize EEG representations for conditional signal generation. Reciprocal generation would ask which information is shared between the two modalities. An EEG window could be mapped to a seven-signal fMRI response and then reconstructed from that response, testing which features survive the round trip. A cycle beginning in fMRI could test whether an inferred EEG signal supports recovery of the original response pattern. Comparing each cycle against mismatched windows would distinguish information about the current window from patterns that recur across a scan. Scoring regional trajectories and cross-target structure separately would distinguish features recovered in both directions from features supported by only one mapping. A cross-dataset cycle could test whether these shared features remain recoverable when both mappings are evaluated on a new acquisition setting without refitting. Scores for the two directions could then be reported separately against the same mappings evaluated on matched windows in the original corpus. This would distinguish errors tied to one direction from a loss of shared information when the dataset changes.

\FloatBarrier
\setcounter{topnumber}{2}
\bibliographystyle{plainnat}
\bibliography{references/nvcode,references/nvcode_extra}

@article{kovalev2022fmri,
  title={fMRI from EEG is only Deep Learning away: the use of interpretable DL to unravel EEG-fMRI relationships},
  author={Kovalev, Alexander and Mikheev, Ilia and Ossadtchi, Alexei},
  journal={arXiv preprint arXiv:2211.02024},
  year={2022}
}

@article{yang2023biot,
  title={Biot: Biosignal transformer for cross-data learning in the wild},
  author={Yang, Chaoqi and Westover, M and Sun, Jimeng},
  journal={Advances in Neural Information Processing Systems},
  volume={36},
  pages={78240--78260},
  year={2023}
}

@article{xiao2025brainomni,
  title={Brainomni: A brain foundation model for unified eeg and meg signals},
  author={Xiao, Qinfan and Cui, Ziyun and Zhang, Chi and Chen, Siqi and Wu, Wen and Thwaites, Andrew and Woolgar, Alexandra and Zhou, Bowen and Zhang, Chao},
  journal={arXiv preprint arXiv:2505.18185},
  year={2025}
}

@inproceedings{peh2022transformer,
  title={Transformer convolutional neural networks for automated artifact detection in scalp EEG},
  author={Peh, Wei Yan and Yao, Yuanyuan and Dauwels, Justin},
  booktitle={2022 44th Annual International Conference of the IEEE Engineering in Medicine \& Biology Society (EMBC)},
  pages={3599--3602},
  year={2022},
  organization={IEEE}
}

@article{yang2023self,
  title={Self-supervised electroencephalogram representation learning for automatic sleep staging: model development and evaluation study},
  author={Yang, Chaoqi and Xiao, Cao and Westover, M Brandon and Sun, Jimeng},
  journal={JMIR AI},
  volume={2},
  number={1},
  pages={e46769},
  year={2023},
  publisher={JMIR Publications Inc., Toronto, Canada}
}

@article{li2022motor,
  title={Motor imagery EEG classification algorithm based on CNN-LSTM feature fusion network},
  author={Li, Hongli and Ding, Man and Zhang, Ronghua and Xiu, Chunbo},
  journal={Biomedical signal processing and control},
  volume={72},
  pages={103342},
  year={2022},
  publisher={Elsevier}
}

@article{jiang2024large,
  title={Large brain model for learning generic representations with tremendous EEG data in BCI},
  author={Jiang, Wei-Bang and Zhao, Li-Ming and Lu, Bao-Liang},
  journal={arXiv preprint arXiv:2405.18765},
  year={2024}
}

@inproceedings{li2024leveraging,
  title={Leveraging sinusoidal representation networks to predict fMRI signals from EEG},
  author={Li, Yamin and Lou, Ange and Xu, Ziyuan and Wang, Shiyu and Chang, Catie},
  booktitle={Medical Imaging 2024: Image Processing},
  volume={12926},
  pages={795--800},
  year={2024},
  organization={SPIE}
}

@article{li2024neurobolt,
  title={Neurobolt: Resting-state eeg-to-fmri synthesis with multi-dimensional feature mapping},
  author={Li, Yamin and Lou, Ange and Xu, Ziyuan and Zhang, Shengchao and Wang, Shiyu and Englot, Dario J and Kolouri, Soheil and Moyer, Daniel and Bayrak, Roza G and Chang, Catie},
  journal={Advances in neural information processing systems},
  volume={37},
  pages={23378--23405},
  year={2024}
}

@article{ouahidi2025reve,
  title={REVE: A Foundation Model for EEG--Adapting to Any Setup with Large-Scale Pretraining on 25,000 Subjects},
  author={Ouahidi, Yassine El and Lys, Jonathan and Th{\"o}lke, Philipp and Farrugia, Nicolas and Pasdeloup, Bastien and Gripon, Vincent and Jerbi, Karim and Lioi, Giulia},
  journal={arXiv preprint arXiv:2510.21585},
  year={2025}
}

@article{jing2023development,
  title={Development of expert-level classification of seizures and rhythmic and periodic patterns during EEG interpretation},
  author={Jing, Jin and Ge, Wendong and Hong, Shenda and Fernandes, Marta Bento and Lin, Zhen and Yang, Chaoqi and An, Sungtae and Struck, Aaron F and Herlopian, Aline and Karakis, Ioannis and others},
  journal={Neurology},
  volume={100},
  number={17},
  pages={e1750--e1762},
  year={2023},
  publisher={Lippincott Williams \& Wilkins Hagerstown, MD}
}

@article{song2021transformer,
  title={Transformer-based spatial-temporal feature learning for EEG decoding},
  author={Song, Yonghao and Jia, Xueyu and Yang, Lie and Xie, Longhan},
  journal={arXiv preprint arXiv:2106.11170},
  year={2021}
}

@article{mirza2014conditional,
  title={Conditional generative adversarial nets},
  author={Mirza, Mehdi and Osindero, Simon},
  journal={arXiv preprint arXiv:1411.1784},
  year={2014}
}

@article{yang2024consistency,
  title={Consistency flow matching: Defining straight flows with velocity consistency},
  author={Yang, Ling and Zhang, Zixiang and Zhang, Zhilong and Liu, Xingchao and Xu, Minkai and Zhang, Wentao and Meng, Chenlin and Ermon, Stefano and Cui, Bin},
  journal={arXiv preprint arXiv:2407.02398},
  year={2024}
}

@article{ho2020denoising,
  title={Denoising diffusion probabilistic models},
  author={Ho, Jonathan and Jain, Ajay and Abbeel, Pieter},
  journal={Advances in neural information processing systems},
  volume={33},
  pages={6840--6851},
  year={2020}
}

@article{karras2022elucidating,
  title={Elucidating the design space of diffusion-based generative models},
  author={Karras, Tero and Aittala, Miika and Aila, Timo and Laine, Samuli},
  journal={Advances in neural information processing systems},
  volume={35},
  pages={26565--26577},
  year={2022}
}

@article{liu2022flow,
  title={Flow straight and fast: Learning to generate and transfer data with rectified flow},
  author={Liu, Xingchao and Gong, Chengyue and Liu, Qiang},
  journal={arXiv preprint arXiv:2209.03003},
  year={2022}
}

@article{song2020score,
  title={Score-based generative modeling through stochastic differential equations},
  author={Song, Yang and Sohl-Dickstein, Jascha and Kingma, Diederik P and Kumar, Abhishek and Ermon, Stefano and Poole, Ben},
  journal={arXiv preprint arXiv:2011.13456},
  year={2020}
}

@misc{ds003768:1.0.13,
  author = {Yameng Gu AND Feng Han AND Lucas E. Sainburg AND Margeaux M. Schade AND Xiao Liu},
  title = {"Simultaneous EEG and fMRI signals during sleep from humans"},
  year = {2026},
  doi = {doi:10.18112/openneuro.ds003768.v1.0.13},
  publisher = {OpenNeuro}
}

@misc{ds005795:1.0.0,
  author = {Jörg Stadler AND Torsten Stöter AND Nicole Angenstein AND Andreas Fügner AND Marcel Lommerzheim AND Artur Mathysiak AND Anke Michalsky AND Gabriele Schöps AND Johann van der Meer AND Susann Wolff AND André Brechmann},
  title = {"MULTI-CLARID (Multimodal Category Learning and Resting-state Imaging Data)"},
  year = {2025},
  doi = {doi:10.18112/openneuro.ds005795.v1.0.0},
  publisher = {OpenNeuro}
}

@misc{ds006040:1.0.2,
  author = {Younghwa Cha AND Yeji Lee AND Eunhee Ji AND SoHyun Han AND Sunhyun Min AND Hyoungkyu Kim AND Minseo Cho AND Hae Seong Lee AND Youngjai Park AND Joon-Young Moon},
  title = {"Sustained Attention Task (gradCPT) Dataset using simultaneous EEG-fMRI and DTI"},
  year = {2026},
  doi = {doi:10.18112/openneuro.ds006040.v1.0.2},
  publisher = {OpenNeuro}
}

@misc{ds002336:2.0.2,
  author = {Giulia Lioi AND Claire Cury AND Lorraine Perronnet AND Marsel Mano AND Elise Bannier AND Anatole Lecuyer AND Christian Barillot},
  title = {"A multi-modal human neuroimaging dataset for data integration: simultaneous EEG and fMRI acquisition during a motor imagery neurofeedback task: XP1"},
  year = {2021},
  doi = {10.18112/openneuro.ds002336.v2.0.2},
  publisher = {OpenNeuro}
}

@misc{ds002725:1.0.0,
  author = {Ian Daly AND Nicoletta Nicolaou AND Duncan Williams AND Faustina Hwang AND Alexis Kirke AND Eduardo Miranda AND Slawomir J. Nasuto},
  title = {"A dataset recording joint EEG-fMRI during affective music listening "},
  year = {2020},
  doi = {10.18112/openneuro.ds002725.v1.0.0},
  publisher = {OpenNeuro}
}

@article{telesford2023open,
  title={An open-access dataset of naturalistic viewing using simultaneous EEG-fMRI},
  author={Telesford, Qawi K and Gonzalez-Moreira, Eduardo and Xu, Ting and Tian, Yiwen and Colcombe, Stanley J and Cloud, Jessica and Russ, Brian E and Falchier, Arnaud and Nentwich, Maximilian and Madsen, Jens and others},
  journal={Scientific Data},
  volume={10},
  number={1},
  pages={554},
  year={2023},
  publisher={Nature Publishing Group UK London}
}

@misc{ds007216:1.0.0,
  author = {Aaron Kucyi AND Lotus Shareef-Trudeau AND David Braun AND Huiling Peng AND Tiara Bounyarith AND Janet Z. Li},
  title = {"A multi-session simultaneous EEG-fMRI dataset with online experience sampling"},
  year = {2026},
  doi = {doi:10.18112/openneuro.ds007216.v1.0.0},
  publisher = {OpenNeuro}
}

@article{lipman2022flow,
  title={Flow matching for generative modeling},
  author={Lipman, Yaron and Chen, Ricky TQ and Ben-Hamu, Heli and Nickel, Maximilian and Le, Matt},
  journal={arXiv preprint arXiv:2210.02747},
  year={2022}
}

@article{tong2023improving,
  title={Improving and generalizing flow-based generative models with minibatch optimal transport},
  author={Tong, Alexander and Fatras, Kilian and Malkin, Nikolay and Huguet, Guillaume and Zhang, Yanlei and Rector-Brooks, Jarrid and Wolf, Guy and Bengio, Yoshua},
  journal={arXiv preprint arXiv:2302.00482},
  year={2023}
}

@article{chen2018neural,
  title={Neural ordinary differential equations},
  author={Chen, Ricky TQ and Rubanova, Yulia and Bettencourt, Jesse and Duvenaud, David K},
  journal={Advances in neural information processing systems},
  volume={31},
  year={2018}
}

@article{grathwohl2018ffjord,
  title={Ffjord: Free-form continuous dynamics for scalable reversible generative models},
  author={Grathwohl, Will and Chen, Ricky TQ and Bettencourt, Jesse and Sutskever, Ilya and Duvenaud, David},
  journal={arXiv preprint arXiv:1810.01367},
  year={2018}
}

@article{goodfellow2014generative,
  title={Generative adversarial nets},
  author={Goodfellow, Ian J and Pouget-Abadie, Jean and Mirza, Mehdi and Xu, Bing and Warde-Farley, David and Ozair, Sherjil and Courville, Aaron and Bengio, Yoshua},
  journal={Advances in neural information processing systems},
  volume={27},
  year={2014}
}

@article{kingma2013auto,
  title={Auto-encoding variational bayes},
  author={Kingma, Diederik P and Welling, Max},
  journal={arXiv preprint arXiv:1312.6114},
  year={2013}
}

@inproceedings{sohl2015deep,
  title={Deep unsupervised learning using nonequilibrium thermodynamics},
  author={Sohl-Dickstein, Jascha and Weiss, Eric and Maheswaranathan, Niru and Ganguli, Surya},
  booktitle={International conference on machine learning},
  pages={2256--2265},
  year={2015},
  organization={pmlr}
}

@article{song2019generative,
  title={Generative modeling by estimating gradients of the data distribution},
  author={Song, Yang and Ermon, Stefano},
  journal={Advances in neural information processing systems},
  volume={32},
  year={2019}
}

@article{dhariwal2021diffusion,
  title={Diffusion models beat gans on image synthesis},
  author={Dhariwal, Prafulla and Nichol, Alexander},
  journal={Advances in neural information processing systems},
  volume={34},
  pages={8780--8794},
  year={2021}
}

@article{sohn2015learning,
  title={Learning structured output representation using deep conditional generative models},
  author={Sohn, Kihyuk and Lee, Honglak and Yan, Xinchen},
  journal={Advances in neural information processing systems},
  volume={28},
  year={2015}
}

@inproceedings{isola2017image,
  title={Image-to-image translation with conditional adversarial networks},
  author={Isola, Phillip and Zhu, Jun-Yan and Zhou, Tinghui and Efros, Alexei A},
  booktitle={Proceedings of the IEEE conference on computer vision and pattern recognition},
  pages={1125--1134},
  year={2017}
}

@inproceedings{saharia2022palette,
  title={Palette: Image-to-image diffusion models},
  author={Saharia, Chitwan and Chan, William and Chang, Huiwen and Lee, Chris and Ho, Jonathan and Salimans, Tim and Fleet, David and Norouzi, Mohammad},
  booktitle={ACM SIGGRAPH 2022 conference proceedings},
  pages={1--10},
  year={2022}
}

@article{vaswani2017attention,
  title={Attention is all you need},
  author={Vaswani, Ashish and Shazeer, Noam and Parmar, Niki and Uszkoreit, Jakob and Jones, Llion and Gomez, Aidan N and Kaiser, {\L}ukasz and Polosukhin, Illia},
  journal={Advances in neural information processing systems},
  volume={30},
  year={2017}
}

@article{dosovitskiy2020image,
  title={An image is worth 16x16 words: Transformers for image recognition at scale},
  author={Dosovitskiy, Alexey and Beyer, Lucas and Kolesnikov, Alexander and Weissenborn, Dirk and Zhai, Xiaohua and Unterthiner, Thomas and Dehghani, Mostafa and Minderer, Matthias and Heigold, Georg and Gelly, Sylvain and others},
  journal={arXiv preprint arXiv:2010.11929},
  year={2020}
}

@inproceedings{devlin2019bert,
  title={Bert: Pre-training of deep bidirectional transformers for language understanding},
  author={Devlin, Jacob and Chang, Ming-Wei and Lee, Kenton and Toutanova, Kristina},
  booktitle={Proceedings of the 2019 conference of the North American chapter of the association for computational linguistics: human language technologies, volume 1 (long and short papers)},
  pages={4171--4186},
  year={2019}
}

@article{lawhern2018eegnet,
  title={EEGNet: a compact convolutional neural network for EEG-based brain--computer interfaces},
  author={Lawhern, Vernon J and Solon, Amelia J and Waytowich, Nicholas R and Gordon, Stephen M and Hung, Chou P and Lance, Brent J},
  journal={Journal of neural engineering},
  volume={15},
  number={5},
  pages={056013},
  year={2018},
  publisher={iOP Publishing}
}

@article{schirrmeister2017deep,
  title={Deep learning with convolutional neural networks for EEG decoding and visualization},
  author={Schirrmeister, Robin Tibor and Springenberg, Jost Tobias and Fiederer, Lukas Dominique Josef and Glasstetter, Martin and Eggensperger, Katharina and Tangermann, Michael and Hutter, Frank and Burgard, Wolfram and Ball, Tonio},
  journal={Human brain mapping},
  volume={38},
  number={11},
  pages={5391--5420},
  year={2017},
  publisher={Wiley Online Library}
}

@article{supratak2017deepsleepnet,
  title={DeepSleepNet: A model for automatic sleep stage scoring based on raw single-channel EEG},
  author={Supratak, Akara and Dong, Hao and Wu, Chao and Guo, Yike},
  journal={IEEE transactions on neural systems and rehabilitation engineering},
  volume={25},
  number={11},
  pages={1998--2008},
  year={2017},
  publisher={IEEE}
}

@article{phan2022sleeptransformer,
  title={Sleeptransformer: Automatic sleep staging with interpretability and uncertainty quantification},
  author={Phan, Huy and Mikkelsen, Kaare and Ch{\'e}n, Oliver Y and Koch, Philipp and Mertins, Alfred and De Vos, Maarten},
  journal={IEEE Transactions on Biomedical Engineering},
  volume={69},
  number={8},
  pages={2456--2467},
  year={2022},
  publisher={IEEE}
}

@article{roy2019deep,
  title={Deep learning-based electroencephalography analysis: a systematic review},
  author={Roy, Yannick and Banville, Hubert and Albuquerque, Isabela and Gramfort, Alexandre and Falk, Tiago H and Faubert, Jocelyn},
  journal={Journal of neural engineering},
  volume={16},
  number={5},
  pages={051001},
  year={2019},
  publisher={IOP Publishing}
}

@article{banville2021uncovering,
  title={Uncovering the structure of clinical EEG signals with self-supervised learning},
  author={Banville, Hubert and Chehab, Omar and Hyv{\"a}rinen, Aapo and Engemann, Denis-Alexander and Gramfort, Alexandre},
  journal={Journal of Neural Engineering},
  volume={18},
  number={4},
  pages={046020},
  year={2021},
  publisher={IOP Publishing}
}

@article{allen2000method,
  title={A method for removing imaging artifact from continuous EEG recorded during functional MRI},
  author={Allen, Philip J and Josephs, Oliver and Turner, Robert},
  journal={Neuroimage},
  volume={12},
  number={2},
  pages={230--239},
  year={2000},
  publisher={Elsevier}
}

@article{niazy2005removal,
  title={Removal of FMRI environment artifacts from EEG data using optimal basis sets},
  author={Niazy, Rami K and Beckmann, Christian F and Iannetti, Gian Domenico and Brady, J Michael and Smith, Stephen M},
  journal={Neuroimage},
  volume={28},
  number={3},
  pages={720--737},
  year={2005},
  publisher={Elsevier}
}

@article{murta2015electrophysiological,
  title={Electrophysiological correlates of the BOLD signal for EEG-informed fMRI},
  author={Murta, Teresa and Leite, Marco and Carmichael, David W and Figueiredo, Patr{\'\i}cia and Lemieux, Louis},
  journal={Human brain mapping},
  volume={36},
  number={1},
  pages={391--414},
  year={2015},
  publisher={Wiley Online Library}
}

@article{laufs2003eeg,
  title={EEG-correlated fMRI of human alpha activity},
  author={Laufs, Helmut and Kleinschmidt, Andreas and Beyerle, Astrid and Eger, Evelyn and Salek-Haddadi, Afraim and Preibisch, Christine and Krakow, Karsten},
  journal={Neuroimage},
  volume={19},
  number={4},
  pages={1463--1476},
  year={2003},
  publisher={Elsevier}
}

@article{abreu2018eeg,
  title={EEG-informed fMRI: a review of data analysis methods},
  author={Abreu, Rodolfo and Leal, Alberto and Figueiredo, Patr{\'\i}cia},
  journal={Frontiers in human neuroscience},
  volume={12},
  pages={29},
  year={2018},
  publisher={Frontiers Media SA}
}

@article{huster2012methods,
  title={Methods for simultaneous EEG-fMRI: an introductory review},
  author={Huster, Ren{\'e} J and Debener, Stefan and Eichele, Tom and Herrmann, Christoph S},
  journal={Journal of Neuroscience},
  volume={32},
  number={18},
  pages={6053--6060},
  year={2012},
  publisher={Society for Neuroscience}
}

@article{dadi2020fine,
  title={Fine-grain atlases of functional modes for fMRI analysis},
  author={Dadi, Kamalaker and Varoquaux, Ga{\"e}l and Machlouzarides-Shalit, Antonia and Gorgolewski, Krzysztof J and Wassermann, Demian and Thirion, Bertrand and Mensch, Arthur},
  journal={NeuroImage},
  volume={221},
  pages={117126},
  year={2020},
  publisher={Elsevier}
}

@article{glasser2016multi,
  title={A multi-modal parcellation of human cerebral cortex},
  author={Glasser, Matthew F and Coalson, Timothy S and Robinson, Emma C and Hacker, Carl D and Harwell, John and Yacoub, Essa and Ugurbil, Kamil and Andersson, Jesper and Beckmann, Christian F and Jenkinson, Mark and others},
  journal={Nature},
  volume={536},
  number={7615},
  pages={171--178},
  year={2016},
  publisher={Nature Publishing Group}
}

@article{yeo2011organization,
  title={The organization of the human cerebral cortex estimated by intrinsic functional connectivity},
  author={Yeo, BT Thomas and Krienen, Fenna M and Sepulcre, Jorge and Sabuncu, Mert R and Lashkari, Danial and Hollinshead, Marisa and Roffman, Joshua L and Smoller, Jordan W and Z{\"o}llei, Lilla and Polimeni, Jonathan R and others},
  journal={Journal of neurophysiology},
  year={2011},
  publisher={American Physiological Society Bethesda, MD}
}

@article{smith2013resting,
  title={Resting-state fMRI in the human connectome project},
  author={Smith, Stephen M and Beckmann, Christian F and Andersson, Jesper and Auerbach, Edward J and Bijsterbosch, Janine and Douaud, Gwena{\"e}lle and Duff, Eugene and Feinberg, David A and Griffanti, Ludovica and Harms, Michael P and others},
  journal={Neuroimage},
  volume={80},
  pages={144--168},
  year={2013},
  publisher={Elsevier}
}

@article{van2010exploring,
  title={Exploring the brain network: a review on resting-state fMRI functional connectivity},
  author={Van Den Heuvel, Martijn P and Pol, Hilleke E Hulshoff},
  journal={European neuropsychopharmacology},
  volume={20},
  number={8},
  pages={519--534},
  year={2010},
  publisher={Elsevier}
}

@article{bassett2017network,
  title={Network neuroscience},
  author={Bassett, Danielle S and Sporns, Olaf},
  journal={Nature neuroscience},
  volume={20},
  number={3},
  pages={353--364},
  year={2017},
  publisher={Nature Publishing Group US New York}
}

@article{friston2011functional,
  title={Functional and effective connectivity: a review},
  author={Friston, Karl J},
  journal={Brain connectivity},
  volume={1},
  number={1},
  pages={13--36},
  year={2011},
  publisher={SAGE Publications Sage CA: Los Angeles, CA}
}

@article{biswal1995functional,
  title={Functional connectivity in the motor cortex of resting human brain using echo-planar MRI},
  author={Biswal, Bharat and Zerrin Yetkin, F and Haughton, Victor M and Hyde, James S},
  journal={Magnetic resonance in medicine},
  volume={34},
  number={4},
  pages={537--541},
  year={1995},
  publisher={Wiley Online Library}
}

@article{logothetis2008we,
  title={What we can do and what we cannot do with fMRI},
  author={Logothetis, Nikos K},
  journal={Nature},
  volume={453},
  number={7197},
  pages={869--878},
  year={2008},
  publisher={Nature Publishing Group UK London}
}

@article{ogawa1990brain,
  title={Brain magnetic resonance imaging with contrast dependent on blood oxygenation.},
  author={Ogawa, Seiji and Lee, Tso-Ming and Kay, Alan R and Tank, David W},
  journal={proceedings of the National Academy of Sciences},
  volume={87},
  number={24},
  pages={9868--9872},
  year={1990}
}

@article{buzsaki2012brain,
  title={Brain rhythms and neural syntax: implications for efficient coding of cognitive content and neuropsychiatric disease.},
  author={Buzs{\'a}ki, Gy{\"o}rgy and Watson, Brendon O},
  journal={Dialogues in clinical neuroscience},
  volume={14},
  number={4},
  pages={345--367},
  year={2012},
  publisher={Taylor \& Francis}
}

@article{cohen2017does,
  title={Where does EEG come from and what does it mean?},
  author={Cohen, Michael X},
  journal={Trends in neurosciences},
  volume={40},
  number={4},
  pages={208--218},
  year={2017},
  publisher={Elsevier}
}

@inproceedings{takagi2023high,
  title={High-resolution image reconstruction with latent diffusion models from human brain activity},
  author={Takagi, Yu and Nishimoto, Shinji},
  booktitle={Proceedings of the IEEE/CVF conference on computer vision and pattern recognition},
  pages={14453--14463},
  year={2023}
}

@article{scotti2023reconstructing,
  title={Reconstructing the mind's eye: fmri-to-image with contrastive learning and diffusion priors},
  author={Scotti, Paul and Banerjee, Atmadeep and Goode, Jimmie and Shabalin, Stepan and Nguyen, Alex and Dempster, Aidan and Verlinde, Nathalie and Yundler, Elad and Weisberg, David and Norman, Kenneth and others},
  journal={Advances in Neural Information Processing Systems},
  volume={36},
  pages={24705--24728},
  year={2023}
}

@article{gneiting2007strictly,
  title={Strictly proper scoring rules, prediction, and estimation},
  author={Gneiting, Tilmann and Raftery, Adrian E},
  journal={Journal of the American statistical Association},
  volume={102},
  number={477},
  pages={359--378},
  year={2007},
  publisher={Taylor \& Francis}
}

@article{matheson1976scoring,
  title={Scoring rules for continuous probability distributions},
  author={Matheson, James E and Winkler, Robert L},
  journal={Management science},
  volume={22},
  number={10},
  pages={1087--1096},
  year={1976},
  publisher={INFORMS}
}

@article{kingma2014adam,
  title={Adam: A method for stochastic optimization},
  author={Kingma, Diederik P and Ba, Jimmy},
  journal={arXiv preprint arXiv:1412.6980},
  year={2014}
}

@article{loshchilov2017decoupled,
  title={Decoupled weight decay regularization},
  author={Loshchilov, Ilya and Hutter, Frank},
  journal={arXiv preprint arXiv:1711.05101},
  year={2017}
}

@article{ba2016layer,
  title={Layer normalization},
  author={Ba, Jimmy Lei and Kiros, Jamie Ryan and Hinton, Geoffrey E},
  journal={arXiv preprint arXiv:1607.06450},
  year={2016}
}

@article{deligianni2014relating,
  title={Relating resting-state fMRI and EEG whole-brain connectomes across frequency bands},
  author={Deligianni, Fani and Centeno, Maria and Carmichael, David W and Clayden, Jonathan D},
  journal={Frontiers in neuroscience},
  volume={8},
  pages={258},
  year={2014},
  publisher={Frontiers Media SA}
}

@article{calhoun2016multimodal,
  title={Multimodal fusion of brain imaging data: a key to finding the missing link (s) in complex mental illness},
  author={Calhoun, Vince D and Sui, Jing},
  journal={Biological psychiatry: cognitive neuroscience and neuroimaging},
  volume={1},
  number={3},
  pages={230--244},
  year={2016},
  publisher={Elsevier}
}

@article{vidaurre2018discovering,
  title={Discovering dynamic brain networks from big data in rest and task},
  author={Vidaurre, Diego and Abeysuriya, Romesh and Becker, Robert and Quinn, Andrew J and Alfaro-Almagro, Fidel and Smith, Stephen M and Woolrich, Mark W},
  journal={NeuroImage},
  volume={180},
  pages={646--656},
  year={2018},
  publisher={Elsevier}
}

@article{dar2019image,
  title={Image synthesis in multi-contrast MRI with conditional generative adversarial networks},
  author={Dar, Salman UH and Yurt, Mahmut and Karacan, Levent and Erdem, Aykut and Erdem, Erkut and Cukur, Tolga},
  journal={IEEE transactions on medical imaging},
  volume={38},
  number={10},
  pages={2375--2388},
  year={2019},
  publisher={IEEE}
}

@article{ozbey2023unsupervised,
  title={Unsupervised medical image translation with adversarial diffusion models},
  author={{\"O}zbey, Muzaffer and Dalmaz, Onat and Dar, Salman UH and Bedel, Hasan A and {\"O}zturk, {\c{S}}aban and G{\"u}ng{\"o}r, Alper and Cukur, Tolga},
  journal={IEEE Transactions on Medical Imaging},
  volume={42},
  number={12},
  pages={3524--3539},
  year={2023},
  publisher={IEEE}
}

@article{kostas2021bendr,
  title={BENDR: Using transformers and a contrastive self-supervised learning task to learn from massive amounts of EEG data},
  author={Kostas, Demetres and Aroca-Ouellette, Stephane and Rudzicz, Frank},
  journal={Frontiers in Human Neuroscience},
  volume={15},
  pages={653659},
  year={2021},
  publisher={Frontiers Media SA}
}

@inproceedings{cui2024neuro,
  title={Neuro-gpt: Towards a foundation model for eeg},
  author={Cui, Wenhui and Jeong, Woojae and Th{\"o}lke, Philipp and Medani, Takfarinas and Jerbi, Karim and Joshi, Anand A and Leahy, Richard M},
  booktitle={2024 IEEE International Symposium on Biomedical Imaging (ISBI)},
  pages={1--5},
  year={2024},
  organization={IEEE}
}

@article{yi2023learning,
  title={Learning topology-agnostic EEG representations with geometry-aware modeling},
  author={Yi, Ke and Wang, Yansen and Ren, Kan and Li, Dongsheng},
  journal={Advances in Neural Information Processing Systems},
  volume={36},
  pages={53875--53891},
  year={2023}
}

@article{nie2022time,
  title={A time series is worth 64 words: Long-term forecasting with transformers},
  author={Nie, Yuqi and Nguyen, Nam H and Sinthong, Phanwadee and Kalagnanam, Jayant},
  journal={arXiv preprint arXiv:2211.14730},
  year={2022}
}

@article{mcfarland2011brain,
  title={Brain-computer interfaces for communication and control},
  author={McFarland, Dennis J and Wolpaw, Jonathan R},
  journal={Communications of the ACM},
  volume={54},
  number={5},
  pages={60--66},
  year={2011},
  publisher={ACM New York, NY, USA}
}

@article{lotte2018review,
  title={A review of classification algorithms for EEG-based brain--computer interfaces: a 10 year update},
  author={Lotte, Fabien and Bougrain, Laurent and Cichocki, Andrzej and Clerc, Maureen and Congedo, Marco and Rakotomamonjy, Alain and Yger, Florian},
  journal={Journal of neural engineering},
  volume={15},
  number={3},
  pages={031005},
  year={2018},
  publisher={iOP Publishing}
}

@article{bommasani2021opportunities,
  title={On the opportunities and risks of foundation models},
  author={Bommasani, Rishi and Hudson, Drew A and Adeli, Ehsan and Altman, Russ and Arora, Simran and von Arx, Sydney and Bernstein, Michael S and Bohg, Jeannette and Bosselut, Antoine and Brunskill, Emma and others},
  journal={arXiv preprint arXiv:2108.07258},
  year={2021}
}

@article{wang2024eegpt,
  title={Eegpt: Pretrained transformer for universal and reliable representation of eeg signals},
  author={Wang, Guagnyu and Liu, Wenchao and He, Yuhong and Xu, Cong and Ma, Lin and Li, Haifeng},
  journal={Advances in Neural Information Processing Systems},
  volume={37},
  pages={39249--39280},
  year={2024}
}

@article{caro2023brainlm,
  title={BrainLM: A foundation model for brain activity recordings},
  author={Caro, Josue Ortega and Fonseca, Antonio H de O and Averill, Christopher and Rizvi, Syed A and Rosati, Matteo and Cross, James L and Mittal, Prateek and Zappala, Emanuele and Levine, Daniel and Dhodapkar, Rahul M and others},
  journal={BioRxiv},
  pages={2023--09},
  year={2023},
  publisher={Cold Spring Harbor Laboratory}
}

@article{bullock2021artifact,
  title={Artifact reduction in simultaneous EEG-fMRI: a systematic review of methods and contemporary usage},
  author={Bullock, Madeleine and Jackson, Graeme D and Abbott, David F},
  journal={Frontiers in neurology},
  volume={12},
  pages={622719},
  year={2021},
  publisher={Frontiers Media SA}
}

@inproceedings{song2023consistency,
  title={Consistency models},
  author={Song, Yang and Dhariwal, Prafulla and Chen, Mark and Sutskever, Ilya},
  booktitle={Proceedings of the 40th International Conference on Machine Learning},
  pages={32211--32252},
  year={2023}
}

@article{albergo2025stochastic,
  title={Stochastic interpolants: A unifying framework for flows and diffusions},
  author={Albergo, Michael and Boffi, Nicholas M and Vanden-Eijnden, Eric},
  journal={Journal of Machine Learning Research},
  volume={26},
  number={209},
  pages={1--80},
  year={2025}
}

@article{fox2007spontaneous,
  title={Spontaneous fluctuations in brain activity observed with functional magnetic resonance imaging},
  author={Fox, Michael D and Raichle, Marcus E},
  journal={Nature reviews neuroscience},
  volume={8},
  number={9},
  pages={700--711},
  year={2007},
  publisher={Nature Publishing Group UK London}
}

@article{cordes2001frequencies,
  title={Frequencies contributing to functional connectivity in the cerebral cortex in “resting-state” data},
  author={Cordes, Dietmar and Haughton, Victor M and Arfanakis, Konstantinos and Carew, John D and Turski, Patrick A and Moritz, Chad H and Quigley, Michelle A and Meyerand, M Elizabeth},
  journal={American journal of neuroradiology},
  volume={22},
  number={7},
  pages={1326--1333},
  year={2001},
  publisher={American Journal of Neuroradiology}
}

@article{murphy2009impact,
  title={The impact of global signal regression on resting state correlations: are anti-correlated networks introduced?},
  author={Murphy, Kevin and Birn, Rasmus M and Handwerker, Daniel A and Jones, Tyler B and Bandettini, Peter A},
  journal={Neuroimage},
  volume={44},
  number={3},
  pages={893--905},
  year={2009},
  publisher={Elsevier}
}

@article{pearson1895vii,
  title={VII. Note on regression and inheritance in the case of two parents},
  author={Pearson, Karl},
  journal={proceedings of the royal society of London},
  volume={58},
  number={347-352},
  pages={240--242},
  year={1895},
  publisher={The Royal Society London}
}

@article{mcinnes2018umap,
  title={Umap: Uniform manifold approximation and projection for dimension reduction},
  author={McInnes, Leland and Healy, John and Melville, James},
  journal={arXiv preprint arXiv:1802.03426},
  year={2018}
}

@inproceedings{li2026unebolt,
  title={UnEBOLT: A Unified Model for EEG-to-BOLD Translation and Functional Connectivity Reconstruction},
  author={Li, Yamin and Lou, Ange and Li, Chang and Wang, Shiyu and Pourmotabbed, Haatef and Xu, Ziyuan and Zhang, Shengchao and Englot, Dario J and Kolouri, Soheil and Moyer, Daniel and others},
  booktitle={Medical Imaging with Deep Learning},
  year={2026}
}

@article{scheuerer2015variogram,
  author = {Scheuerer, Michael and Hamill, Thomas M.},
  title = {Variogram-Based Proper Scoring Rules for Probabilistic Forecasts of Multivariate Quantities},
  journal = {Monthly Weather Review},
  year = {2015},
  volume = {143},
  number = {4},
  pages = {1321--1334},
  doi = {10.1175/MWR-D-14-00269.1}
}
\clearpage
\appendix
\section*{Appendix}

This appendix presents additional results and documents the experimental and model details.

\begin{itemize}[leftmargin=*, itemsep=1pt]
  \item Appendix~\ref{app:additional} reports results by dataset and illustrates individual scans.
  \item Appendix~\ref{app:experiment} describes the evaluation protocols, comparisons, and ablations.
  \item Appendix~\ref{app:method} gives the derivations, architecture, training, and sampling details.
\end{itemize}

\subsection*{Table of Contents}

\begingroup
\newcommand{\NVCodeAppendixItem}[1]{%
  \textbf{\ref{#1}~~~~\nameref{#1}} \dotfill \pageref{#1}%
}
\newcommand{\NVCodeAppendixSubItem}[1]{%
  \ref{#1}~~~~\nameref{#1} \dotfill \pageref{#1}%
}
\begin{itemize}[label={}, leftmargin=0pt]
  \item \NVCodeAppendixItem{app:additional}
    \begin{itemize}[label={}]
      \item \NVCodeAppendixSubItem{app:additional:aggregate}
      \item \NVCodeAppendixSubItem{app:additional:encoders}
      \item \NVCodeAppendixSubItem{app:additional:generators}
      \item \NVCodeAppendixSubItem{app:additional:scan-atlases}
    \end{itemize}
  \item \NVCodeAppendixItem{app:experiment}
    \begin{itemize}[label={}]
      \item \NVCodeAppendixSubItem{app:setup}
      \item \NVCodeAppendixSubItem{app:main-exp}
      \item \NVCodeAppendixSubItem{app:downstream-utility}
      \item \NVCodeAppendixSubItem{app:ablation-study}
    \end{itemize}
  \item \NVCodeAppendixItem{app:method}
    \begin{itemize}[label={}]
      \item \NVCodeAppendixSubItem{app:method:nvsi-proof}
      \item \NVCodeAppendixSubItem{app:method:enc}
      \item \NVCodeAppendixSubItem{app:method:aggregator}
      \item \NVCodeAppendixSubItem{app:method:velocity}
      \item \NVCodeAppendixSubItem{app:method:optim}
      \item \NVCodeAppendixSubItem{app:method:generation}
    \end{itemize}
\end{itemize}
\endgroup

\clearpage
\newcommand{\NVAppendixText}{%
  \par\noindent\begin{minipage}[t]{.455\linewidth}\vspace{0pt}%
  \setlength{\parindent}{0pt}\setlength{\parskip}{5pt}%
}
\newcommand{\NVAppendixTable}{%
  \par\end{minipage}\hfill\begin{minipage}[t]{.48925\linewidth}\vspace{0pt}%
  \captionsetup{type=table,width=\linewidth,justification=justified,singlelinecheck=false,skip=6pt}%
  \UseCVPRTablePalette
}
\newcommand{\NVAppendixEnd}{\par\end{minipage}\par}

\section{Additional Results}
\label{app:additional}

\begingroup
\renewcommand{\ResultBlockRule}{\specialrule{0.6pt}{0pt}{1.2pt}}
% Match Table 10's on-page type size and share one column grid across all datasets.
\newsavebox{\NVCodeDatasetBox}
\newenvironment{NVCodeDatasetTable}[4]{%
  \begin{table}[H]
  \centering
  \renewcommand{\theContinuedFloat}{\alph{continuedfloat}}%
  \continuedfloat#1%
  \captionsetup{width=.95\linewidth,justification=justified}%
  \caption{\textbf{#2 comparison on #3.}}%
  \label{#4}%
  \UseCVPRTablePalette
  \normalfont\normalsize
  \renewcommand{\ResultStd}[1]{\scalebox{0.86}{\textpm##1}}%
  \renewcommand{\arraystretch}{0.92}%
  \setlength{\tabcolsep}{2.80pt}%
  \begin{lrbox}{\NVCodeDatasetBox}%
  \begin{tabular}{@{}W{c}{12.5pt}W{l}{89.3pt}W{c}{57pt}W{c}{84.1pt}W{c}{84.1pt}%
                        W{c}{59.6pt}W{c}{57pt}W{c}{56.1pt}W{c}{75.3pt}%
                        W{c}{47.3pt}W{c}{42.1pt}W{c}{57.8pt}@{}}
  \ResultTableHeader{#2}%
}{%
  \ResultBottomRule
  \end{tabular}%
  \end{lrbox}%
  \resizebox{.95\linewidth}{!}{\usebox{\NVCodeDatasetBox}}%
  \end{table}%
}
\renewcommand{\ResultTableSetup}{%
  \setlength{\tabcolsep}{4.5pt}%
  \def\arraystretch{1.0}%
}

\subsection{Main EEG to fMRI Synthesis Results}
\label{app:additional:aggregate}

\begin{table}[H]
\centering
\UseCVPRTablePalette
\captionsetup{width=.95\linewidth,justification=justified}
\caption{\textbf{End-to-end EEG-to-fMRI comparison.} Models train on all eight datasets pooled. Avg.~$R$ weights all seven targets equally, CRPS scores the marginals, and FC-MAE compares predictive-mean connectivity.}
\label{tab:app:end-to-end}
\ResultTableSetup
\resizebox{.95\linewidth}{!}{%
\begin{tabular}{@{}clcccccccccc@{}}
\ResultTableHeader{Method}
\multirow{3}{*}{\textbf{Intra}} & \ResultMethod{NeuroBOLT} & 0.418 & 0.392 & 0.405 & 0.436 & \Second{0.477} & 0.439 & \Second{0.514} & 0.440 & 0.400 & 0.323 \\
        & \ResultMethod{UnEBOLT} & \Second{0.436} & \Best{0.493} & \Second{0.472} & \Second{0.476} & 0.435 & \Best{0.503} & 0.509 & \Second{0.475} & \Second{0.361} & \SecondRightEdge{0.287} \\
        & \ResultMethod{NVCode (Ours)} & \Best{0.473} & \Second{0.482} & \Best{0.510} & \Best{0.536} & \Best{0.498} & \Second{0.475} & \Best{0.528} & \Best{0.500} & \Best{0.247} & \BestRightEdge{0.210} \\
\ResultBlockRule
\multirow{3}{*}{\textbf{Inter}} & \ResultMethod{NeuroBOLT} & \Best{0.316} & 0.324 & 0.260 & 0.247 & 0.281 & 0.317 & 0.369 & 0.302 & 0.473 & 0.385 \\
        & \ResultMethod{UnEBOLT} & \Second{0.304} & \Second{0.382} & \Best{0.419} & \Second{0.475} & \Second{0.436} & \Second{0.420} & \Second{0.383} & \Second{0.403} & \Second{0.384} & \SecondRightEdge{0.262} \\
        & \ResultMethod{NVCode (Ours)} & 0.294 & \Best{0.436} & \Second{0.381} & \Best{0.495} & \Best{0.487} & \Best{0.461} & \Best{0.463} & \Best{0.431} & \Best{0.316} & \BestRightEdge{0.172} \\
\ResultBottomRule
\end{tabular}%
}
\end{table}

\begin{table}[H]
\centering
\UseCVPRTablePalette
\captionsetup{width=.95\linewidth,justification=justified}
\caption{\textbf{EEG-to-fMRI synthesis by encoder and generator.} Signal columns report Pearson $R$ for each target. Avg.~$R$, CRPS, and FC-MAE summarize performance across the eight datasets. Ranking applies within each split and comparison. \RankingCaption}
\label{tab:app-main-combined}
\ResultTableSetup
\renewcommand{\arraystretch}{0.92}
% Citations are already given in the main-text comparison tables; omitting the
% repeated marks here keeps the dense all-method summary readable at full width.
\renewcommand{\citep}[1]{}
\resizebox{.95\linewidth}{!}{%
\begin{tabular}{@{}clcccccccccc@{}}
\ResultTableHeader{Encoder}
        \IntraBlock{12} & \ResultMethod{BEIRA~\citep{kovalev2022fmri}} & 0.302 & 0.372 & 0.257 & \Second{0.372} & 0.384 & 0.344 & 0.392 & 0.346 & 0.393 & 0.327 \\
        & \ResultMethod{BIOT~\citep{yang2023biot}} & 0.275 & 0.218 & 0.299 & 0.239 & 0.280 & 0.245 & 0.278 & 0.262 & 0.363 & 0.328 \\
        & \ResultMethod{BrainOmni~\citep{xiao2025brainomni}} & 0.240 & 0.386 & 0.297 & 0.308 & 0.258 & 0.256 & 0.196 & 0.277 & \Second{0.360} & 0.331 \\
        & \ResultMethod{CNN-Trans.~\citep{peh2022transformer}} & 0.247 & \Second{0.415} & 0.264 & 0.309 & 0.362 & 0.314 & 0.332 & 0.320 & 0.363 & 0.332 \\
        & \ResultMethod{ContraWR~\citep{yang2023self}} & 0.311 & 0.370 & 0.351 & 0.314 & 0.334 & 0.264 & 0.356 & 0.329 & 0.390 & 0.348 \\
        & \ResultMethod{FFCL~\citep{li2022motor}} & 0.342 & 0.264 & 0.361 & 0.288 & 0.378 & 0.292 & 0.278 & 0.315 & 0.389 & 0.351 \\
        & \ResultMethod{LaBraM~\citep{jiang2024large}} & 0.280 & 0.201 & 0.280 & 0.282 & 0.303 & 0.278 & 0.231 & 0.265 & \Second{0.360} & 0.319 \\
        & \ResultMethod{SIREN-based~\citep{li2024leveraging}} & \Second{0.443} & 0.406 & \Best{0.407} & 0.321 & \Second{0.438} & \Second{0.404} & 0.400 & \Second{0.403} & 0.383 & \SecondRightEdge{0.315} \\
        & \ResultMethod{REVE~\citep{ouahidi2025reve}} & 0.116 & 0.302 & 0.214 & 0.239 & 0.242 & 0.155 & 0.271 & 0.220 & 0.365 & 0.340 \\
        & \ResultMethod{SPaRCNet~\citep{jing2023development}} & 0.334 & 0.341 & 0.243 & 0.257 & 0.359 & 0.367 & \Second{0.410} & 0.330 & 0.379 & 0.330 \\
        & \ResultMethod{STT-Trans.~\citep{song2021transformer}} & 0.276 & 0.296 & 0.268 & 0.257 & 0.227 & 0.269 & 0.216 & 0.258 & 0.381 & 0.323 \\
        \addlinespace[1pt]
        & \ResultMethod{NVCode (Ours)} & \Best{0.449} & \Best{0.503} & \Second{0.400} & \Best{0.506} & \Best{0.500} & \Best{0.477} & \Best{0.524} & \Best{0.480} & \Best{0.352} & \BestRightEdge{0.280} \\
\ResultBlockRule
        \InterBlock{12} & \ResultMethod{BEIRA~\citep{kovalev2022fmri}} & 0.250 & 0.180 & 0.128 & 0.177 & 0.118 & 0.102 & 0.148 & 0.158 & 0.369 & 0.204 \\
        & \ResultMethod{BIOT~\citep{yang2023biot}} & 0.232 & 0.204 & 0.215 & 0.324 & 0.231 & 0.281 & 0.236 & 0.246 & 0.348 & 0.172 \\
        & \ResultMethod{BrainOmni~\citep{xiao2025brainomni}} & 0.216 & 0.304 & 0.257 & 0.308 & 0.316 & 0.293 & 0.227 & 0.274 & 0.347 & 0.168 \\
        & \ResultMethod{CNN-Trans.~\citep{peh2022transformer}} & 0.309 & 0.299 & 0.186 & 0.242 & 0.334 & 0.235 & 0.144 & 0.250 & 0.347 & 0.190 \\
        & \ResultMethod{ContraWR~\citep{yang2023self}} & 0.192 & 0.128 & 0.355 & 0.333 & 0.240 & 0.230 & 0.189 & 0.238 & 0.358 & 0.221 \\
        & \ResultMethod{FFCL~\citep{li2022motor}} & 0.287 & 0.250 & 0.320 & 0.293 & 0.203 & 0.267 & 0.225 & 0.264 & 0.363 & 0.197 \\
        & \ResultMethod{LaBraM~\citep{jiang2024large}} & 0.304 & 0.346 & 0.289 & 0.332 & 0.290 & 0.286 & 0.280 & 0.304 & 0.345 & 0.187 \\
        & \ResultMethod{SIREN-based~\citep{li2024leveraging}} & 0.346 & 0.347 & 0.215 & \Second{0.338} & 0.266 & 0.307 & 0.289 & 0.301 & 0.354 & 0.189 \\
        & \ResultMethod{REVE~\citep{ouahidi2025reve}} & 0.218 & 0.203 & 0.162 & 0.177 & 0.165 & 0.110 & 0.182 & 0.174 & 0.351 & 0.200 \\
        & \ResultMethod{SPaRCNet~\citep{jing2023development}} & \Best{0.397} & \Second{0.362} & \Second{0.365} & 0.309 & \Second{0.370} & \Second{0.345} & \Second{0.324} & \Second{0.353} & \Second{0.344} & 0.163 \\
        & \ResultMethod{STT-Trans.~\citep{song2021transformer}} & 0.209 & 0.195 & 0.241 & 0.141 & 0.198 & 0.199 & 0.158 & 0.192 & 0.362 & \SecondRightEdge{0.163} \\
        \addlinespace[1pt]
        & \ResultMethod{NVCode (Ours)} & \Second{0.392} & \Best{0.475} & \Best{0.434} & \Best{0.370} & \Best{0.405} & \Best{0.396} & \Best{0.487} & \Best{0.423} & \Best{0.344} & \BestRightEdge{0.135} \\
\ResultTableSubHeader{Generator}
        \IntraBlock{8} & \ResultMethod{cGAN~\citep{mirza2014conditional}} & 0.232 & 0.312 & 0.297 & 0.308 & 0.389 & 0.301 & 0.335 & 0.310 & 0.462 & 0.505 \\
        & \ResultMethod{Consistency~\citep{yang2024consistency}} & 0.294 & 0.315 & 0.362 & 0.333 & 0.318 & 0.314 & 0.360 & 0.328 & 0.452 & 0.357 \\
        & \ResultMethod{CVAE~\citep{sohn2015learning}} & \Second{0.351} & \Second{0.409} & \Best{0.439} & \Second{0.411} & \Second{0.435} & 0.344 & \Second{0.441} & \Second{0.404} & 0.378 & 0.336 \\
        & \ResultMethod{DDPM~\citep{ho2020denoising}} & 0.285 & 0.243 & 0.345 & 0.286 & 0.302 & 0.280 & 0.375 & 0.302 & \Second{0.358} & \SecondRightEdge{0.309} \\
        & \ResultMethod{EDM~\citep{karras2022elucidating}} & 0.329 & 0.331 & 0.352 & 0.323 & 0.383 & \Second{0.378} & 0.363 & 0.351 & 0.361 & 0.323 \\
        & \ResultMethod{Rectified Flow~\citep{liu2022flow}} & 0.338 & 0.288 & 0.282 & 0.356 & 0.316 & 0.311 & 0.223 & 0.302 & 0.364 & 0.328 \\
        & \ResultMethod{Score SDE~\citep{song2020score}} & 0.300 & 0.348 & 0.372 & 0.345 & 0.394 & 0.343 & 0.296 & 0.343 & 0.361 & 0.346 \\
        \addlinespace[1pt]
        & \ResultMethod{NVCode (Ours)} & \Best{0.449} & \Best{0.503} & \Second{0.400} & \Best{0.506} & \Best{0.500} & \Best{0.477} & \Best{0.524} & \Best{0.480} & \Best{0.352} & \BestRightEdge{0.280} \\
\ResultBlockRule
        \InterBlock{8} & \ResultMethod{cGAN~\citep{mirza2014conditional}} & 0.175 & 0.309 & 0.119 & 0.171 & 0.265 & 0.224 & 0.216 & 0.211 & 0.368 & 0.297 \\
        & \ResultMethod{Consistency~\citep{yang2024consistency}} & 0.328 & 0.178 & 0.297 & \Second{0.318} & 0.294 & 0.313 & 0.254 & 0.283 & 0.431 & 0.240 \\
        & \ResultMethod{CVAE~\citep{sohn2015learning}} & 0.231 & 0.309 & \Second{0.321} & 0.200 & 0.277 & 0.294 & 0.324 & 0.280 & 0.345 & 0.244 \\
        & \ResultMethod{DDPM~\citep{ho2020denoising}} & \Second{0.330} & 0.221 & 0.152 & 0.182 & 0.168 & 0.160 & 0.164 & 0.197 & 0.346 & \SecondRightEdge{0.170} \\
        & \ResultMethod{EDM~\citep{karras2022elucidating}} & 0.211 & \Second{0.340} & 0.266 & 0.216 & \Second{0.360} & 0.252 & \Second{0.354} & \Second{0.286} & \Second{0.343} & 0.211 \\
        & \ResultMethod{Rectified Flow~\citep{liu2022flow}} & 0.194 & 0.248 & 0.207 & 0.154 & 0.297 & \Second{0.313} & 0.223 & 0.234 & 0.350 & 0.242 \\
        & \ResultMethod{Score SDE~\citep{song2020score}} & 0.203 & 0.329 & 0.268 & 0.220 & 0.334 & 0.309 & 0.295 & 0.280 & \Best{0.339} & 0.202 \\
        \addlinespace[1pt]
        & \ResultMethod{NVCode (Ours)} & \Best{0.392} & \Best{0.475} & \Best{0.434} & \Best{0.370} & \Best{0.405} & \Best{0.396} & \Best{0.487} & \Best{0.423} & 0.344 & \BestRightEdge{0.135} \\
\ResultBottomRule
\end{tabular}%
}
\end{table}

Table~\ref{tab:app:end-to-end} gives the seven target scores for models trained on pooled data. Table~\ref{tab:app-main-combined} summarizes separate fits within each dataset: encoder comparisons hold the generator architecture fixed, and generation-network comparisons use the same cached EEG codes. Tables~\ref{tab:app:encoder-all-datasets}--\ref{tab:app:encoder-all-datasets:natview} and~\ref{tab:app:generator-all-datasets}--\ref{tab:app:generator-all-datasets:natview} give the individual dataset scores underlying the component comparisons. Section~\ref{app:main-exp} defines the split and dataset averages used in these comparisons.

The target columns show how performance varies within each evaluation split. In the pooled Inter setting, NVCode reaches $R=0.495$ in the precuneus and $R=0.487$ in the putamen. These correlations describe agreement over time for individual signals. The lower FC-MAE of $0.172$ versus $0.262$ for UnEBOLT shows that the pooled results also improve agreement between the predictive-mean and recorded connectivity matrices.

\Needspace{245pt}
\subsection{Per Dataset Encoder Results}
\label{app:additional:encoders}

\setlength{\intextsep}{8pt plus 1pt minus 1pt}
Seven regional encoders and single-encoder baselines use one generator architecture fitted separately for each configuration. Signal columns report the mean $\pm$ standard deviation across seeds $42$, $2026$, and $114514$ in both comparisons. \XPone\ and NatView retained 10 and 2 scans, respectively (Table~\ref{tab:datasets}). With so few scans, the often large standard deviations across seeds offer little basis for comparing methods. We report only the means for these two datasets in both splits. Avg.~$R$, CRPS, and FC-MAE average the same runs. {\UseCVPRTablePalette\RankingCaption}

\begin{NVCodeDatasetTable}{*}{Encoder}{\SleepEEGfMRI\ (Resting State)}{tab:app:encoder-all-datasets}
\IntraBlock{12} & \ResultMethod{BEIRA} & \MeanStd{0.311}{0.161} & \MeanStd{0.263}{0.183} & \BestMeanStd{0.525}{0.186} & \MeanStd{0.372}{0.231} & \MeanStd{0.378}{0.181} & \SecondMeanStd{0.443}{0.175} & \MeanStd{0.302}{0.202} & 0.371 & 0.393 & 0.247 \\*
         & \ResultMethod{BIOT} & \MeanStd{0.239}{0.109} & \MeanStd{0.318}{0.105} & \MeanStd{0.265}{0.107} & \MeanStd{0.217}{0.111} & \MeanStd{0.301}{0.103} & \MeanStd{0.292}{0.126} & \MeanStd{0.252}{0.116} & 0.269 & 0.342 & 0.276 \\*
         & \ResultMethod{BrainOmni} & \MeanStd{0.276}{0.098} & \MeanStd{0.319}{0.158} & \MeanStd{0.301}{0.110} & \MeanStd{0.307}{0.143} & \MeanStd{0.306}{0.122} & \MeanStd{0.262}{0.114} & \MeanStd{0.260}{0.124} & 0.290 & 0.347 & 0.255 \\*
         & \ResultMethod{CNN-Trans.} & \MeanStd{0.351}{0.134} & \MeanStd{0.319}{0.179} & \MeanStd{0.419}{0.177} & \MeanStd{0.387}{0.183} & \MeanStd{0.309}{0.161} & \MeanStd{0.293}{0.126} & \MeanStd{0.336}{0.185} & 0.345 & \Best{0.335} & \SecondRightEdge{0.245} \\*
         & \ResultMethod{ContraWR} & \MeanStd{0.351}{0.195} & \MeanStd{0.308}{0.175} & \MeanStd{0.442}{0.187} & \MeanStd{0.297}{0.239} & \MeanStd{0.396}{0.195} & \MeanStd{0.315}{0.207} & \SecondMeanStd{0.392}{0.203} & 0.357 & 0.361 & 0.281 \\*
         & \ResultMethod{FFCL} & \MeanStd{0.324}{0.200} & \MeanStd{0.418}{0.157} & \SecondMeanStd{0.487}{0.159} & \MeanStd{0.361}{0.221} & \SecondMeanStd{0.487}{0.180} & \MeanStd{0.405}{0.195} & \MeanStd{0.340}{0.193} & \Second{0.403} & 0.367 & 0.290 \\*
         & \ResultMethod{LaBraM} & \MeanStd{0.221}{0.114} & \MeanStd{0.181}{0.121} & \MeanStd{0.228}{0.126} & \MeanStd{0.236}{0.112} & \MeanStd{0.282}{0.088} & \MeanStd{0.271}{0.128} & \MeanStd{0.223}{0.127} & 0.235 & 0.344 & 0.280 \\*
         & \ResultMethod{SIREN-based} & \MeanStd{0.268}{0.196} & \MeanStd{0.423}{0.172} & \MeanStd{0.461}{0.133} & \SecondMeanStd{0.475}{0.198} & \MeanStd{0.384}{0.190} & \MeanStd{0.303}{0.159} & \MeanStd{0.368}{0.153} & 0.383 & 0.358 & 0.259 \\*
         & \ResultMethod{REVE} & \MeanStd{0.275}{0.102} & \MeanStd{0.380}{0.119} & \MeanStd{0.233}{0.133} & \MeanStd{0.299}{0.146} & \MeanStd{0.342}{0.155} & \MeanStd{0.334}{0.108} & \MeanStd{0.304}{0.123} & 0.309 & 0.354 & 0.309 \\*
         & \ResultMethod{SPaRCNet} & \BestMeanStd{0.384}{0.188} & \SecondMeanStd{0.436}{0.179} & \MeanStd{0.381}{0.173} & \MeanStd{0.459}{0.198} & \MeanStd{0.433}{0.151} & \MeanStd{0.366}{0.160} & \MeanStd{0.353}{0.164} & 0.402 & 0.348 & 0.277 \\*
         & \ResultMethod{STT-Trans.} & \MeanStd{0.253}{0.121} & \MeanStd{0.414}{0.160} & \MeanStd{0.387}{0.120} & \MeanStd{0.335}{0.111} & \MeanStd{0.296}{0.130} & \MeanStd{0.239}{0.171} & \MeanStd{0.363}{0.145} & 0.327 & 0.348 & 0.262 \\*
        \addlinespace[1pt]
         & \ResultMethod{NVCode (Ours)} & \SecondMeanStd{0.366}{0.175} & \BestMeanStd{0.522}{0.186} & \MeanStd{0.472}{0.143} & \BestMeanStd{0.524}{0.135} & \BestMeanStd{0.608}{0.118} & \BestMeanStd{0.491}{0.131} & \BestMeanStd{0.534}{0.145} & \Best{0.502} & \Second{0.339} & \BestRightEdge{0.239} \\
\ResultBlockRule
\InterBlock{12} & \ResultMethod{BEIRA} & \MeanStd{0.140}{0.058} & \MeanStd{0.290}{0.233} & \MeanStd{0.248}{0.125} & \MeanStd{0.405}{0.080} & \MeanStd{0.283}{0.203} & \MeanStd{0.170}{0.068} & \MeanStd{0.248}{0.125} & 0.255 & 0.332 & 0.126 \\*
         & \ResultMethod{BIOT} & \MeanStd{0.235}{0.073} & \MeanStd{0.320}{0.127} & \MeanStd{0.334}{0.104} & \MeanStd{0.299}{0.185} & \MeanStd{0.324}{0.119} & \MeanStd{0.302}{0.182} & \MeanStd{0.343}{0.092} & 0.308 & 0.331 & 0.100 \\*
         & \ResultMethod{BrainOmni} & \MeanStd{0.201}{0.108} & \MeanStd{0.219}{0.125} & \MeanStd{0.207}{0.113} & \MeanStd{0.272}{0.221} & \MeanStd{0.332}{0.208} & \MeanStd{0.332}{0.209} & \MeanStd{0.064}{0.054} & 0.232 & 0.338 & \SecondRightEdge{0.093} \\*
         & \ResultMethod{CNN-Trans.} & \MeanStd{0.212}{0.121} & \MeanStd{0.291}{0.315} & \MeanStd{0.227}{0.137} & \MeanStd{0.183}{0.099} & \MeanStd{0.267}{0.211} & \MeanStd{0.152}{0.082} & \MeanStd{0.185}{0.100} & 0.217 & 0.336 & 0.097 \\*
         & \ResultMethod{ContraWR} & \MeanStd{0.153}{0.115} & \MeanStd{0.292}{0.400} & \MeanStd{0.489}{0.094} & \BestMeanStd{0.543}{0.076} & \BestMeanStd{0.573}{0.069} & \MeanStd{0.213}{0.170} & \BestMeanStd{0.677}{0.052} & 0.420 & 0.332 & 0.208 \\*
         & \ResultMethod{FFCL} & \MeanStd{0.221}{0.204} & \MeanStd{0.191}{0.163} & \BestMeanStd{0.582}{0.075} & \MeanStd{0.475}{0.113} & \MeanStd{0.395}{0.179} & \MeanStd{0.266}{0.327} & \SecondMeanStd{0.625}{0.066} & 0.394 & 0.336 & 0.098 \\*
         & \ResultMethod{LaBraM} & \SecondMeanStd{0.310}{0.300} & \MeanStd{0.452}{0.078} & \MeanStd{0.249}{0.165} & \MeanStd{0.358}{0.152} & \MeanStd{0.492}{0.065} & \MeanStd{0.282}{0.258} & \MeanStd{0.378}{0.127} & 0.360 & 0.333 & 0.129 \\*
         & \ResultMethod{SIREN-based} & \BestMeanStd{0.328}{0.366} & \SecondMeanStd{0.539}{0.089} & \MeanStd{0.459}{0.125} & \SecondMeanStd{0.529}{0.093} & \MeanStd{0.437}{0.141} & \BestMeanStd{0.451}{0.131} & \MeanStd{0.401}{0.177} & \Second{0.449} & 0.329 & 0.158 \\*
         & \ResultMethod{REVE} & \MeanStd{0.239}{0.172} & \MeanStd{0.139}{0.086} & \MeanStd{0.228}{0.154} & \MeanStd{0.078}{0.066} & \MeanStd{0.262}{0.223} & \MeanStd{0.180}{0.108} & \MeanStd{0.126}{0.081} & 0.179 & 0.338 & 0.132 \\*
         & \ResultMethod{SPaRCNet} & \MeanStd{0.250}{0.266} & \MeanStd{0.536}{0.086} & \MeanStd{0.449}{0.126} & \MeanStd{0.362}{0.233} & \SecondMeanStd{0.565}{0.078} & \MeanStd{0.415}{0.153} & \MeanStd{0.394}{0.177} & 0.425 & \Second{0.328} & \BestRightEdge{0.092} \\*
         & \ResultMethod{STT-Trans.} & \MeanStd{0.147}{0.133} & \MeanStd{0.151}{0.136} & \MeanStd{0.126}{0.120} & \MeanStd{0.105}{0.110} & \MeanStd{0.027}{0.082} & \MeanStd{0.207}{0.193} & \MeanStd{0.079}{0.098} & 0.120 & 0.347 & 0.107 \\*
        \addlinespace[1pt]
         & \ResultMethod{NVCode (Ours)} & \MeanStd{0.194}{0.207} & \BestMeanStd{0.613}{0.142} & \SecondMeanStd{0.553}{0.178} & \MeanStd{0.410}{0.400} & \MeanStd{0.557}{0.175} & \SecondMeanStd{0.433}{0.362} & \MeanStd{0.430}{0.373} & \Best{0.456} & \Best{0.327} & 0.096 \\
\end{NVCodeDatasetTable}

\begin{NVCodeDatasetTable}{}{Encoder}{\MultiCLARID\ (Resting State)}{tab:app:encoder-all-datasets:multi-clarid}
\IntraBlock{12} & \ResultMethod{BEIRA} & \MeanStd{0.408}{0.127} & \MeanStd{0.466}{0.131} & \MeanStd{0.470}{0.126} & \BestMeanStd{0.544}{0.112} & \MeanStd{0.523}{0.131} & \MeanStd{0.608}{0.113} & \MeanStd{0.464}{0.114} & 0.498 & 0.442 & 0.316 \\*
         & \ResultMethod{BIOT} & \MeanStd{0.452}{0.089} & \MeanStd{0.406}{0.096} & \MeanStd{0.360}{0.102} & \MeanStd{0.377}{0.103} & \MeanStd{0.414}{0.106} & \MeanStd{0.377}{0.085} & \SecondMeanStd{0.492}{0.100} & 0.411 & 0.406 & 0.312 \\*
         & \ResultMethod{BrainOmni} & \MeanStd{0.428}{0.087} & \MeanStd{0.441}{0.091} & \MeanStd{0.466}{0.090} & \MeanStd{0.371}{0.086} & \MeanStd{0.418}{0.100} & \MeanStd{0.424}{0.093} & \MeanStd{0.423}{0.096} & 0.424 & \Second{0.395} & 0.308 \\*
         & \ResultMethod{CNN-Trans.} & \MeanStd{0.415}{0.126} & \MeanStd{0.472}{0.117} & \MeanStd{0.533}{0.138} & \MeanStd{0.351}{0.082} & \MeanStd{0.552}{0.114} & \MeanStd{0.494}{0.119} & \MeanStd{0.439}{0.115} & 0.465 & 0.421 & 0.346 \\*
         & \ResultMethod{ContraWR} & \MeanStd{0.409}{0.102} & \MeanStd{0.428}{0.120} & \SecondMeanStd{0.591}{0.103} & \MeanStd{0.367}{0.109} & \MeanStd{0.540}{0.106} & \MeanStd{0.512}{0.097} & \MeanStd{0.409}{0.094} & 0.465 & 0.437 & 0.335 \\*
         & \ResultMethod{FFCL} & \MeanStd{0.438}{0.094} & \MeanStd{0.474}{0.110} & \MeanStd{0.460}{0.108} & \MeanStd{0.457}{0.086} & \MeanStd{0.430}{0.113} & \MeanStd{0.468}{0.119} & \MeanStd{0.440}{0.120} & 0.453 & 0.448 & 0.325 \\*
         & \ResultMethod{LaBraM} & \MeanStd{0.492}{0.072} & \MeanStd{0.465}{0.108} & \MeanStd{0.432}{0.090} & \MeanStd{0.426}{0.090} & \MeanStd{0.507}{0.107} & \MeanStd{0.478}{0.113} & \MeanStd{0.410}{0.084} & 0.458 & 0.400 & \SecondRightEdge{0.295} \\*
         & \ResultMethod{SIREN-based} & \SecondMeanStd{0.506}{0.099} & \SecondMeanStd{0.478}{0.141} & \MeanStd{0.578}{0.131} & \MeanStd{0.404}{0.113} & \BestMeanStd{0.651}{0.106} & \SecondMeanStd{0.638}{0.120} & \MeanStd{0.388}{0.106} & \Second{0.520} & 0.424 & 0.310 \\*
         & \ResultMethod{REVE} & \MeanStd{0.330}{0.085} & \MeanStd{0.342}{0.070} & \MeanStd{0.363}{0.068} & \MeanStd{0.337}{0.071} & \MeanStd{0.392}{0.064} & \MeanStd{0.390}{0.076} & \MeanStd{0.395}{0.057} & 0.364 & 0.408 & 0.311 \\*
         & \ResultMethod{SPaRCNet} & \MeanStd{0.428}{0.096} & \MeanStd{0.424}{0.123} & \MeanStd{0.442}{0.124} & \SecondMeanStd{0.485}{0.109} & \SecondMeanStd{0.609}{0.124} & \MeanStd{0.618}{0.124} & \BestMeanStd{0.564}{0.105} & 0.510 & 0.426 & 0.344 \\*
         & \ResultMethod{STT-Trans.} & \MeanStd{0.422}{0.101} & \MeanStd{0.403}{0.109} & \MeanStd{0.396}{0.130} & \MeanStd{0.383}{0.107} & \MeanStd{0.377}{0.105} & \MeanStd{0.234}{0.124} & \MeanStd{0.326}{0.093} & 0.363 & 0.446 & 0.350 \\*
        \addlinespace[1pt]
         & \ResultMethod{NVCode (Ours)} & \BestMeanStd{0.524}{0.100} & \BestMeanStd{0.484}{0.128} & \BestMeanStd{0.639}{0.124} & \MeanStd{0.416}{0.096} & \MeanStd{0.606}{0.119} & \BestMeanStd{0.650}{0.117} & \MeanStd{0.482}{0.124} & \Best{0.543} & \Best{0.384} & \BestRightEdge{0.276} \\
\ResultBlockRule
\InterBlock{12} & \ResultMethod{BEIRA} & \MeanStd{0.163}{0.053} & \BestMeanStd{0.421}{0.058} & \MeanStd{0.394}{0.070} & \MeanStd{0.311}{0.183} & \MeanStd{0.468}{0.045} & \MeanStd{0.403}{0.065} & \MeanStd{0.314}{0.172} & 0.354 & 0.394 & 0.196 \\*
         & \ResultMethod{BIOT} & \MeanStd{0.282}{0.113} & \SecondMeanStd{0.364}{0.063} & \MeanStd{0.381}{0.054} & \MeanStd{0.355}{0.069} & \MeanStd{0.269}{0.092} & \MeanStd{0.302}{0.157} & \MeanStd{0.308}{0.137} & 0.323 & 0.381 & \SecondRightEdge{0.098} \\*
         & \ResultMethod{BrainOmni} & \MeanStd{0.256}{0.079} & \MeanStd{0.170}{0.039} & \MeanStd{0.341}{0.083} & \MeanStd{0.297}{0.155} & \MeanStd{0.290}{0.132} & \MeanStd{0.257}{0.080} & \MeanStd{0.290}{0.134} & 0.271 & 0.385 & 0.106 \\*
         & \ResultMethod{CNN-Trans.} & \MeanStd{0.219}{0.086} & \MeanStd{0.193}{0.071} & \MeanStd{0.308}{0.220} & \MeanStd{0.356}{0.109} & \MeanStd{0.563}{0.034} & \MeanStd{0.525}{0.039} & \SecondMeanStd{0.325}{0.161} & 0.355 & \Best{0.377} & 0.187 \\*
         & \ResultMethod{ContraWR} & \MeanStd{0.162}{0.048} & \MeanStd{0.251}{0.097} & \MeanStd{0.257}{0.104} & \MeanStd{0.344}{0.102} & \MeanStd{0.447}{0.046} & \MeanStd{0.356}{0.089} & \MeanStd{0.313}{0.161} & 0.304 & 0.401 & 0.122 \\*
         & \ResultMethod{FFCL} & \MeanStd{0.215}{0.058} & \MeanStd{0.267}{0.099} & \MeanStd{0.372}{0.065} & \SecondMeanStd{0.376}{0.063} & \MeanStd{0.387}{0.057} & \MeanStd{0.289}{0.141} & \MeanStd{0.291}{0.149} & 0.314 & 0.399 & 0.144 \\*
         & \ResultMethod{LaBraM} & \SecondMeanStd{0.326}{0.100} & \MeanStd{0.285}{0.120} & \MeanStd{0.356}{0.069} & \MeanStd{0.323}{0.106} & \MeanStd{0.467}{0.032} & \MeanStd{0.399}{0.048} & \MeanStd{0.319}{0.113} & 0.353 & 0.382 & 0.179 \\*
         & \ResultMethod{SIREN-based} & \MeanStd{0.237}{0.066} & \MeanStd{0.293}{0.146} & \MeanStd{0.323}{0.108} & \MeanStd{0.313}{0.129} & \MeanStd{0.537}{0.025} & \MeanStd{0.544}{0.024} & \MeanStd{0.282}{0.119} & \Second{0.361} & \Second{0.380} & 0.173 \\*
         & \ResultMethod{REVE} & \MeanStd{0.302}{0.115} & \MeanStd{0.337}{0.063} & \MeanStd{0.319}{0.081} & \MeanStd{0.306}{0.106} & \MeanStd{0.404}{0.034} & \MeanStd{0.346}{0.056} & \MeanStd{0.308}{0.101} & 0.332 & 0.382 & \BestRightEdge{0.094} \\*
         & \ResultMethod{SPaRCNet} & \MeanStd{0.178}{0.073} & \MeanStd{0.153}{0.063} & \MeanStd{0.293}{0.250} & \MeanStd{0.224}{0.102} & \SecondMeanStd{0.610}{0.034} & \SecondMeanStd{0.547}{0.041} & \MeanStd{0.249}{0.131} & 0.322 & 0.384 & 0.147 \\*
         & \ResultMethod{STT-Trans.} & \BestMeanStd{0.398}{0.047} & \MeanStd{0.361}{0.064} & \SecondMeanStd{0.408}{0.044} & \MeanStd{0.078}{0.025} & \MeanStd{0.306}{0.141} & \MeanStd{0.353}{0.070} & \MeanStd{0.313}{0.122} & 0.317 & 0.406 & 0.137 \\*
        \addlinespace[1pt]
         & \ResultMethod{NVCode (Ours)} & \MeanStd{0.324}{0.134} & \MeanStd{0.361}{0.197} & \BestMeanStd{0.428}{0.229} & \BestMeanStd{0.395}{0.349} & \BestMeanStd{0.768}{0.038} & \BestMeanStd{0.742}{0.041} & \BestMeanStd{0.436}{0.206} & \Best{0.493} & \Second{0.380} & 0.105 \\
\end{NVCodeDatasetTable}

\begin{NVCodeDatasetTable}{}{Encoder}{\GradCPT\ (Resting State)}{tab:app:encoder-all-datasets:gradcpt}
\IntraBlock{12} & \ResultMethod{BEIRA} & \BestMeanStd{0.476}{0.173} & \MeanStd{0.316}{0.199} & \MeanStd{0.322}{0.191} & \MeanStd{0.380}{0.139} & \MeanStd{0.431}{0.155} & \MeanStd{0.296}{0.183} & \MeanStd{0.439}{0.172} & 0.380 & 0.383 & 0.359 \\*
         & \ResultMethod{BIOT} & \MeanStd{0.345}{0.142} & \MeanStd{0.249}{0.148} & \MeanStd{0.208}{0.132} & \MeanStd{0.289}{0.134} & \MeanStd{0.279}{0.124} & \MeanStd{0.375}{0.141} & \MeanStd{0.315}{0.110} & 0.294 & 0.332 & \SecondRightEdge{0.326} \\*
         & \ResultMethod{BrainOmni} & \MeanStd{0.214}{0.118} & \MeanStd{0.346}{0.117} & \MeanStd{0.388}{0.146} & \MeanStd{0.313}{0.152} & \MeanStd{0.333}{0.152} & \MeanStd{0.309}{0.126} & \MeanStd{0.290}{0.158} & 0.313 & 0.333 & 0.332 \\*
         & \ResultMethod{CNN-Trans.} & \MeanStd{0.332}{0.126} & \MeanStd{0.363}{0.140} & \SecondMeanStd{0.446}{0.129} & \MeanStd{0.409}{0.153} & \MeanStd{0.464}{0.151} & \MeanStd{0.344}{0.149} & \MeanStd{0.409}{0.166} & 0.395 & \Second{0.330} & 0.336 \\*
         & \ResultMethod{ContraWR} & \MeanStd{0.347}{0.163} & \SecondMeanStd{0.447}{0.164} & \MeanStd{0.399}{0.144} & \SecondMeanStd{0.478}{0.181} & \SecondMeanStd{0.545}{0.157} & \MeanStd{0.400}{0.168} & \SecondMeanStd{0.509}{0.161} & \Second{0.447} & 0.352 & 0.353 \\*
         & \ResultMethod{FFCL} & \MeanStd{0.344}{0.163} & \MeanStd{0.378}{0.165} & \MeanStd{0.347}{0.169} & \MeanStd{0.446}{0.161} & \MeanStd{0.382}{0.151} & \MeanStd{0.393}{0.146} & \MeanStd{0.365}{0.159} & 0.379 & 0.359 & 0.368 \\*
         & \ResultMethod{LaBraM} & \SecondMeanStd{0.417}{0.130} & \MeanStd{0.384}{0.161} & \MeanStd{0.388}{0.146} & \MeanStd{0.245}{0.093} & \MeanStd{0.336}{0.152} & \MeanStd{0.294}{0.140} & \MeanStd{0.325}{0.138} & 0.341 & 0.333 & 0.326 \\*
         & \ResultMethod{SIREN-based} & \MeanStd{0.383}{0.165} & \MeanStd{0.446}{0.158} & \MeanStd{0.393}{0.142} & \BestMeanStd{0.483}{0.137} & \MeanStd{0.439}{0.148} & \SecondMeanStd{0.448}{0.149} & \MeanStd{0.441}{0.155} & 0.433 & \Best{0.329} & 0.354 \\*
         & \ResultMethod{REVE} & \MeanStd{0.337}{0.136} & \MeanStd{0.410}{0.144} & \MeanStd{0.313}{0.135} & \MeanStd{0.281}{0.131} & \MeanStd{0.331}{0.132} & \MeanStd{0.377}{0.135} & \MeanStd{0.287}{0.137} & 0.334 & 0.335 & 0.357 \\*
         & \ResultMethod{SPaRCNet} & \MeanStd{0.311}{0.143} & \MeanStd{0.412}{0.150} & \BestMeanStd{0.493}{0.168} & \MeanStd{0.265}{0.116} & \MeanStd{0.411}{0.153} & \MeanStd{0.361}{0.149} & \MeanStd{0.400}{0.145} & 0.379 & 0.345 & 0.372 \\*
         & \ResultMethod{STT-Trans.} & \MeanStd{0.395}{0.140} & \MeanStd{0.411}{0.128} & \MeanStd{0.393}{0.126} & \MeanStd{0.303}{0.146} & \MeanStd{0.308}{0.127} & \MeanStd{0.359}{0.154} & \MeanStd{0.311}{0.150} & 0.354 & 0.337 & 0.346 \\*
        \addlinespace[1pt]
         & \ResultMethod{NVCode (Ours)} & \MeanStd{0.392}{0.200} & \BestMeanStd{0.601}{0.181} & \MeanStd{0.424}{0.144} & \MeanStd{0.463}{0.180} & \BestMeanStd{0.577}{0.188} & \BestMeanStd{0.538}{0.174} & \BestMeanStd{0.609}{0.166} & \Best{0.515} & \Best{0.329} & \BestRightEdge{0.309} \\
\ResultBlockRule
\InterBlock{12} & \ResultMethod{BEIRA} & \MeanStd{0.238}{0.133} & \MeanStd{0.412}{0.089} & \MeanStd{0.221}{0.114} & \MeanStd{0.359}{0.137} & \MeanStd{0.250}{0.151} & \MeanStd{0.256}{0.162} & \MeanStd{0.302}{0.324} & 0.291 & 0.318 & 0.149 \\*
         & \ResultMethod{BIOT} & \MeanStd{0.182}{0.116} & \MeanStd{0.183}{0.117} & \MeanStd{0.177}{0.113} & \MeanStd{0.274}{0.277} & \MeanStd{0.267}{0.252} & \MeanStd{0.309}{0.376} & \MeanStd{0.353}{0.198} & 0.249 & 0.321 & 0.104 \\*
         & \ResultMethod{BrainOmni} & \MeanStd{0.210}{0.124} & \MeanStd{0.287}{0.306} & \MeanStd{0.204}{0.119} & \MeanStd{0.249}{0.177} & \MeanStd{0.272}{0.235} & \MeanStd{0.362}{0.157} & \MeanStd{0.361}{0.160} & 0.278 & 0.318 & 0.098 \\*
         & \ResultMethod{CNN-Trans.} & \MeanStd{0.275}{0.169} & \MeanStd{0.230}{0.100} & \MeanStd{0.245}{0.115} & \MeanStd{0.268}{0.154} & \MeanStd{0.351}{0.121} & \MeanStd{0.301}{0.271} & \MeanStd{0.214}{0.087} & 0.269 & 0.321 & \SecondRightEdge{0.064} \\*
         & \ResultMethod{ContraWR} & \MeanStd{0.175}{0.106} & \MeanStd{0.222}{0.148} & \MeanStd{0.431}{0.102} & \MeanStd{0.217}{0.142} & \MeanStd{0.290}{0.349} & \MeanStd{0.174}{0.105} & \MeanStd{0.176}{0.106} & 0.241 & 0.325 & 0.210 \\*
         & \ResultMethod{FFCL} & \MeanStd{0.245}{0.187} & \MeanStd{0.268}{0.248} & \MeanStd{0.299}{0.400} & \MeanStd{0.051}{0.061} & \MeanStd{0.058}{0.063} & \MeanStd{0.118}{0.080} & \MeanStd{0.158}{0.098} & 0.171 & 0.328 & 0.112 \\*
         & \ResultMethod{LaBraM} & \MeanStd{0.243}{0.114} & \MeanStd{0.266}{0.151} & \MeanStd{0.313}{0.209} & \MeanStd{0.149}{0.058} & \MeanStd{0.301}{0.269} & \MeanStd{0.280}{0.184} & \MeanStd{0.200}{0.079} & 0.250 & 0.319 & 0.085 \\*
         & \ResultMethod{SIREN-based} & \MeanStd{0.365}{0.342} & \MeanStd{0.546}{0.126} & \BestMeanStd{0.505}{0.147} & \SecondMeanStd{0.378}{0.306} & \MeanStd{0.469}{0.173} & \MeanStd{0.380}{0.301} & \SecondMeanStd{0.554}{0.123} & \Second{0.457} & \Best{0.311} & 0.131 \\*
         & \ResultMethod{REVE} & \MeanStd{0.205}{0.122} & \MeanStd{0.227}{0.145} & \MeanStd{0.195}{0.113} & \MeanStd{0.116}{0.073} & \MeanStd{0.189}{0.109} & \MeanStd{0.234}{0.155} & \MeanStd{0.305}{0.367} & 0.210 & 0.323 & 0.145 \\*
         & \ResultMethod{SPaRCNet} & \SecondMeanStd{0.485}{0.142} & \BestMeanStd{0.579}{0.100} & \MeanStd{0.355}{0.336} & \MeanStd{0.268}{0.400} & \BestMeanStd{0.516}{0.125} & \BestMeanStd{0.419}{0.201} & \MeanStd{0.498}{0.134} & 0.446 & \Second{0.313} & 0.137 \\*
         & \ResultMethod{STT-Trans.} & \MeanStd{0.248}{0.158} & \MeanStd{0.257}{0.176} & \MeanStd{0.222}{0.123} & \BestMeanStd{0.463}{0.072} & \MeanStd{0.258}{0.178} & \MeanStd{0.263}{0.190} & \MeanStd{0.282}{0.252} & 0.285 & 0.319 & 0.153 \\*
        \addlinespace[1pt]
         & \ResultMethod{NVCode (Ours)} & \BestMeanStd{0.576}{0.219} & \SecondMeanStd{0.560}{0.236} & \SecondMeanStd{0.482}{0.388} & \MeanStd{0.222}{0.217} & \SecondMeanStd{0.509}{0.317} & \SecondMeanStd{0.412}{0.400} & \BestMeanStd{0.587}{0.208} & \Best{0.478} & 0.314 & \BestRightEdge{0.044} \\
\end{NVCodeDatasetTable}

\begin{NVCodeDatasetTable}{}{Encoder}{NeuroBOLT (Resting State)}{tab:app:encoder-all-datasets:neurobolt}
\IntraBlock{12} & \ResultMethod{BEIRA} & \MeanStd{0.447}{0.105} & \MeanStd{0.453}{0.107} & \MeanStd{0.356}{0.126} & \MeanStd{0.387}{0.102} & \MeanStd{0.303}{0.102} & \MeanStd{0.322}{0.102} & \MeanStd{0.490}{0.113} & 0.394 & 0.317 & 0.229 \\*
         & \ResultMethod{BIOT} & \MeanStd{0.212}{0.066} & \MeanStd{0.200}{0.062} & \MeanStd{0.219}{0.063} & \MeanStd{0.208}{0.070} & \MeanStd{0.144}{0.059} & \MeanStd{0.203}{0.069} & \MeanStd{0.210}{0.072} & 0.199 & 0.305 & 0.182 \\*
         & \ResultMethod{BrainOmni} & \MeanStd{0.335}{0.088} & \MeanStd{0.326}{0.087} & \MeanStd{0.273}{0.103} & \MeanStd{0.323}{0.098} & \MeanStd{0.225}{0.074} & \MeanStd{0.208}{0.082} & \MeanStd{0.393}{0.100} & 0.297 & 0.301 & 0.249 \\*
         & \ResultMethod{CNN-Trans.} & \MeanStd{0.411}{0.108} & \MeanStd{0.400}{0.096} & \MeanStd{0.326}{0.112} & \MeanStd{0.304}{0.104} & \MeanStd{0.292}{0.075} & \MeanStd{0.274}{0.090} & \MeanStd{0.400}{0.110} & 0.344 & 0.314 & 0.224 \\*
         & \ResultMethod{ContraWR} & \MeanStd{0.386}{0.106} & \MeanStd{0.338}{0.112} & \MeanStd{0.256}{0.092} & \MeanStd{0.364}{0.103} & \MeanStd{0.237}{0.094} & \MeanStd{0.218}{0.084} & \MeanStd{0.402}{0.109} & 0.314 & 0.347 & 0.209 \\*
         & \ResultMethod{FFCL} & \MeanStd{0.447}{0.105} & \MeanStd{0.377}{0.110} & \MeanStd{0.290}{0.103} & \MeanStd{0.353}{0.112} & \MeanStd{0.266}{0.096} & \MeanStd{0.287}{0.093} & \MeanStd{0.440}{0.115} & 0.351 & 0.328 & 0.191 \\*
         & \ResultMethod{LaBraM} & \MeanStd{0.463}{0.113} & \MeanStd{0.452}{0.107} & \MeanStd{0.375}{0.101} & \MeanStd{0.389}{0.121} & \MeanStd{0.343}{0.076} & \MeanStd{0.371}{0.091} & \MeanStd{0.503}{0.096} & 0.414 & \Second{0.289} & 0.188 \\*
         & \ResultMethod{SIREN-based} & \MeanStd{0.527}{0.096} & \MeanStd{0.469}{0.105} & \SecondMeanStd{0.427}{0.080} & \MeanStd{0.436}{0.109} & \MeanStd{0.348}{0.092} & \MeanStd{0.435}{0.097} & \MeanStd{0.562}{0.093} & 0.458 & 0.300 & 0.188 \\*
         & \ResultMethod{REVE} & \MeanStd{0.174}{0.063} & \MeanStd{0.149}{0.062} & \MeanStd{0.103}{0.075} & \MeanStd{0.161}{0.067} & \MeanStd{0.085}{0.065} & \MeanStd{0.058}{0.059} & \MeanStd{0.149}{0.066} & 0.125 & 0.310 & 0.259 \\*
         & \ResultMethod{SPaRCNet} & \SecondMeanStd{0.531}{0.099} & \SecondMeanStd{0.525}{0.096} & \MeanStd{0.379}{0.109} & \SecondMeanStd{0.489}{0.109} & \SecondMeanStd{0.405}{0.104} & \SecondMeanStd{0.457}{0.080} & \SecondMeanStd{0.647}{0.081} & \Second{0.491} & 0.312 & \BestRightEdge{0.162} \\*
         & \ResultMethod{STT-Trans.} & \MeanStd{0.104}{0.066} & \MeanStd{0.203}{0.070} & \MeanStd{0.180}{0.066} & \MeanStd{0.136}{0.068} & \MeanStd{0.128}{0.070} & \MeanStd{0.106}{0.082} & \MeanStd{0.178}{0.062} & 0.148 & 0.326 & 0.231 \\*
        \addlinespace[1pt]
         & \ResultMethod{NVCode (Ours)} & \BestMeanStd{0.639}{0.113} & \BestMeanStd{0.671}{0.107} & \BestMeanStd{0.489}{0.153} & \BestMeanStd{0.564}{0.130} & \BestMeanStd{0.439}{0.175} & \BestMeanStd{0.515}{0.145} & \BestMeanStd{0.653}{0.110} & \Best{0.567} & \Best{0.219} & \SecondRightEdge{0.167} \\
\ResultBlockRule
\InterBlock{12} & \ResultMethod{BEIRA} & \MeanStd{0.379}{0.052} & \MeanStd{0.380}{0.052} & \MeanStd{0.282}{0.071} & \MeanStd{0.343}{0.058} & \MeanStd{0.126}{0.176} & \MeanStd{0.142}{0.153} & \MeanStd{0.389}{0.051} & 0.291 & 0.290 & 0.183 \\*
         & \ResultMethod{BIOT} & \MeanStd{0.284}{0.081} & \MeanStd{0.303}{0.076} & \MeanStd{0.224}{0.105} & \MeanStd{0.261}{0.088} & \MeanStd{0.178}{0.135} & \MeanStd{0.195}{0.122} & \MeanStd{0.318}{0.072} & 0.252 & 0.294 & 0.152 \\*
         & \ResultMethod{BrainOmni} & \MeanStd{0.291}{0.067} & \MeanStd{0.297}{0.065} & \MeanStd{0.228}{0.087} & \MeanStd{0.264}{0.074} & \MeanStd{0.160}{0.129} & \MeanStd{0.204}{0.099} & \MeanStd{0.346}{0.056} & 0.256 & 0.293 & 0.109 \\*
         & \ResultMethod{CNN-Trans.} & \MeanStd{0.405}{0.077} & \MeanStd{0.487}{0.063} & \MeanStd{0.360}{0.087} & \MeanStd{0.424}{0.073} & \MeanStd{0.263}{0.122} & \MeanStd{0.304}{0.104} & \MeanStd{0.495}{0.062} & 0.391 & 0.284 & 0.098 \\*
         & \ResultMethod{ContraWR} & \MeanStd{0.293}{0.076} & \MeanStd{0.270}{0.082} & \MeanStd{0.301}{0.073} & \MeanStd{0.297}{0.074} & \MeanStd{0.166}{0.141} & \MeanStd{0.209}{0.109} & \MeanStd{0.353}{0.062} & 0.270 & 0.302 & 0.102 \\*
         & \ResultMethod{FFCL} & \MeanStd{0.404}{0.065} & \MeanStd{0.394}{0.067} & \MeanStd{0.301}{0.089} & \MeanStd{0.314}{0.085} & \MeanStd{0.231}{0.118} & \MeanStd{0.252}{0.108} & \MeanStd{0.415}{0.063} & 0.330 & 0.290 & 0.101 \\*
         & \ResultMethod{LaBraM} & \MeanStd{0.445}{0.064} & \MeanStd{0.447}{0.064} & \MeanStd{0.356}{0.082} & \MeanStd{0.421}{0.068} & \MeanStd{0.200}{0.152} & \MeanStd{0.352}{0.082} & \MeanStd{0.482}{0.059} & 0.386 & 0.285 & \SecondRightEdge{0.065} \\*
         & \ResultMethod{SIREN-based} & \MeanStd{0.475}{0.076} & \SecondMeanStd{0.521}{0.069} & \MeanStd{0.414}{0.088} & \MeanStd{0.420}{0.087} & \SecondMeanStd{0.318}{0.116} & \MeanStd{0.401}{0.091} & \MeanStd{0.546}{0.066} & 0.442 & 0.279 & 0.112 \\*
         & \ResultMethod{REVE} & \MeanStd{0.154}{0.041} & \MeanStd{0.059}{0.133} & \MeanStd{0.081}{0.089} & \MeanStd{0.138}{0.046} & \MeanStd{0.031}{0.079} & \MeanStd{0.105}{0.064} & \MeanStd{0.168}{0.037} & 0.105 & 0.303 & 0.153 \\*
         & \ResultMethod{SPaRCNet} & \SecondMeanStd{0.489}{0.064} & \MeanStd{0.515}{0.061} & \SecondMeanStd{0.435}{0.073} & \SecondMeanStd{0.506}{0.062} & \MeanStd{0.252}{0.131} & \SecondMeanStd{0.438}{0.072} & \SecondMeanStd{0.554}{0.057} & \Second{0.455} & \Second{0.275} & 0.104 \\*
         & \ResultMethod{STT-Trans.} & \MeanStd{0.090}{0.082} & \MeanStd{0.099}{0.073} & \MeanStd{0.080}{0.096} & \MeanStd{0.076}{0.103} & \MeanStd{0.093}{0.079} & \MeanStd{0.088}{0.085} & \MeanStd{0.131}{0.052} & 0.094 & 0.303 & 0.133 \\*
        \addlinespace[1pt]
         & \ResultMethod{NVCode (Ours)} & \BestMeanStd{0.616}{0.049} & \BestMeanStd{0.614}{0.050} & \BestMeanStd{0.436}{0.070} & \BestMeanStd{0.523}{0.058} & \BestMeanStd{0.341}{0.090} & \BestMeanStd{0.468}{0.065} & \BestMeanStd{0.585}{0.052} & \Best{0.512} & \Best{0.268} & \BestRightEdge{0.043} \\
\end{NVCodeDatasetTable}

\begin{NVCodeDatasetTable}{}{Encoder}{\XPone\ (Active)}{tab:app:encoder-all-datasets:xp1}
\IntraBlock{12} & \ResultMethod{BEIRA} & \MeanOnly{0.467}{0.310} & \MeanOnly{0.233}{0.324} & \MeanOnly{0.216}{0.332} & \BestMeanOnly{0.616}{0.321} & \MeanOnly{0.379}{0.289} & \SecondMeanOnly{0.376}{0.287} & \MeanOnly{0.502}{0.329} & 0.398 & 0.539 & 0.390 \\*
         & \ResultMethod{BIOT} & \MeanOnly{0.009}{0.172} & \MeanOnly{-0.054}{0.209} & \MeanOnly{0.228}{0.204} & \MeanOnly{0.246}{0.210} & \MeanOnly{0.086}{0.228} & \MeanOnly{0.086}{0.297} & \MeanOnly{0.098}{0.212} & 0.100 & 0.515 & 0.413 \\*
         & \ResultMethod{BrainOmni} & \MeanOnly{0.142}{0.262} & \MeanOnly{0.347}{0.223} & \MeanOnly{0.223}{0.233} & \MeanOnly{0.373}{0.247} & \MeanOnly{0.339}{0.277} & \MeanOnly{0.064}{0.254} & \MeanOnly{0.273}{0.282} & 0.252 & 0.502 & 0.400 \\*
         & \ResultMethod{CNN-Trans.} & \MeanOnly{0.356}{0.350} & \SecondMeanOnly{0.447}{0.268} & \MeanOnly{0.208}{0.351} & \MeanOnly{0.184}{0.306} & \MeanOnly{0.240}{0.257} & \MeanOnly{0.212}{0.272} & \MeanOnly{0.198}{0.358} & 0.263 & \Second{0.493} & 0.408 \\*
         & \ResultMethod{ContraWR} & \MeanOnly{0.128}{0.203} & \MeanOnly{0.171}{0.237} & \SecondMeanOnly{0.375}{0.310} & \MeanOnly{0.121}{0.379} & \MeanOnly{0.064}{0.280} & \MeanOnly{0.119}{0.318} & \MeanOnly{-0.010}{0.234} & 0.138 & 0.547 & 0.431 \\*
         & \ResultMethod{FFCL} & \MeanOnly{0.384}{0.216} & \MeanOnly{-0.049}{0.344} & \MeanOnly{0.292}{0.412} & \MeanOnly{0.227}{0.395} & \SecondMeanOnly{0.410}{0.234} & \MeanOnly{0.269}{0.281} & \MeanOnly{0.232}{0.373} & 0.252 & 0.580 & 0.446 \\*
         & \ResultMethod{LaBraM} & \MeanOnly{0.239}{0.214} & \MeanOnly{0.004}{0.272} & \MeanOnly{0.143}{0.218} & \MeanOnly{0.240}{0.316} & \MeanOnly{0.277}{0.230} & \MeanOnly{0.179}{0.161} & \MeanOnly{0.197}{0.256} & 0.183 & 0.506 & 0.412 \\*
         & \ResultMethod{SIREN-based} & \SecondMeanOnly{0.473}{0.243} & \MeanOnly{0.418}{0.285} & \BestMeanOnly{0.471}{0.316} & \MeanOnly{0.236}{0.419} & \MeanOnly{0.394}{0.341} & \BestMeanOnly{0.408}{0.275} & \BestMeanOnly{0.537}{0.246} & \Second{0.420} & 0.512 & \SecondRightEdge{0.388} \\*
         & \ResultMethod{REVE} & \MeanOnly{-0.097}{0.196} & \MeanOnly{0.133}{0.278} & \MeanOnly{0.011}{0.279} & \MeanOnly{0.196}{0.273} & \MeanOnly{0.218}{0.246} & \MeanOnly{0.233}{0.228} & \MeanOnly{0.286}{0.277} & 0.140 & 0.511 & 0.397 \\*
         & \ResultMethod{SPaRCNet} & \MeanOnly{0.364}{0.254} & \MeanOnly{0.250}{0.284} & \MeanOnly{0.311}{0.351} & \MeanOnly{0.067}{0.321} & \MeanOnly{0.372}{0.342} & \MeanOnly{0.329}{0.273} & \MeanOnly{0.188}{0.266} & 0.269 & 0.531 & 0.419 \\*
         & \ResultMethod{STT-Trans.} & \MeanOnly{0.263}{0.219} & \MeanOnly{0.387}{0.292} & \MeanOnly{0.226}{0.220} & \MeanOnly{0.115}{0.162} & \MeanOnly{0.213}{0.238} & \MeanOnly{0.221}{0.237} & \MeanOnly{0.285}{0.219} & 0.244 & 0.512 & 0.405 \\*
        \addlinespace[1pt]
         & \ResultMethod{NVCode (Ours)} & \BestMeanOnly{0.707}{0.293} & \BestMeanOnly{0.495}{0.329} & \MeanOnly{0.291}{0.315} & \SecondMeanOnly{0.461}{0.378} & \BestMeanOnly{0.608}{0.342} & \MeanOnly{0.328}{0.357} & \SecondMeanOnly{0.536}{0.450} & \Best{0.489} & \Best{0.432} & \BestRightEdge{0.361} \\
\ResultBlockRule
\InterBlock{12} & \ResultMethod{BEIRA} & \MeanOnly{0.178}{0.310} & \MeanOnly{-0.076}{0.121} & \MeanOnly{-0.129}{0.107} & \MeanOnly{0.024}{0.160} & \MeanOnly{-0.062}{0.125} & \MeanOnly{-0.141}{0.105} & \MeanOnly{0.142}{0.254} & -0.009 & 0.513 & \SecondRightEdge{0.269} \\*
         & \ResultMethod{BIOT} & \MeanOnly{0.381}{0.305} & \MeanOnly{0.109}{0.160} & \MeanOnly{0.236}{0.355} & \MeanOnly{0.195}{0.254} & \BestMeanOnly{0.470}{0.175} & \BestMeanOnly{0.371}{0.333} & \MeanOnly{0.194}{0.253} & 0.279 & 0.483 & 0.272 \\*
         & \ResultMethod{BrainOmni} & \MeanOnly{0.438}{0.159} & \MeanOnly{0.224}{0.246} & \MeanOnly{0.225}{0.248} & \MeanOnly{0.365}{0.270} & \MeanOnly{0.228}{0.254} & \MeanOnly{0.348}{0.321} & \MeanOnly{0.337}{0.369} & 0.310 & 0.490 & 0.314 \\*
         & \ResultMethod{CNN-Trans.} & \MeanOnly{0.254}{0.347} & \MeanOnly{0.242}{0.305} & \SecondMeanOnly{0.276}{0.400} & \MeanOnly{0.208}{0.225} & \SecondMeanOnly{0.459}{0.150} & \SecondMeanOnly{0.354}{0.316} & \MeanOnly{0.070}{0.110} & 0.266 & 0.489 & 0.348 \\*
         & \ResultMethod{ContraWR} & \MeanOnly{0.497}{0.127} & \MeanOnly{0.136}{0.148} & \MeanOnly{0.259}{0.371} & \MeanOnly{0.352}{0.330} & \MeanOnly{0.196}{0.209} & \MeanOnly{0.233}{0.282} & \MeanOnly{0.186}{0.196} & 0.265 & 0.490 & 0.384 \\*
         & \ResultMethod{FFCL} & \MeanOnly{0.485}{0.092} & \MeanOnly{0.143}{0.105} & \MeanOnly{0.211}{0.160} & \MeanOnly{0.353}{0.223} & \MeanOnly{0.307}{0.400} & \MeanOnly{0.313}{0.387} & \MeanOnly{0.125}{0.096} & 0.277 & 0.493 & 0.356 \\*
         & \ResultMethod{LaBraM} & \MeanOnly{0.428}{0.171} & \MeanOnly{0.387}{0.227} & \MeanOnly{0.202}{0.208} & \MeanOnly{0.507}{0.116} & \MeanOnly{0.280}{0.400} & \MeanOnly{0.337}{0.372} & \MeanOnly{0.307}{0.400} & 0.350 & \Second{0.480} & 0.282 \\*
         & \ResultMethod{SIREN-based} & \MeanOnly{0.495}{0.145} & \MeanOnly{0.276}{0.400} & \MeanOnly{0.135}{0.166} & \SecondMeanOnly{0.578}{0.107} & \MeanOnly{0.406}{0.234} & \MeanOnly{0.266}{0.400} & \MeanOnly{0.338}{0.400} & 0.356 & 0.492 & 0.343 \\*
         & \ResultMethod{REVE} & \MeanOnly{0.367}{0.364} & \MeanOnly{0.233}{0.364} & \MeanOnly{0.138}{0.192} & \MeanOnly{0.520}{0.150} & \MeanOnly{0.232}{0.360} & \MeanOnly{0.053}{0.136} & \SecondMeanOnly{0.504}{0.160} & 0.292 & 0.497 & 0.308 \\*
         & \ResultMethod{SPaRCNet} & \SecondMeanOnly{0.587}{0.154} & \SecondMeanOnly{0.481}{0.228} & \MeanOnly{0.259}{0.400} & \MeanOnly{0.465}{0.245} & \MeanOnly{0.432}{0.293} & \MeanOnly{0.313}{0.400} & \MeanOnly{0.435}{0.288} & \Second{0.425} & 0.499 & 0.318 \\*
         & \ResultMethod{STT-Trans.} & \MeanOnly{0.353}{0.299} & \MeanOnly{0.225}{0.242} & \MeanOnly{0.198}{0.196} & \MeanOnly{0.099}{0.115} & \MeanOnly{0.310}{0.400} & \MeanOnly{0.100}{0.116} & \MeanOnly{0.178}{0.171} & 0.209 & 0.489 & \BestRightEdge{0.265} \\*
        \addlinespace[1pt]
         & \ResultMethod{NVCode (Ours)} & \BestMeanOnly{0.654}{0.213} & \BestMeanOnly{0.660}{0.209} & \BestMeanOnly{0.342}{0.400} & \BestMeanOnly{0.588}{0.260} & \MeanOnly{-0.012}{0.203} & \MeanOnly{0.182}{0.400} & \BestMeanOnly{0.660}{0.209} & \Best{0.439} & \Best{0.466} & 0.310 \\
\end{NVCodeDatasetTable}

\begin{NVCodeDatasetTable}{}{Encoder}{\AffectiveMusic\ (Active)}{tab:app:encoder-all-datasets:affective-music}
\IntraBlock{12} & \ResultMethod{BEIRA} & \MeanStd{0.279}{0.118} & \MeanStd{0.413}{0.137} & \MeanStd{0.477}{0.126} & \MeanStd{0.317}{0.107} & \SecondMeanStd{0.469}{0.103} & \MeanStd{0.455}{0.108} & \MeanStd{0.273}{0.112} & 0.383 & \Second{0.357} & 0.283 \\*
         & \ResultMethod{BIOT} & \MeanStd{0.194}{0.083} & \MeanStd{0.235}{0.065} & \MeanStd{0.252}{0.086} & \MeanStd{0.218}{0.050} & \MeanStd{0.198}{0.055} & \MeanStd{0.251}{0.081} & \MeanStd{0.258}{0.074} & 0.230 & 0.357 & 0.261 \\*
         & \ResultMethod{BrainOmni} & \MeanStd{0.276}{0.104} & \MeanStd{0.320}{0.085} & \MeanStd{0.393}{0.099} & \MeanStd{0.293}{0.086} & \MeanStd{0.297}{0.067} & \MeanStd{0.354}{0.063} & \MeanStd{0.261}{0.093} & 0.313 & \Best{0.345} & 0.248 \\*
         & \ResultMethod{CNN-Trans.} & \MeanStd{0.324}{0.083} & \MeanStd{0.396}{0.092} & \MeanStd{0.419}{0.147} & \MeanStd{0.330}{0.064} & \MeanStd{0.393}{0.133} & \MeanStd{0.449}{0.103} & \MeanStd{0.290}{0.088} & 0.371 & 0.369 & 0.239 \\*
         & \ResultMethod{ContraWR} & \MeanStd{0.314}{0.123} & \MeanStd{0.367}{0.091} & \BestMeanStd{0.553}{0.116} & \MeanStd{0.314}{0.111} & \MeanStd{0.410}{0.117} & \MeanStd{0.481}{0.105} & \MeanStd{0.265}{0.092} & 0.386 & 0.396 & 0.248 \\*
         & \ResultMethod{FFCL} & \MeanStd{0.309}{0.117} & \MeanStd{0.360}{0.094} & \MeanStd{0.476}{0.108} & \MeanStd{0.255}{0.092} & \MeanStd{0.396}{0.122} & \SecondMeanStd{0.486}{0.074} & \MeanStd{0.278}{0.079} & 0.366 & 0.405 & 0.245 \\*
         & \ResultMethod{LaBraM} & \MeanStd{0.250}{0.074} & \MeanStd{0.270}{0.097} & \MeanStd{0.299}{0.121} & \MeanStd{0.172}{0.088} & \MeanStd{0.282}{0.089} & \MeanStd{0.367}{0.093} & \MeanStd{0.210}{0.072} & 0.264 & 0.360 & 0.243 \\*
         & \ResultMethod{SIREN-based} & \MeanStd{0.311}{0.097} & \SecondMeanStd{0.463}{0.099} & \MeanStd{0.446}{0.088} & \MeanStd{0.294}{0.080} & \MeanStd{0.450}{0.112} & \MeanStd{0.469}{0.110} & \SecondMeanStd{0.311}{0.074} & \Second{0.392} & 0.386 & 0.241 \\*
         & \ResultMethod{REVE} & \MeanStd{0.202}{0.049} & \MeanStd{0.259}{0.051} & \MeanStd{0.245}{0.059} & \MeanStd{0.213}{0.047} & \MeanStd{0.274}{0.055} & \MeanStd{0.289}{0.056} & \MeanStd{0.216}{0.055} & 0.242 & 0.360 & 0.246 \\*
         & \ResultMethod{SPaRCNet} & \SecondMeanStd{0.324}{0.132} & \MeanStd{0.426}{0.099} & \MeanStd{0.357}{0.140} & \SecondMeanStd{0.336}{0.095} & \MeanStd{0.376}{0.145} & \MeanStd{0.387}{0.112} & \MeanStd{0.269}{0.088} & 0.353 & 0.382 & 0.261 \\*
         & \ResultMethod{STT-Trans.} & \MeanStd{0.207}{0.104} & \MeanStd{0.252}{0.077} & \MeanStd{0.316}{0.110} & \MeanStd{0.225}{0.071} & \MeanStd{0.170}{0.094} & \MeanStd{0.215}{0.085} & \MeanStd{0.204}{0.066} & 0.227 & 0.423 & \SecondRightEdge{0.229} \\*
        \addlinespace[1pt]
         & \ResultMethod{NVCode (Ours)} & \BestMeanStd{0.405}{0.084} & \BestMeanStd{0.542}{0.100} & \SecondMeanStd{0.496}{0.113} & \BestMeanStd{0.442}{0.069} & \BestMeanStd{0.491}{0.098} & \BestMeanStd{0.543}{0.120} & \BestMeanStd{0.381}{0.121} & \Best{0.471} & 0.395 & \BestRightEdge{0.226} \\
\ResultBlockRule
\InterBlock{12} & \ResultMethod{BEIRA} & \MeanStd{0.328}{0.056} & \MeanStd{0.341}{0.052} & \MeanStd{0.434}{0.034} & \MeanStd{0.337}{0.053} & \MeanStd{0.285}{0.075} & \MeanStd{0.412}{0.037} & \MeanStd{0.219}{0.158} & 0.337 & 0.344 & 0.158 \\*
         & \ResultMethod{BIOT} & \MeanStd{0.229}{0.056} & \MeanStd{0.289}{0.030} & \MeanStd{0.190}{0.077} & \MeanStd{0.263}{0.037} & \MeanStd{0.227}{0.057} & \MeanStd{0.196}{0.087} & \MeanStd{0.150}{0.043} & 0.220 & 0.339 & 0.160 \\*
         & \ResultMethod{BrainOmni} & \MeanStd{0.253}{0.085} & \MeanStd{0.263}{0.076} & \MeanStd{0.452}{0.027} & \MeanStd{0.439}{0.028} & \MeanStd{0.323}{0.048} & \MeanStd{0.361}{0.039} & \MeanStd{0.283}{0.064} & 0.339 & \Best{0.328} & 0.157 \\*
         & \ResultMethod{CNN-Trans.} & \MeanStd{0.207}{0.079} & \MeanStd{0.237}{0.048} & \MeanStd{0.183}{0.066} & \MeanStd{0.109}{0.029} & \MeanStd{0.163}{0.048} & \MeanStd{0.186}{0.070} & \MeanStd{0.217}{0.065} & 0.186 & 0.335 & \BestRightEdge{0.112} \\*
         & \ResultMethod{ContraWR} & \SecondMeanStd{0.465}{0.042} & \SecondMeanStd{0.376}{0.059} & \BestMeanStd{0.508}{0.037} & \MeanStd{0.430}{0.047} & \SecondMeanStd{0.342}{0.070} & \SecondMeanStd{0.473}{0.041} & \MeanStd{0.286}{0.101} & \Best{0.411} & 0.345 & 0.211 \\*
         & \ResultMethod{FFCL} & \MeanStd{0.349}{0.072} & \MeanStd{0.368}{0.065} & \MeanStd{0.485}{0.042} & \BestMeanStd{0.474}{0.043} & \MeanStd{0.335}{0.077} & \BestMeanStd{0.491}{0.041} & \SecondMeanStd{0.339}{0.076} & \Second{0.406} & 0.344 & 0.161 \\*
         & \ResultMethod{LaBraM} & \MeanStd{0.219}{0.090} & \MeanStd{0.200}{0.130} & \MeanStd{0.318}{0.033} & \MeanStd{0.360}{0.026} & \MeanStd{0.251}{0.058} & \MeanStd{0.292}{0.040} & \MeanStd{0.280}{0.044} & 0.274 & 0.344 & \SecondRightEdge{0.123} \\*
         & \ResultMethod{SIREN-based} & \MeanStd{0.321}{0.050} & \MeanStd{0.353}{0.042} & \MeanStd{0.317}{0.052} & \MeanStd{0.363}{0.040} & \MeanStd{0.194}{0.176} & \MeanStd{0.397}{0.034} & \MeanStd{0.221}{0.133} & 0.309 & 0.335 & 0.137 \\*
         & \ResultMethod{REVE} & \MeanStd{0.267}{0.044} & \MeanStd{0.266}{0.045} & \MeanStd{0.310}{0.032} & \MeanStd{0.242}{0.058} & \MeanStd{0.141}{0.048} & \MeanStd{0.176}{0.074} & \MeanStd{0.187}{0.089} & 0.227 & 0.337 & 0.177 \\*
         & \ResultMethod{SPaRCNet} & \BestMeanStd{0.498}{0.029} & \BestMeanStd{0.411}{0.040} & \MeanStd{0.365}{0.048} & \MeanStd{0.353}{0.051} & \BestMeanStd{0.388}{0.044} & \MeanStd{0.335}{0.057} & \MeanStd{0.198}{0.234} & 0.364 & \Second{0.334} & 0.143 \\*
         & \ResultMethod{STT-Trans.} & \MeanStd{0.228}{0.066} & \MeanStd{0.295}{0.033} & \MeanStd{0.159}{0.055} & \MeanStd{0.194}{0.096} & \MeanStd{0.278}{0.038} & \MeanStd{0.243}{0.053} & \MeanStd{0.142}{0.045} & 0.220 & 0.380 & 0.131 \\*
        \addlinespace[1pt]
         & \ResultMethod{NVCode (Ours)} & \MeanStd{0.406}{0.094} & \MeanStd{0.366}{0.130} & \SecondMeanStd{0.494}{0.059} & \SecondMeanStd{0.451}{0.072} & \MeanStd{0.266}{0.187} & \MeanStd{0.331}{0.194} & \BestMeanStd{0.389}{0.108} & 0.386 & 0.351 & 0.167 \\
\end{NVCodeDatasetTable}

\begin{NVCodeDatasetTable}{}{Encoder}{\ExperienceSampling\ (Active)}{tab:app:encoder-all-datasets:experience-sampling}
\IntraBlock{12} & \ResultMethod{BEIRA} & \MeanStd{0.446}{0.110} & \MeanStd{0.426}{0.111} & \MeanStd{0.360}{0.108} & \BestMeanStd{0.467}{0.104} & \MeanStd{0.367}{0.096} & \MeanStd{0.410}{0.099} & \MeanStd{0.430}{0.104} & 0.415 & 0.368 & 0.337 \\*
         & \ResultMethod{BIOT} & \MeanStd{0.409}{0.089} & \MeanStd{0.314}{0.077} & \MeanStd{0.341}{0.075} & \SecondMeanStd{0.433}{0.089} & \MeanStd{0.385}{0.082} & \MeanStd{0.403}{0.070} & \MeanStd{0.371}{0.094} & 0.379 & \Best{0.338} & 0.292 \\*
         & \ResultMethod{BrainOmni} & \MeanStd{0.331}{0.073} & \MeanStd{0.335}{0.059} & \MeanStd{0.422}{0.072} & \MeanStd{0.375}{0.076} & \MeanStd{0.336}{0.068} & \MeanStd{0.414}{0.064} & \MeanStd{0.320}{0.076} & 0.362 & 0.343 & 0.323 \\*
         & \ResultMethod{CNN-Trans.} & \MeanStd{0.436}{0.075} & \MeanStd{0.415}{0.085} & \MeanStd{0.382}{0.095} & \MeanStd{0.314}{0.079} & \SecondMeanStd{0.443}{0.089} & \MeanStd{0.395}{0.089} & \MeanStd{0.393}{0.094} & 0.397 & \Second{0.339} & 0.337 \\*
         & \ResultMethod{ContraWR} & \MeanStd{0.450}{0.101} & \MeanStd{0.432}{0.089} & \BestMeanStd{0.460}{0.098} & \MeanStd{0.373}{0.098} & \MeanStd{0.386}{0.077} & \MeanStd{0.379}{0.082} & \MeanStd{0.388}{0.099} & 0.410 & 0.352 & 0.340 \\*
         & \ResultMethod{FFCL} & \MeanStd{0.435}{0.119} & \MeanStd{0.416}{0.100} & \MeanStd{0.424}{0.113} & \MeanStd{0.416}{0.116} & \MeanStd{0.382}{0.086} & \MeanStd{0.405}{0.098} & \MeanStd{0.369}{0.110} & 0.407 & 0.356 & 0.325 \\*
         & \ResultMethod{LaBraM} & \MeanStd{0.369}{0.077} & \MeanStd{0.333}{0.079} & \MeanStd{0.377}{0.074} & \MeanStd{0.399}{0.089} & \MeanStd{0.373}{0.088} & \MeanStd{0.375}{0.077} & \MeanStd{0.283}{0.069} & 0.359 & 0.340 & \SecondRightEdge{0.289} \\*
         & \ResultMethod{SIREN-based} & \BestMeanStd{0.469}{0.101} & \SecondMeanStd{0.490}{0.079} & \MeanStd{0.373}{0.096} & \MeanStd{0.413}{0.090} & \MeanStd{0.430}{0.091} & \MeanStd{0.426}{0.092} & \SecondMeanStd{0.444}{0.068} & \Second{0.435} & 0.352 & 0.331 \\*
         & \ResultMethod{REVE} & \MeanStd{0.420}{0.089} & \MeanStd{0.313}{0.082} & \MeanStd{0.290}{0.071} & \MeanStd{0.363}{0.075} & \MeanStd{0.394}{0.084} & \MeanStd{0.379}{0.083} & \MeanStd{0.319}{0.065} & 0.354 & 0.346 & 0.322 \\*
         & \ResultMethod{SPaRCNet} & \SecondMeanStd{0.459}{0.128} & \MeanStd{0.322}{0.095} & \MeanStd{0.399}{0.100} & \MeanStd{0.431}{0.104} & \MeanStd{0.419}{0.095} & \BestMeanStd{0.468}{0.103} & \MeanStd{0.412}{0.092} & 0.416 & 0.388 & 0.349 \\*
         & \ResultMethod{STT-Trans.} & \MeanStd{0.395}{0.075} & \MeanStd{0.346}{0.082} & \SecondMeanStd{0.425}{0.091} & \MeanStd{0.315}{0.084} & \MeanStd{0.320}{0.079} & \MeanStd{0.377}{0.081} & \MeanStd{0.364}{0.087} & 0.363 & 0.343 & 0.300 \\*
        \addlinespace[1pt]
         & \ResultMethod{NVCode (Ours)} & \MeanStd{0.452}{0.117} & \BestMeanStd{0.527}{0.119} & \MeanStd{0.306}{0.089} & \MeanStd{0.363}{0.108} & \BestMeanStd{0.478}{0.107} & \SecondMeanStd{0.447}{0.106} & \BestMeanStd{0.478}{0.113} & \Best{0.436} & 0.351 & \BestRightEdge{0.274} \\
\ResultBlockRule
\InterBlock{12} & \ResultMethod{BEIRA} & \MeanStd{0.154}{0.040} & \MeanStd{0.209}{0.055} & \MeanStd{0.115}{0.034} & \MeanStd{0.206}{0.054} & \MeanStd{0.307}{0.158} & \MeanStd{0.331}{0.164} & \MeanStd{0.283}{0.108} & 0.229 & 0.331 & 0.100 \\*
         & \ResultMethod{BIOT} & \MeanStd{0.147}{0.039} & \MeanStd{0.303}{0.146} & \MeanStd{0.281}{0.106} & \MeanStd{0.269}{0.092} & \MeanStd{0.262}{0.085} & \MeanStd{0.231}{0.065} & \MeanStd{0.183}{0.047} & 0.239 & 0.335 & 0.130 \\*
         & \ResultMethod{BrainOmni} & \BestMeanStd{0.347}{0.108} & \MeanStd{0.246}{0.063} & \MeanStd{0.272}{0.082} & \MeanStd{0.290}{0.102} & \MeanStd{0.306}{0.134} & \SecondMeanStd{0.377}{0.074} & \MeanStd{0.283}{0.093} & 0.303 & 0.328 & 0.080 \\*
         & \ResultMethod{CNN-Trans.} & \MeanStd{0.175}{0.051} & \MeanStd{0.345}{0.146} & \MeanStd{0.150}{0.045} & \MeanStd{0.302}{0.163} & \MeanStd{0.203}{0.060} & \MeanStd{0.290}{0.136} & \MeanStd{0.247}{0.084} & 0.245 & 0.334 & 0.089 \\*
         & \ResultMethod{ContraWR} & \MeanStd{0.059}{0.037} & \MeanStd{0.209}{0.074} & \MeanStd{0.297}{0.178} & \MeanStd{0.251}{0.102} & \MeanStd{0.262}{0.114} & \MeanStd{0.240}{0.093} & \MeanStd{0.290}{0.159} & 0.230 & 0.345 & 0.135 \\*
         & \ResultMethod{FFCL} & \MeanStd{0.170}{0.036} & \MeanStd{0.346}{0.103} & \MeanStd{0.328}{0.142} & \MeanStd{0.278}{0.083} & \SecondMeanStd{0.462}{0.037} & \MeanStd{0.275}{0.080} & \SecondMeanStd{0.434}{0.044} & 0.327 & 0.339 & 0.125 \\*
         & \ResultMethod{LaBraM} & \SecondMeanStd{0.298}{0.135} & \MeanStd{0.309}{0.164} & \MeanStd{0.250}{0.076} & \BestMeanStd{0.395}{0.073} & \MeanStd{0.456}{0.048} & \MeanStd{0.348}{0.124} & \MeanStd{0.380}{0.085} & \Second{0.348} & \Best{0.326} & 0.092 \\*
         & \ResultMethod{SIREN-based} & \MeanStd{0.258}{0.098} & \MeanStd{0.278}{0.123} & \MeanStd{0.290}{0.143} & \MeanStd{0.261}{0.101} & \MeanStd{0.249}{0.091} & \MeanStd{0.246}{0.088} & \MeanStd{0.251}{0.092} & 0.262 & \Second{0.328} & 0.087 \\*
         & \ResultMethod{REVE} & \MeanStd{0.156}{0.045} & \MeanStd{0.300}{0.154} & \MeanStd{0.199}{0.058} & \MeanStd{0.260}{0.092} & \MeanStd{0.256}{0.089} & \MeanStd{0.311}{0.190} & \MeanStd{0.206}{0.060} & 0.241 & 0.334 & \BestRightEdge{0.049} \\*
         & \ResultMethod{SPaRCNet} & \MeanStd{0.286}{0.072} & \BestMeanStd{0.373}{0.057} & \SecondMeanStd{0.339}{0.090} & \MeanStd{0.321}{0.128} & \MeanStd{0.295}{0.081} & \MeanStd{0.336}{0.094} & \MeanStd{0.277}{0.064} & 0.318 & 0.329 & 0.070 \\*
         & \ResultMethod{STT-Trans.} & \MeanStd{0.195}{0.053} & \MeanStd{0.307}{0.163} & \MeanStd{0.195}{0.053} & \MeanStd{0.172}{0.046} & \MeanStd{0.326}{0.189} & \MeanStd{0.261}{0.088} & \MeanStd{0.205}{0.056} & 0.237 & 0.334 & 0.090 \\*
        \addlinespace[1pt]
         & \ResultMethod{NVCode (Ours)} & \MeanStd{0.235}{0.126} & \SecondMeanStd{0.348}{0.232} & \BestMeanStd{0.342}{0.256} & \SecondMeanStd{0.378}{0.160} & \BestMeanStd{0.575}{0.053} & \BestMeanStd{0.423}{0.111} & \BestMeanStd{0.440}{0.098} & \Best{0.392} & 0.332 & \SecondRightEdge{0.066} \\
\end{NVCodeDatasetTable}

\begin{NVCodeDatasetTable}{}{Encoder}{NatView (Active)}{tab:app:encoder-all-datasets:natview}
\IntraBlock{12} & \ResultMethod{BEIRA} & \MeanOnly{-0.418}{0.582} & \MeanOnly{0.406}{0.594} & \MeanOnly{-0.671}{0.329} & \MeanOnly{-0.105}{0.895} & \MeanOnly{0.221}{0.779} & \MeanOnly{-0.155}{0.845} & \MeanOnly{0.236}{0.764} & -0.070 & 0.347 & 0.456 \\*
         & \ResultMethod{BIOT} & \SecondMeanOnly{0.344}{0.656} & \MeanOnly{0.072}{0.928} & \BestMeanOnly{0.521}{0.479} & \MeanOnly{-0.075}{0.925} & \BestMeanOnly{0.436}{0.564} & \MeanOnly{-0.026}{0.974} & \MeanOnly{0.228}{0.772} & \Second{0.214} & 0.307 & 0.563 \\*
         & \ResultMethod{BrainOmni} & \MeanOnly{-0.085}{0.207} & \BestMeanOnly{0.652}{0.348} & \MeanOnly{-0.085}{0.915} & \MeanOnly{0.107}{0.893} & \MeanOnly{-0.187}{0.813} & \MeanOnly{0.012}{0.988} & \MeanOnly{-0.651}{0.349} & -0.034 & 0.312 & 0.529 \\*
         & \ResultMethod{CNN-Trans.} & \MeanOnly{-0.646}{0.354} & \SecondMeanOnly{0.506}{0.494} & \MeanOnly{-0.620}{0.117} & \MeanOnly{0.190}{0.810} & \MeanOnly{0.205}{0.795} & \MeanOnly{0.053}{0.139} & \MeanOnly{0.187}{0.347} & -0.018 & 0.301 & 0.521 \\*
         & \ResultMethod{ContraWR} & \MeanOnly{0.100}{0.900} & \MeanOnly{0.471}{0.529} & \MeanOnly{-0.269}{0.731} & \MeanOnly{0.198}{0.802} & \MeanOnly{0.095}{0.905} & \MeanOnly{-0.313}{0.687} & \SecondMeanOnly{0.494}{0.506} & 0.111 & 0.328 & 0.585 \\*
         & \ResultMethod{FFCL} & \MeanOnly{0.053}{0.947} & \MeanOnly{-0.260}{0.740} & \MeanOnly{0.111}{0.889} & \MeanOnly{-0.214}{0.786} & \MeanOnly{0.267}{0.229} & \MeanOnly{-0.378}{0.622} & \MeanOnly{-0.238}{0.762} & -0.094 & \Best{0.271} & 0.618 \\*
         & \ResultMethod{LaBraM} & \MeanOnly{-0.207}{0.702} & \MeanOnly{-0.479}{0.521} & \MeanOnly{-0.003}{0.997} & \MeanOnly{0.150}{0.850} & \MeanOnly{0.024}{0.976} & \MeanOnly{-0.114}{0.886} & \MeanOnly{-0.301}{0.699} & -0.133 & 0.309 & 0.521 \\*
         & \ResultMethod{SIREN-based} & \BestMeanOnly{0.607}{0.393} & \MeanOnly{0.059}{0.941} & \MeanOnly{0.106}{0.894} & \MeanOnly{-0.174}{0.003} & \SecondMeanOnly{0.405}{0.595} & \MeanOnly{0.106}{0.894} & \MeanOnly{0.148}{0.852} & 0.180 & 0.405 & \SecondRightEdge{0.448} \\*
         & \ResultMethod{REVE} & \MeanOnly{-0.712}{0.288} & \MeanOnly{0.428}{0.572} & \SecondMeanOnly{0.150}{0.850} & \MeanOnly{0.062}{0.938} & \MeanOnly{-0.102}{0.898} & \MeanOnly{-0.816}{0.184} & \MeanOnly{0.213}{0.787} & -0.111 & \Second{0.295} & 0.517 \\*
         & \ResultMethod{SPaRCNet} & \MeanOnly{-0.133}{0.867} & \MeanOnly{-0.066}{0.934} & \MeanOnly{-0.815}{0.185} & \MeanOnly{-0.477}{0.523} & \MeanOnly{-0.148}{0.411} & \MeanOnly{-0.049}{0.951} & \MeanOnly{0.447}{0.553} & -0.177 & 0.300 & 0.453 \\*
         & \ResultMethod{STT-Trans.} & \MeanOnly{0.168}{0.832} & \MeanOnly{-0.047}{0.953} & \MeanOnly{-0.178}{0.822} & \SecondMeanOnly{0.244}{0.756} & \MeanOnly{0.004}{0.045} & \BestMeanOnly{0.398}{0.602} & \MeanOnly{-0.308}{0.692} & 0.040 & 0.314 & 0.462 \\*
        \addlinespace[1pt]
         & \ResultMethod{NVCode (Ours)} & \MeanOnly{0.103}{0.897} & \MeanOnly{0.184}{0.816} & \MeanOnly{0.082}{0.918} & \BestMeanOnly{0.816}{0.184} & \MeanOnly{0.194}{0.806} & \SecondMeanOnly{0.301}{0.699} & \BestMeanOnly{0.518}{0.482} & \Best{0.314} & 0.368 & \BestRightEdge{0.385} \\
\ResultBlockRule
\InterBlock{12} & \ResultMethod{BEIRA} & \SecondMeanOnly{0.422}{0.400} & \MeanOnly{-0.541}{0.400} & \MeanOnly{-0.538}{0.400} & \MeanOnly{-0.570}{0.400} & \MeanOnly{-0.711}{0.289} & \MeanOnly{-0.760}{0.240} & \MeanOnly{-0.714}{0.286} & -0.488 & 0.426 & 0.453 \\*
         & \ResultMethod{BIOT} & \MeanOnly{0.115}{0.400} & \MeanOnly{-0.238}{0.400} & \MeanOnly{-0.102}{0.400} & \BestMeanOnly{0.676}{0.167} & \MeanOnly{-0.150}{0.400} & \BestMeanOnly{0.341}{0.314} & \SecondMeanOnly{0.039}{0.400} & 0.097 & 0.297 & 0.358 \\*
         & \ResultMethod{BrainOmni} & \MeanOnly{-0.267}{0.400} & \BestMeanOnly{0.722}{0.212} & \MeanOnly{0.125}{0.400} & \SecondMeanOnly{0.290}{0.400} & \BestMeanOnly{0.616}{0.246} & \MeanOnly{0.103}{0.400} & \MeanOnly{-0.150}{0.400} & \Second{0.205} & 0.292 & 0.387 \\*
         & \ResultMethod{CNN-Trans.} & \BestMeanOnly{0.726}{0.223} & \MeanOnly{0.369}{0.400} & \MeanOnly{-0.260}{0.400} & \MeanOnly{0.088}{0.400} & \SecondMeanOnly{0.403}{0.385} & \MeanOnly{-0.232}{0.400} & \MeanOnly{-0.602}{0.266} & 0.070 & 0.300 & 0.525 \\*
         & \ResultMethod{ContraWR} & \MeanOnly{-0.265}{0.400} & \MeanOnly{-0.734}{0.266} & \MeanOnly{0.299}{0.400} & \MeanOnly{0.229}{0.400} & \MeanOnly{-0.357}{0.400} & \MeanOnly{-0.057}{0.400} & \MeanOnly{-0.773}{0.227} & -0.237 & 0.320 & 0.398 \\*
         & \ResultMethod{FFCL} & \MeanOnly{0.208}{0.298} & \MeanOnly{0.025}{0.400} & \MeanOnly{-0.019}{0.400} & \MeanOnly{0.027}{0.400} & \MeanOnly{-0.547}{0.126} & \MeanOnly{0.129}{0.400} & \MeanOnly{-0.589}{0.117} & -0.109 & 0.375 & 0.475 \\*
         & \ResultMethod{LaBraM} & \MeanOnly{0.163}{0.400} & \SecondMeanOnly{0.420}{0.197} & \MeanOnly{0.272}{0.290} & \MeanOnly{0.147}{0.400} & \MeanOnly{-0.127}{0.400} & \MeanOnly{0.002}{0.400} & \MeanOnly{-0.107}{0.400} & 0.110 & \Best{0.288} & 0.544 \\*
         & \ResultMethod{SIREN-based} & \MeanOnly{0.291}{0.399} & \MeanOnly{-0.031}{0.400} & \MeanOnly{-0.727}{0.173} & \MeanOnly{-0.137}{0.400} & \MeanOnly{-0.480}{0.254} & \MeanOnly{-0.227}{0.400} & \MeanOnly{-0.278}{0.400} & -0.227 & 0.379 & 0.372 \\*
         & \ResultMethod{REVE} & \MeanOnly{0.051}{0.400} & \MeanOnly{0.067}{0.400} & \MeanOnly{-0.172}{0.360} & \MeanOnly{-0.242}{0.271} & \MeanOnly{-0.195}{0.325} & \MeanOnly{-0.529}{0.134} & \MeanOnly{-0.346}{0.198} & -0.195 & 0.294 & 0.544 \\*
         & \ResultMethod{SPaRCNet} & \MeanOnly{0.406}{0.189} & \MeanOnly{-0.149}{0.400} & \SecondMeanOnly{0.429}{0.179} & \MeanOnly{-0.027}{0.400} & \MeanOnly{-0.095}{0.400} & \MeanOnly{-0.043}{0.400} & \MeanOnly{-0.010}{0.400} & 0.073 & \Second{0.289} & 0.296 \\*
         & \ResultMethod{STT-Trans.} & \MeanOnly{0.011}{0.400} & \MeanOnly{-0.135}{0.333} & \BestMeanOnly{0.543}{0.100} & \MeanOnly{-0.060}{0.400} & \MeanOnly{-0.014}{0.400} & \MeanOnly{0.073}{0.400} & \MeanOnly{-0.065}{0.400} & 0.050 & 0.320 & \SecondRightEdge{0.288} \\*
        \addlinespace[1pt]
         & \ResultMethod{NVCode (Ours)} & \MeanOnly{0.127}{0.362} & \MeanOnly{0.274}{0.357} & \MeanOnly{0.394}{0.174} & \MeanOnly{-0.003}{0.168} & \MeanOnly{0.235}{0.400} & \SecondMeanOnly{0.177}{0.400} & \BestMeanOnly{0.369}{0.195} & \Best{0.225} & 0.311 & \BestRightEdge{0.249} \\
\end{NVCodeDatasetTable}

\Needspace{215pt}
\subsection{Per Dataset Generator Results}
\label{app:additional:generators}

All generation networks share the seven frozen NVCode encoders and the frozen NVCode-trained projector. Each table reports the two evaluation splits for one dataset using the same aggregation and ranking conventions as the encoder tables.

\begin{NVCodeDatasetTable}{*}{Generator}{\SleepEEGfMRI\ (Resting State)}{tab:app:generator-all-datasets}
\IntraBlock{8} & \ResultMethod{cGAN} & \MeanStd{0.276}{0.152} & \MeanStd{0.307}{0.113} & \MeanStd{0.440}{0.111} & \MeanStd{0.330}{0.115} & \MeanStd{0.398}{0.103} & \MeanStd{0.358}{0.139} & \MeanStd{0.304}{0.146} & 0.344 & 0.459 & 0.567 \\*
         & \ResultMethod{Consistency} & \MeanStd{0.413}{0.108} & \MeanStd{0.405}{0.109} & \MeanStd{0.471}{0.136} & \MeanStd{0.383}{0.110} & \MeanStd{0.326}{0.109} & \MeanStd{0.379}{0.096} & \MeanStd{0.396}{0.154} & 0.396 & 0.439 & 0.287 \\*
         & \ResultMethod{CVAE} & \BestMeanStd{0.538}{0.126} & \MeanStd{0.438}{0.112} & \BestMeanStd{0.573}{0.149} & \MeanStd{0.432}{0.149} & \MeanStd{0.428}{0.136} & \MeanStd{0.325}{0.140} & \MeanStd{0.489}{0.142} & 0.461 & 0.351 & 0.276 \\*
         & \ResultMethod{DDPM} & \SecondMeanStd{0.491}{0.132} & \MeanStd{0.496}{0.112} & \MeanStd{0.436}{0.134} & \MeanStd{0.407}{0.161} & \MeanStd{0.424}{0.146} & \MeanStd{0.386}{0.127} & \MeanStd{0.405}{0.133} & 0.435 & 0.343 & \SecondRightEdge{0.249} \\*
         & \ResultMethod{EDM} & \MeanStd{0.436}{0.154} & \MeanStd{0.510}{0.173} & \SecondMeanStd{0.550}{0.116} & \BestMeanStd{0.524}{0.176} & \SecondMeanStd{0.472}{0.182} & \MeanStd{0.388}{0.174} & \SecondMeanStd{0.529}{0.177} & \Second{0.487} & \Second{0.340} & 0.258 \\*
         & \ResultMethod{Rectified Flow} & \MeanStd{0.467}{0.150} & \MeanStd{0.437}{0.131} & \MeanStd{0.428}{0.116} & \MeanStd{0.438}{0.098} & \MeanStd{0.457}{0.106} & \SecondMeanStd{0.458}{0.116} & \MeanStd{0.363}{0.135} & 0.435 & 0.346 & 0.280 \\*
         & \ResultMethod{Score SDE} & \MeanStd{0.455}{0.116} & \BestMeanStd{0.554}{0.138} & \MeanStd{0.422}{0.145} & \MeanStd{0.472}{0.138} & \MeanStd{0.463}{0.143} & \MeanStd{0.408}{0.149} & \MeanStd{0.499}{0.120} & 0.468 & 0.343 & 0.299 \\*
        \addlinespace[1pt]
         & \ResultMethod{NVCode (Ours)} & \MeanStd{0.366}{0.175} & \SecondMeanStd{0.522}{0.186} & \MeanStd{0.472}{0.143} & \SecondMeanStd{0.524}{0.135} & \BestMeanStd{0.608}{0.118} & \BestMeanStd{0.491}{0.131} & \BestMeanStd{0.534}{0.145} & \Best{0.502} & \Best{0.339} & \BestRightEdge{0.239} \\
\ResultBlockRule
\InterBlock{8} & \ResultMethod{cGAN} & \SecondMeanStd{0.400}{0.113} & \MeanStd{0.412}{0.110} & \MeanStd{0.330}{0.134} & \MeanStd{0.219}{0.192} & \MeanStd{0.310}{0.142} & \MeanStd{0.306}{0.143} & \MeanStd{0.384}{0.117} & 0.337 & 0.338 & 0.180 \\*
         & \ResultMethod{Consistency} & \BestMeanStd{0.425}{0.105} & \MeanStd{0.450}{0.099} & \MeanStd{0.399}{0.111} & \MeanStd{0.381}{0.116} & \MeanStd{0.383}{0.115} & \MeanStd{0.316}{0.137} & \MeanStd{0.365}{0.120} & 0.388 & 0.433 & 0.169 \\*
         & \ResultMethod{CVAE} & \MeanStd{0.267}{0.252} & \BestMeanStd{0.642}{0.113} & \SecondMeanStd{0.470}{0.151} & \BestMeanStd{0.436}{0.162} & \SecondMeanStd{0.433}{0.163} & \BestMeanStd{0.462}{0.154} & \SecondMeanStd{0.437}{0.162} & \Second{0.449} & 0.334 & 0.272 \\*
         & \ResultMethod{DDPM} & \MeanStd{0.261}{0.157} & \MeanStd{0.443}{0.098} & \MeanStd{0.357}{0.119} & \MeanStd{0.364}{0.117} & \MeanStd{0.317}{0.132} & \MeanStd{0.340}{0.124} & \MeanStd{0.396}{0.108} & 0.354 & 0.331 & \SecondRightEdge{0.128} \\*
         & \ResultMethod{EDM} & \MeanStd{0.220}{0.214} & \MeanStd{0.589}{0.088} & \MeanStd{0.362}{0.138} & \MeanStd{0.312}{0.158} & \MeanStd{0.339}{0.146} & \MeanStd{0.405}{0.125} & \MeanStd{0.352}{0.141} & 0.368 & 0.335 & 0.173 \\*
         & \ResultMethod{Rectified Flow} & \MeanStd{0.381}{0.139} & \MeanStd{0.446}{0.121} & \MeanStd{0.446}{0.120} & \MeanStd{0.254}{0.199} & \MeanStd{0.276}{0.185} & \MeanStd{0.349}{0.151} & \BestMeanStd{0.475}{0.114} & 0.375 & 0.336 & 0.189 \\*
         & \ResultMethod{Score SDE} & \MeanStd{0.290}{0.178} & \MeanStd{0.594}{0.092} & \MeanStd{0.399}{0.133} & \MeanStd{0.322}{0.162} & \MeanStd{0.371}{0.143} & \MeanStd{0.406}{0.132} & \MeanStd{0.335}{0.156} & 0.388 & \Second{0.329} & 0.157 \\*
        \addlinespace[1pt]
         & \ResultMethod{NVCode (Ours)} & \MeanStd{0.194}{0.207} & \SecondMeanStd{0.613}{0.142} & \BestMeanStd{0.553}{0.178} & \SecondMeanStd{0.410}{0.400} & \BestMeanStd{0.557}{0.175} & \SecondMeanStd{0.433}{0.362} & \MeanStd{0.430}{0.373} & \Best{0.456} & \Best{0.327} & \BestRightEdge{0.096} \\
\end{NVCodeDatasetTable}

\begin{NVCodeDatasetTable}{}{Generator}{\MultiCLARID\ (Resting State)}{tab:app:generator-all-datasets:multi-clarid}
\IntraBlock{8} & \ResultMethod{cGAN} & \MeanStd{0.374}{0.107} & \MeanStd{0.365}{0.125} & \MeanStd{0.415}{0.108} & \MeanStd{0.414}{0.092} & \SecondMeanStd{0.571}{0.101} & \MeanStd{0.504}{0.131} & \SecondMeanStd{0.382}{0.116} & 0.432 & 0.519 & 0.548 \\*
         & \ResultMethod{Consistency} & \MeanStd{0.400}{0.087} & \MeanStd{0.400}{0.109} & \MeanStd{0.414}{0.106} & \MeanStd{0.376}{0.099} & \MeanStd{0.428}{0.105} & \MeanStd{0.447}{0.112} & \MeanStd{0.311}{0.101} & 0.397 & 0.505 & 0.386 \\*
         & \ResultMethod{CVAE} & \MeanStd{0.390}{0.108} & \SecondMeanStd{0.408}{0.116} & \SecondMeanStd{0.459}{0.131} & \MeanStd{0.395}{0.085} & \MeanStd{0.566}{0.116} & \SecondMeanStd{0.509}{0.131} & \MeanStd{0.377}{0.111} & \Second{0.444} & 0.431 & 0.331 \\*
         & \ResultMethod{DDPM} & \MeanStd{0.341}{0.094} & \MeanStd{0.401}{0.102} & \MeanStd{0.438}{0.119} & \MeanStd{0.331}{0.080} & \MeanStd{0.512}{0.119} & \MeanStd{0.479}{0.134} & \MeanStd{0.366}{0.107} & 0.410 & \Second{0.397} & 0.334 \\*
         & \ResultMethod{EDM} & \SecondMeanStd{0.420}{0.094} & \MeanStd{0.405}{0.104} & \MeanStd{0.458}{0.107} & \MeanStd{0.389}{0.084} & \MeanStd{0.506}{0.118} & \MeanStd{0.477}{0.122} & \MeanStd{0.343}{0.099} & 0.428 & 0.413 & \SecondRightEdge{0.322} \\*
         & \ResultMethod{Rectified Flow} & \MeanStd{0.405}{0.081} & \MeanStd{0.382}{0.113} & \MeanStd{0.434}{0.103} & \BestMeanStd{0.425}{0.089} & \MeanStd{0.484}{0.118} & \MeanStd{0.469}{0.114} & \MeanStd{0.321}{0.097} & 0.417 & 0.417 & 0.359 \\*
         & \ResultMethod{Score SDE} & \MeanStd{0.363}{0.090} & \MeanStd{0.386}{0.119} & \MeanStd{0.425}{0.119} & \MeanStd{0.333}{0.091} & \MeanStd{0.552}{0.102} & \MeanStd{0.472}{0.118} & \MeanStd{0.350}{0.104} & 0.411 & 0.409 & 0.327 \\*
        \addlinespace[1pt]
         & \ResultMethod{NVCode (Ours)} & \BestMeanStd{0.524}{0.100} & \BestMeanStd{0.484}{0.128} & \BestMeanStd{0.639}{0.124} & \SecondMeanStd{0.416}{0.096} & \BestMeanStd{0.606}{0.119} & \BestMeanStd{0.650}{0.117} & \BestMeanStd{0.482}{0.124} & \Best{0.543} & \Best{0.384} & \BestRightEdge{0.276} \\
\ResultBlockRule
\InterBlock{8} & \ResultMethod{cGAN} & \MeanStd{0.316}{0.106} & \BestMeanStd{0.378}{0.091} & \SecondMeanStd{0.461}{0.076} & \MeanStd{0.365}{0.094} & \MeanStd{0.695}{0.052} & \MeanStd{0.654}{0.055} & \MeanStd{0.363}{0.094} & \Second{0.462} & 0.394 & 0.263 \\*
         & \ResultMethod{Consistency} & \MeanStd{0.354}{0.092} & \MeanStd{0.270}{0.117} & \BestMeanStd{0.466}{0.072} & \MeanStd{0.385}{0.085} & \MeanStd{0.697}{0.049} & \SecondMeanStd{0.666}{0.051} & \MeanStd{0.360}{0.091} & 0.457 & 0.472 & 0.292 \\*
         & \ResultMethod{CVAE} & \SecondMeanStd{0.389}{0.092} & \MeanStd{0.318}{0.111} & \MeanStd{0.447}{0.081} & \MeanStd{0.328}{0.108} & \MeanStd{0.694}{0.054} & \MeanStd{0.627}{0.059} & \MeanStd{0.376}{0.095} & 0.454 & 0.381 & 0.191 \\*
         & \ResultMethod{DDPM} & \BestMeanStd{0.399}{0.104} & \MeanStd{0.316}{0.128} & \MeanStd{0.418}{0.099} & \BestMeanStd{0.469}{0.090} & \MeanStd{0.616}{0.069} & \MeanStd{0.538}{0.079} & \MeanStd{0.300}{0.134} & 0.437 & 0.382 & \SecondRightEdge{0.101} \\*
         & \ResultMethod{EDM} & \MeanStd{0.339}{0.113} & \MeanStd{0.315}{0.121} & \MeanStd{0.436}{0.090} & \MeanStd{0.390}{0.100} & \MeanStd{0.632}{0.064} & \MeanStd{0.591}{0.068} & \BestMeanStd{0.459}{0.086} & 0.452 & \Best{0.374} & \BestRightEdge{0.089} \\*
         & \ResultMethod{Rectified Flow} & \MeanStd{0.368}{0.078} & \MeanStd{0.207}{0.129} & \MeanStd{0.415}{0.070} & \MeanStd{0.313}{0.090} & \SecondMeanStd{0.705}{0.043} & \MeanStd{0.629}{0.047} & \MeanStd{0.362}{0.079} & 0.429 & \Second{0.377} & 0.135 \\*
         & \ResultMethod{Score SDE} & \MeanStd{0.382}{0.094} & \MeanStd{0.265}{0.131} & \MeanStd{0.331}{0.107} & \MeanStd{0.322}{0.110} & \MeanStd{0.695}{0.054} & \MeanStd{0.614}{0.061} & \MeanStd{0.308}{0.114} & 0.417 & 0.379 & 0.203 \\*
        \addlinespace[1pt]
         & \ResultMethod{NVCode (Ours)} & \MeanStd{0.324}{0.134} & \SecondMeanStd{0.361}{0.197} & \MeanStd{0.428}{0.229} & \SecondMeanStd{0.395}{0.349} & \BestMeanStd{0.768}{0.038} & \BestMeanStd{0.742}{0.041} & \SecondMeanStd{0.436}{0.206} & \Best{0.493} & 0.380 & 0.105 \\
\end{NVCodeDatasetTable}

\begin{NVCodeDatasetTable}{}{Generator}{\GradCPT\ (Resting State)}{tab:app:generator-all-datasets:gradcpt}
\IntraBlock{8} & \ResultMethod{cGAN} & \MeanStd{0.285}{0.163} & \MeanStd{0.272}{0.160} & \MeanStd{0.398}{0.126} & \MeanStd{0.369}{0.142} & \MeanStd{0.404}{0.131} & \MeanStd{0.377}{0.165} & \MeanStd{0.319}{0.142} & 0.346 & 0.451 & 0.533 \\*
         & \ResultMethod{Consistency} & \MeanStd{0.389}{0.115} & \MeanStd{0.385}{0.125} & \MeanStd{0.361}{0.097} & \MeanStd{0.335}{0.121} & \MeanStd{0.356}{0.104} & \MeanStd{0.370}{0.129} & \MeanStd{0.377}{0.095} & 0.367 & 0.418 & 0.357 \\*
         & \ResultMethod{CVAE} & \MeanStd{0.394}{0.157} & \MeanStd{0.315}{0.166} & \BestMeanStd{0.480}{0.124} & \MeanStd{0.333}{0.169} & \MeanStd{0.374}{0.156} & \MeanStd{0.228}{0.129} & \MeanStd{0.326}{0.138} & 0.350 & 0.345 & 0.348 \\*
         & \ResultMethod{DDPM} & \MeanStd{0.330}{0.154} & \MeanStd{0.300}{0.167} & \MeanStd{0.391}{0.143} & \MeanStd{0.441}{0.140} & \MeanStd{0.425}{0.171} & \MeanStd{0.353}{0.142} & \MeanStd{0.396}{0.171} & 0.377 & 0.332 & 0.327 \\*
         & \ResultMethod{EDM} & \SecondMeanStd{0.428}{0.131} & \MeanStd{0.436}{0.175} & \MeanStd{0.383}{0.144} & \MeanStd{0.393}{0.136} & \SecondMeanStd{0.463}{0.149} & \SecondMeanStd{0.423}{0.140} & \MeanStd{0.420}{0.151} & \Second{0.421} & \Second{0.330} & \SecondRightEdge{0.324} \\*
         & \ResultMethod{Rectified Flow} & \BestMeanStd{0.507}{0.118} & \MeanStd{0.416}{0.148} & \MeanStd{0.312}{0.147} & \MeanStd{0.344}{0.152} & \MeanStd{0.411}{0.156} & \MeanStd{0.361}{0.138} & \SecondMeanStd{0.425}{0.143} & 0.397 & 0.333 & 0.332 \\*
         & \ResultMethod{Score SDE} & \MeanStd{0.255}{0.136} & \SecondMeanStd{0.444}{0.138} & \MeanStd{0.379}{0.151} & \BestMeanStd{0.518}{0.138} & \MeanStd{0.427}{0.133} & \MeanStd{0.417}{0.148} & \MeanStd{0.401}{0.152} & 0.406 & 0.333 & 0.359 \\*
        \addlinespace[1pt]
         & \ResultMethod{NVCode (Ours)} & \MeanStd{0.392}{0.200} & \BestMeanStd{0.601}{0.181} & \SecondMeanStd{0.424}{0.144} & \SecondMeanStd{0.463}{0.180} & \BestMeanStd{0.577}{0.188} & \BestMeanStd{0.538}{0.174} & \BestMeanStd{0.609}{0.166} & \Best{0.515} & \Best{0.329} & \BestRightEdge{0.309} \\
\ResultBlockRule
\InterBlock{8} & \ResultMethod{cGAN} & \MeanStd{0.292}{0.228} & \MeanStd{0.427}{0.162} & \MeanStd{0.284}{0.233} & \BestMeanStd{0.403}{0.171} & \MeanStd{0.408}{0.169} & \SecondMeanStd{0.366}{0.186} & \MeanStd{0.413}{0.167} & 0.370 & \Second{0.317} & 0.161 \\*
         & \ResultMethod{Consistency} & \MeanStd{0.325}{0.152} & \MeanStd{0.244}{0.195} & \MeanStd{0.298}{0.164} & \MeanStd{0.342}{0.145} & \MeanStd{0.399}{0.126} & \MeanStd{0.341}{0.145} & \MeanStd{0.425}{0.119} & 0.339 & 0.410 & 0.195 \\*
         & \ResultMethod{CVAE} & \MeanStd{0.385}{0.153} & \MeanStd{0.417}{0.142} & \MeanStd{0.355}{0.164} & \MeanStd{0.088}{0.400} & \MeanStd{0.441}{0.135} & \MeanStd{0.350}{0.166} & \SecondMeanStd{0.463}{0.129} & 0.357 & 0.322 & 0.188 \\*
         & \ResultMethod{DDPM} & \MeanStd{0.315}{0.195} & \SecondMeanStd{0.437}{0.145} & \MeanStd{0.353}{0.176} & \MeanStd{0.324}{0.190} & \SecondMeanStd{0.458}{0.139} & \MeanStd{0.357}{0.174} & \MeanStd{0.459}{0.139} & 0.386 & 0.317 & \SecondRightEdge{0.082} \\*
         & \ResultMethod{EDM} & \MeanStd{0.436}{0.144} & \MeanStd{0.420}{0.149} & \MeanStd{0.257}{0.231} & \SecondMeanStd{0.347}{0.177} & \MeanStd{0.449}{0.140} & \MeanStd{0.321}{0.190} & \MeanStd{0.452}{0.140} & 0.383 & 0.317 & 0.131 \\*
         & \ResultMethod{Rectified Flow} & \MeanStd{0.361}{0.144} & \MeanStd{0.326}{0.158} & \MeanStd{0.313}{0.164} & \MeanStd{0.146}{0.311} & \MeanStd{0.321}{0.161} & \MeanStd{0.298}{0.172} & \MeanStd{0.445}{0.120} & 0.316 & 0.325 & 0.178 \\*
         & \ResultMethod{Score SDE} & \SecondMeanStd{0.451}{0.158} & \MeanStd{0.409}{0.172} & \SecondMeanStd{0.380}{0.184} & \MeanStd{0.267}{0.253} & \MeanStd{0.433}{0.164} & \MeanStd{0.340}{0.204} & \MeanStd{0.449}{0.158} & \Second{0.390} & 0.321 & 0.182 \\*
        \addlinespace[1pt]
         & \ResultMethod{NVCode (Ours)} & \BestMeanStd{0.576}{0.219} & \BestMeanStd{0.560}{0.236} & \BestMeanStd{0.482}{0.388} & \MeanStd{0.222}{0.217} & \BestMeanStd{0.509}{0.317} & \BestMeanStd{0.412}{0.400} & \BestMeanStd{0.587}{0.208} & \Best{0.478} & \Best{0.314} & \BestRightEdge{0.044} \\
\end{NVCodeDatasetTable}

\begin{NVCodeDatasetTable}{}{Generator}{NeuroBOLT (Resting State)}{tab:app:generator-all-datasets:neurobolt}
\IntraBlock{8} & \ResultMethod{cGAN} & \MeanStd{0.459}{0.106} & \MeanStd{0.418}{0.095} & \MeanStd{0.271}{0.083} & \MeanStd{0.367}{0.081} & \MeanStd{0.213}{0.099} & \MeanStd{0.266}{0.089} & \MeanStd{0.444}{0.090} & 0.348 & 0.326 & 0.198 \\*
         & \ResultMethod{Consistency} & \MeanStd{0.370}{0.104} & \MeanStd{0.375}{0.100} & \MeanStd{0.228}{0.093} & \MeanStd{0.277}{0.117} & \MeanStd{0.114}{0.091} & \MeanStd{0.192}{0.086} & \MeanStd{0.317}{0.120} & 0.268 & 0.378 & 0.228 \\*
         & \ResultMethod{CVAE} & \MeanStd{0.487}{0.113} & \SecondMeanStd{0.458}{0.084} & \MeanStd{0.312}{0.110} & \SecondMeanStd{0.384}{0.109} & \SecondMeanStd{0.272}{0.088} & \MeanStd{0.264}{0.097} & \SecondMeanStd{0.480}{0.098} & \Second{0.380} & 0.300 & 0.185 \\*
         & \ResultMethod{DDPM} & \MeanStd{0.451}{0.108} & \MeanStd{0.407}{0.076} & \MeanStd{0.199}{0.096} & \MeanStd{0.319}{0.097} & \MeanStd{0.211}{0.101} & \MeanStd{0.203}{0.090} & \MeanStd{0.414}{0.098} & 0.315 & 0.287 & 0.187 \\*
         & \ResultMethod{EDM} & \SecondMeanStd{0.491}{0.099} & \MeanStd{0.445}{0.091} & \MeanStd{0.301}{0.098} & \MeanStd{0.376}{0.111} & \MeanStd{0.225}{0.089} & \MeanStd{0.258}{0.103} & \MeanStd{0.463}{0.102} & 0.366 & 0.288 & 0.199 \\*
         & \ResultMethod{Rectified Flow} & \MeanStd{0.449}{0.113} & \MeanStd{0.433}{0.100} & \MeanStd{0.274}{0.093} & \MeanStd{0.363}{0.103} & \MeanStd{0.215}{0.098} & \MeanStd{0.271}{0.082} & \MeanStd{0.436}{0.108} & 0.349 & 0.290 & \SecondRightEdge{0.180} \\*
         & \ResultMethod{Score SDE} & \MeanStd{0.472}{0.105} & \MeanStd{0.450}{0.087} & \SecondMeanStd{0.315}{0.085} & \MeanStd{0.379}{0.095} & \MeanStd{0.235}{0.092} & \SecondMeanStd{0.282}{0.083} & \MeanStd{0.461}{0.094} & 0.371 & \Second{0.282} & 0.193 \\*
        \addlinespace[1pt]
         & \ResultMethod{NVCode (Ours)} & \BestMeanStd{0.639}{0.113} & \BestMeanStd{0.671}{0.107} & \BestMeanStd{0.489}{0.153} & \BestMeanStd{0.564}{0.130} & \BestMeanStd{0.439}{0.175} & \BestMeanStd{0.515}{0.145} & \BestMeanStd{0.653}{0.110} & \Best{0.567} & \Best{0.219} & \BestRightEdge{0.167} \\
\ResultBlockRule
\InterBlock{8} & \ResultMethod{cGAN} & \MeanStd{0.569}{0.059} & \MeanStd{0.515}{0.064} & \SecondMeanStd{0.374}{0.086} & \MeanStd{0.474}{0.069} & \MeanStd{0.257}{0.120} & \MeanStd{0.287}{0.109} & \MeanStd{0.519}{0.064} & 0.428 & 0.276 & 0.227 \\*
         & \ResultMethod{Consistency} & \MeanStd{0.567}{0.057} & \MeanStd{0.506}{0.064} & \MeanStd{0.367}{0.085} & \MeanStd{0.467}{0.069} & \MeanStd{0.223}{0.132} & \MeanStd{0.281}{0.108} & \MeanStd{0.525}{0.062} & 0.420 & 0.359 & 0.182 \\*
         & \ResultMethod{CVAE} & \SecondMeanStd{0.580}{0.055} & \MeanStd{0.501}{0.063} & \MeanStd{0.340}{0.090} & \MeanStd{0.444}{0.071} & \MeanStd{0.235}{0.125} & \MeanStd{0.255}{0.116} & \MeanStd{0.503}{0.063} & 0.408 & 0.271 & 0.237 \\*
         & \ResultMethod{DDPM} & \MeanStd{0.540}{0.050} & \MeanStd{0.482}{0.056} & \MeanStd{0.304}{0.085} & \MeanStd{0.399}{0.066} & \MeanStd{0.165}{0.142} & \MeanStd{0.270}{0.094} & \MeanStd{0.470}{0.057} & 0.376 & 0.280 & \SecondRightEdge{0.079} \\*
         & \ResultMethod{EDM} & \MeanStd{0.511}{0.060} & \MeanStd{0.496}{0.061} & \MeanStd{0.363}{0.082} & \MeanStd{0.449}{0.067} & \MeanStd{0.229}{0.122} & \MeanStd{0.284}{0.102} & \MeanStd{0.499}{0.061} & 0.404 & 0.272 & 0.125 \\*
         & \ResultMethod{Rectified Flow} & \MeanStd{0.519}{0.049} & \MeanStd{0.473}{0.054} & \MeanStd{0.302}{0.081} & \MeanStd{0.433}{0.058} & \MeanStd{0.153}{0.143} & \MeanStd{0.269}{0.090} & \MeanStd{0.455}{0.056} & 0.372 & 0.277 & 0.108 \\*
         & \ResultMethod{Score SDE} & \MeanStd{0.573}{0.061} & \SecondMeanStd{0.519}{0.067} & \MeanStd{0.370}{0.092} & \SecondMeanStd{0.486}{0.071} & \SecondMeanStd{0.259}{0.126} & \SecondMeanStd{0.330}{0.102} & \SecondMeanStd{0.530}{0.066} & \Second{0.438} & \Best{0.268} & 0.189 \\*
        \addlinespace[1pt]
         & \ResultMethod{NVCode (Ours)} & \BestMeanStd{0.616}{0.049} & \BestMeanStd{0.614}{0.050} & \BestMeanStd{0.436}{0.070} & \BestMeanStd{0.523}{0.058} & \BestMeanStd{0.341}{0.090} & \BestMeanStd{0.468}{0.065} & \BestMeanStd{0.585}{0.052} & \Best{0.512} & \Second{0.268} & \BestRightEdge{0.043} \\
\end{NVCodeDatasetTable}

\begin{NVCodeDatasetTable}{}{Generator}{\XPone\ (Active)}{tab:app:generator-all-datasets:xp1}
\IntraBlock{8} & \ResultMethod{cGAN} & \MeanOnly{0.312}{0.322} & \MeanOnly{0.324}{0.254} & \MeanOnly{0.004}{0.307} & \MeanOnly{0.081}{0.312} & \MeanOnly{0.534}{0.270} & \MeanOnly{0.276}{0.358} & \MeanOnly{0.168}{0.301} & 0.243 & 0.619 & 0.603 \\*
         & \ResultMethod{Consistency} & \MeanOnly{0.353}{0.149} & \MeanOnly{0.286}{0.192} & \MeanOnly{0.364}{0.186} & \SecondMeanOnly{0.437}{0.282} & \MeanOnly{0.284}{0.288} & \BestMeanOnly{0.348}{0.260} & \MeanOnly{0.177}{0.268} & 0.321 & 0.605 & \SecondRightEdge{0.382} \\*
         & \ResultMethod{CVAE} & \SecondMeanOnly{0.575}{0.339} & \MeanOnly{0.479}{0.269} & \SecondMeanOnly{0.432}{0.266} & \MeanOnly{0.346}{0.341} & \MeanOnly{0.464}{0.324} & \MeanOnly{0.305}{0.318} & \SecondMeanOnly{0.452}{0.338} & \Second{0.436} & 0.508 & 0.411 \\*
         & \ResultMethod{DDPM} & \MeanOnly{0.523}{0.304} & \MeanOnly{0.365}{0.313} & \MeanOnly{0.334}{0.338} & \MeanOnly{0.379}{0.313} & \MeanOnly{0.532}{0.320} & \MeanOnly{0.252}{0.324} & \MeanOnly{0.339}{0.279} & 0.389 & 0.489 & 0.432 \\*
         & \ResultMethod{EDM} & \MeanOnly{0.533}{0.270} & \SecondMeanOnly{0.515}{0.275} & \MeanOnly{0.376}{0.228} & \MeanOnly{0.411}{0.330} & \MeanOnly{0.472}{0.256} & \MeanOnly{0.336}{0.275} & \MeanOnly{0.382}{0.326} & 0.432 & \Second{0.479} & 0.402 \\*
         & \ResultMethod{Rectified Flow} & \MeanOnly{0.480}{0.222} & \MeanOnly{0.458}{0.203} & \MeanOnly{0.299}{0.306} & \MeanOnly{0.362}{0.307} & \SecondMeanOnly{0.543}{0.226} & \MeanOnly{0.339}{0.237} & \MeanOnly{0.325}{0.350} & 0.401 & 0.483 & 0.395 \\*
         & \ResultMethod{Score SDE} & \MeanOnly{0.548}{0.169} & \BestMeanOnly{0.586}{0.218} & \BestMeanOnly{0.467}{0.194} & \MeanOnly{0.250}{0.336} & \MeanOnly{0.394}{0.281} & \SecondMeanOnly{0.340}{0.258} & \MeanOnly{0.433}{0.232} & 0.431 & 0.489 & 0.431 \\*
        \addlinespace[1pt]
         & \ResultMethod{NVCode (Ours)} & \BestMeanOnly{0.707}{0.293} & \MeanOnly{0.495}{0.329} & \MeanOnly{0.291}{0.315} & \BestMeanOnly{0.461}{0.378} & \BestMeanOnly{0.608}{0.342} & \MeanOnly{0.328}{0.357} & \BestMeanOnly{0.536}{0.450} & \Best{0.489} & \Best{0.432} & \BestRightEdge{0.361} \\
\ResultBlockRule
\InterBlock{8} & \ResultMethod{cGAN} & \MeanOnly{0.170}{0.329} & \MeanOnly{-0.101}{0.400} & \MeanOnly{-0.278}{0.218} & \MeanOnly{-0.020}{0.400} & \BestMeanOnly{0.456}{0.140} & \BestMeanOnly{0.481}{0.133} & \MeanOnly{-0.124}{0.400} & 0.083 & 0.495 & 0.436 \\*
         & \ResultMethod{Consistency} & \MeanOnly{0.006}{0.400} & \MeanOnly{-0.046}{0.377} & \SecondMeanOnly{0.128}{0.194} & \MeanOnly{0.131}{0.190} & \MeanOnly{0.188}{0.143} & \MeanOnly{0.141}{0.180} & \MeanOnly{-0.122}{0.201} & 0.061 & 0.577 & 0.422 \\*
         & \ResultMethod{CVAE} & \MeanOnly{0.018}{0.400} & \MeanOnly{-0.130}{0.202} & \MeanOnly{-0.006}{0.400} & \MeanOnly{0.070}{0.314} & \MeanOnly{0.185}{0.153} & \MeanOnly{0.099}{0.247} & \MeanOnly{-0.153}{0.178} & 0.012 & 0.484 & 0.415 \\*
         & \ResultMethod{DDPM} & \SecondMeanOnly{0.537}{0.175} & \SecondMeanOnly{0.157}{0.400} & \MeanOnly{-0.181}{0.400} & \SecondMeanOnly{0.189}{0.400} & \MeanOnly{-0.315}{0.284} & \MeanOnly{-0.101}{0.400} & \SecondMeanOnly{0.139}{0.400} & 0.061 & 0.484 & \BestRightEdge{0.258} \\*
         & \ResultMethod{EDM} & \MeanOnly{0.168}{0.290} & \MeanOnly{0.058}{0.400} & \MeanOnly{-0.164}{0.295} & \MeanOnly{0.043}{0.400} & \MeanOnly{0.303}{0.175} & \SecondMeanOnly{0.292}{0.181} & \MeanOnly{-0.078}{0.400} & \Second{0.089} & 0.476 & 0.374 \\*
         & \ResultMethod{Rectified Flow} & \MeanOnly{0.129}{0.251} & \MeanOnly{-0.160}{0.212} & \MeanOnly{-0.020}{0.400} & \MeanOnly{-0.049}{0.400} & \SecondMeanOnly{0.352}{0.108} & \MeanOnly{0.261}{0.141} & \MeanOnly{-0.197}{0.179} & 0.045 & 0.487 & 0.443 \\*
         & \ResultMethod{Score SDE} & \MeanOnly{0.251}{0.174} & \MeanOnly{-0.088}{0.396} & \MeanOnly{-0.163}{0.250} & \MeanOnly{0.131}{0.296} & \MeanOnly{0.190}{0.221} & \MeanOnly{0.148}{0.270} & \MeanOnly{-0.122}{0.313} & 0.049 & \Second{0.475} & 0.337 \\*
        \addlinespace[1pt]
         & \ResultMethod{NVCode (Ours)} & \BestMeanOnly{0.654}{0.213} & \BestMeanOnly{0.660}{0.209} & \BestMeanOnly{0.342}{0.400} & \BestMeanOnly{0.588}{0.260} & \MeanOnly{-0.012}{0.203} & \MeanOnly{0.182}{0.400} & \BestMeanOnly{0.660}{0.209} & \Best{0.439} & \Best{0.466} & \SecondRightEdge{0.310} \\
\end{NVCodeDatasetTable}

\begin{NVCodeDatasetTable}{}{Generator}{\AffectiveMusic\ (Active)}{tab:app:generator-all-datasets:affective-music}
\IntraBlock{8} & \ResultMethod{cGAN} & \MeanStd{0.329}{0.091} & \MeanStd{0.428}{0.091} & \MeanStd{0.470}{0.069} & \MeanStd{0.356}{0.068} & \MeanStd{0.439}{0.097} & \MeanStd{0.500}{0.102} & \MeanStd{0.301}{0.083} & 0.403 & 0.418 & 0.333 \\*
         & \ResultMethod{Consistency} & \MeanStd{0.301}{0.091} & \MeanStd{0.437}{0.095} & \MeanStd{0.447}{0.065} & \MeanStd{0.366}{0.077} & \MeanStd{0.426}{0.073} & \MeanStd{0.385}{0.088} & \MeanStd{0.364}{0.075} & 0.389 & 0.445 & 0.303 \\*
         & \ResultMethod{CVAE} & \SecondMeanStd{0.398}{0.104} & \MeanStd{0.475}{0.099} & \SecondMeanStd{0.530}{0.083} & \SecondMeanStd{0.411}{0.065} & \SecondMeanStd{0.475}{0.097} & \SecondMeanStd{0.523}{0.108} & \MeanStd{0.376}{0.094} & \Second{0.455} & 0.395 & 0.287 \\*
         & \ResultMethod{DDPM} & \MeanStd{0.336}{0.100} & \MeanStd{0.440}{0.076} & \MeanStd{0.448}{0.096} & \MeanStd{0.370}{0.068} & \MeanStd{0.441}{0.065} & \MeanStd{0.458}{0.088} & \MeanStd{0.350}{0.084} & 0.406 & \Best{0.363} & 0.273 \\*
         & \ResultMethod{EDM} & \MeanStd{0.390}{0.089} & \MeanStd{0.426}{0.080} & \MeanStd{0.515}{0.084} & \MeanStd{0.388}{0.046} & \MeanStd{0.436}{0.077} & \MeanStd{0.482}{0.114} & \MeanStd{0.349}{0.079} & 0.427 & 0.374 & 0.261 \\*
         & \ResultMethod{Rectified Flow} & \MeanStd{0.388}{0.080} & \MeanStd{0.399}{0.090} & \MeanStd{0.479}{0.075} & \MeanStd{0.382}{0.083} & \MeanStd{0.449}{0.075} & \MeanStd{0.492}{0.092} & \MeanStd{0.350}{0.076} & 0.420 & 0.377 & 0.265 \\*
         & \ResultMethod{Score SDE} & \MeanStd{0.357}{0.089} & \SecondMeanStd{0.490}{0.099} & \BestMeanStd{0.532}{0.092} & \MeanStd{0.395}{0.062} & \MeanStd{0.470}{0.090} & \MeanStd{0.497}{0.104} & \BestMeanStd{0.402}{0.082} & 0.449 & \Second{0.364} & \SecondRightEdge{0.245} \\*
        \addlinespace[1pt]
         & \ResultMethod{NVCode (Ours)} & \BestMeanStd{0.405}{0.084} & \BestMeanStd{0.542}{0.100} & \MeanStd{0.496}{0.113} & \BestMeanStd{0.442}{0.069} & \BestMeanStd{0.491}{0.098} & \BestMeanStd{0.543}{0.120} & \SecondMeanStd{0.381}{0.121} & \Best{0.471} & 0.395 & \BestRightEdge{0.226} \\
\ResultBlockRule
\InterBlock{8} & \ResultMethod{cGAN} & \MeanStd{0.365}{0.065} & \MeanStd{0.413}{0.058} & \MeanStd{0.345}{0.069} & \MeanStd{0.322}{0.073} & \SecondMeanStd{0.359}{0.066} & \MeanStd{0.321}{0.073} & \MeanStd{0.293}{0.079} & 0.346 & 0.371 & 0.204 \\*
         & \ResultMethod{Consistency} & \SecondMeanStd{0.384}{0.070} & \BestMeanStd{0.474}{0.058} & \MeanStd{0.429}{0.064} & \MeanStd{0.412}{0.066} & \MeanStd{0.331}{0.081} & \BestMeanStd{0.366}{0.074} & \MeanStd{0.349}{0.077} & \Second{0.392} & 0.405 & 0.178 \\*
         & \ResultMethod{CVAE} & \MeanStd{0.323}{0.065} & \MeanStd{0.451}{0.048} & \MeanStd{0.304}{0.069} & \MeanStd{0.294}{0.071} & \MeanStd{0.288}{0.072} & \SecondMeanStd{0.365}{0.058} & \MeanStd{0.325}{0.065} & 0.336 & \Best{0.332} & \BestRightEdge{0.115} \\*
         & \ResultMethod{DDPM} & \MeanStd{0.326}{0.074} & \SecondMeanStd{0.461}{0.054} & \SecondMeanStd{0.456}{0.055} & \MeanStd{0.413}{0.060} & \MeanStd{0.311}{0.078} & \MeanStd{0.354}{0.069} & \MeanStd{0.287}{0.083} & 0.373 & 0.336 & 0.137 \\*
         & \ResultMethod{EDM} & \MeanStd{0.351}{0.078} & \MeanStd{0.457}{0.061} & \MeanStd{0.448}{0.062} & \SecondMeanStd{0.435}{0.064} & \BestMeanStd{0.371}{0.074} & \MeanStd{0.290}{0.092} & \BestMeanStd{0.394}{0.070} & \Best{0.392} & \Second{0.335} & 0.158 \\*
         & \ResultMethod{Rectified Flow} & \MeanStd{0.333}{0.056} & \MeanStd{0.351}{0.053} & \MeanStd{0.263}{0.069} & \MeanStd{0.269}{0.067} & \MeanStd{0.242}{0.074} & \MeanStd{0.278}{0.065} & \MeanStd{0.263}{0.068} & 0.285 & 0.345 & \SecondRightEdge{0.136} \\*
         & \ResultMethod{Score SDE} & \MeanStd{0.301}{0.073} & \MeanStd{0.456}{0.050} & \MeanStd{0.327}{0.068} & \MeanStd{0.278}{0.078} & \MeanStd{0.242}{0.088} & \MeanStd{0.304}{0.072} & \MeanStd{0.321}{0.069} & 0.318 & 0.336 & 0.147 \\*
        \addlinespace[1pt]
         & \ResultMethod{NVCode (Ours)} & \BestMeanStd{0.406}{0.094} & \MeanStd{0.366}{0.130} & \BestMeanStd{0.494}{0.059} & \BestMeanStd{0.451}{0.072} & \MeanStd{0.266}{0.187} & \MeanStd{0.331}{0.194} & \SecondMeanStd{0.389}{0.108} & 0.386 & 0.351 & 0.167 \\
\end{NVCodeDatasetTable}

\begin{NVCodeDatasetTable}{}{Generator}{\ExperienceSampling\ (Active)}{tab:app:generator-all-datasets:experience-sampling}
\IntraBlock{8} & \ResultMethod{cGAN} & \MeanStd{0.264}{0.073} & \MeanStd{0.377}{0.090} & \MeanStd{0.286}{0.081} & \MeanStd{0.304}{0.086} & \MeanStd{0.332}{0.089} & \MeanStd{0.334}{0.088} & \MeanStd{0.375}{0.087} & 0.325 & 0.464 & 0.502 \\*
         & \ResultMethod{Consistency} & \SecondMeanStd{0.441}{0.069} & \MeanStd{0.386}{0.074} & \SecondMeanStd{0.385}{0.078} & \MeanStd{0.326}{0.076} & \MeanStd{0.383}{0.085} & \MeanStd{0.386}{0.089} & \MeanStd{0.335}{0.087} & \Second{0.378} & 0.438 & 0.340 \\*
         & \ResultMethod{CVAE} & \MeanStd{0.327}{0.088} & \SecondMeanStd{0.420}{0.093} & \MeanStd{0.362}{0.088} & \MeanStd{0.314}{0.093} & \MeanStd{0.403}{0.080} & \MeanStd{0.379}{0.102} & \MeanStd{0.351}{0.092} & 0.365 & 0.370 & 0.347 \\*
         & \ResultMethod{DDPM} & \MeanStd{0.364}{0.073} & \MeanStd{0.363}{0.079} & \BestMeanStd{0.411}{0.062} & \BestMeanStd{0.372}{0.092} & \MeanStd{0.347}{0.082} & \MeanStd{0.358}{0.092} & \MeanStd{0.385}{0.081} & 0.371 & \Second{0.341} & \SecondRightEdge{0.311} \\*
         & \ResultMethod{EDM} & \MeanStd{0.396}{0.084} & \MeanStd{0.359}{0.082} & \MeanStd{0.370}{0.078} & \MeanStd{0.302}{0.080} & \MeanStd{0.318}{0.074} & \SecondMeanStd{0.432}{0.088} & \SecondMeanStd{0.390}{0.081} & 0.367 & 0.346 & 0.330 \\*
         & \ResultMethod{Rectified Flow} & \MeanStd{0.305}{0.085} & \MeanStd{0.394}{0.088} & \MeanStd{0.365}{0.073} & \SecondMeanStd{0.368}{0.092} & \MeanStd{0.405}{0.067} & \MeanStd{0.387}{0.102} & \MeanStd{0.354}{0.077} & 0.368 & \Best{0.340} & 0.357 \\*
         & \ResultMethod{Score SDE} & \MeanStd{0.395}{0.078} & \MeanStd{0.393}{0.080} & \MeanStd{0.325}{0.087} & \MeanStd{0.360}{0.087} & \SecondMeanStd{0.415}{0.086} & \MeanStd{0.408}{0.097} & \MeanStd{0.345}{0.080} & 0.377 & 0.351 & 0.349 \\*
        \addlinespace[1pt]
         & \ResultMethod{NVCode (Ours)} & \BestMeanStd{0.452}{0.117} & \BestMeanStd{0.527}{0.119} & \MeanStd{0.306}{0.089} & \MeanStd{0.363}{0.108} & \BestMeanStd{0.478}{0.107} & \BestMeanStd{0.447}{0.106} & \BestMeanStd{0.478}{0.113} & \Best{0.436} & 0.351 & \BestRightEdge{0.274} \\
\ResultBlockRule
\InterBlock{8} & \ResultMethod{cGAN} & \MeanStd{-0.071}{0.085} & \MeanStd{-0.046}{0.110} & \MeanStd{0.012}{0.180} & \MeanStd{-0.074}{0.083} & \MeanStd{-0.021}{0.153} & \MeanStd{0.022}{0.151} & \MeanStd{-0.009}{0.192} & -0.027 & 0.345 & 0.214 \\*
         & \ResultMethod{Consistency} & \MeanStd{-0.003}{0.131} & \MeanStd{-0.020}{0.093} & \MeanStd{-0.030}{0.080} & \MeanStd{-0.083}{0.045} & \MeanStd{0.023}{0.089} & \MeanStd{-0.013}{0.106} & \MeanStd{-0.014}{0.104} & -0.020 & 0.417 & 0.119 \\*
         & \ResultMethod{CVAE} & \MeanStd{-0.017}{0.122} & \SecondMeanStd{0.009}{0.142} & \SecondMeanStd{0.016}{0.125} & \MeanStd{-0.043}{0.084} & \MeanStd{-0.007}{0.148} & \MeanStd{0.046}{0.081} & \MeanStd{0.015}{0.126} & \Second{0.003} & \Best{0.328} & 0.220 \\*
         & \ResultMethod{DDPM} & \SecondMeanStd{0.014}{0.122} & \MeanStd{0.002}{0.157} & \MeanStd{-0.012}{0.127} & \SecondMeanStd{-0.028}{0.097} & \MeanStd{-0.056}{0.068} & \MeanStd{-0.044}{0.078} & \MeanStd{-0.034}{0.089} & -0.023 & 0.331 & \BestRightEdge{0.066} \\*
         & \ResultMethod{EDM} & \MeanStd{-0.213}{0.036} & \MeanStd{0.000}{0.227} & \MeanStd{0.005}{0.204} & \MeanStd{-0.123}{0.056} & \SecondMeanStd{0.078}{0.078} & \MeanStd{-0.066}{0.086} & \SecondMeanStd{0.033}{0.125} & -0.041 & 0.336 & 0.221 \\*
         & \ResultMethod{Rectified Flow} & \MeanStd{-0.031}{0.121} & \MeanStd{-0.059}{0.086} & \MeanStd{-0.025}{0.131} & \MeanStd{-0.097}{0.062} & \MeanStd{-0.007}{0.182} & \MeanStd{0.012}{0.163} & \MeanStd{-0.064}{0.082} & -0.039 & 0.344 & 0.281 \\*
         & \ResultMethod{Score SDE} & \MeanStd{-0.092}{0.076} & \MeanStd{-0.034}{0.135} & \MeanStd{-0.000}{0.249} & \MeanStd{-0.136}{0.057} & \MeanStd{0.038}{0.128} & \SecondMeanStd{0.065}{0.095} & \MeanStd{-0.023}{0.157} & -0.026 & \Second{0.329} & 0.183 \\*
        \addlinespace[1pt]
         & \ResultMethod{NVCode (Ours)} & \BestMeanStd{0.235}{0.126} & \BestMeanStd{0.348}{0.232} & \BestMeanStd{0.342}{0.256} & \BestMeanStd{0.378}{0.160} & \BestMeanStd{0.575}{0.053} & \BestMeanStd{0.423}{0.111} & \BestMeanStd{0.440}{0.098} & \Best{0.392} & 0.332 & \SecondRightEdge{0.066} \\
\end{NVCodeDatasetTable}

\begin{NVCodeDatasetTable}{}{Generator}{NatView (Active)}{tab:app:generator-all-datasets:natview}
\IntraBlock{8} & \ResultMethod{cGAN} & \MeanOnly{-0.445}{0.315} & \MeanOnly{0.004}{0.996} & \MeanOnly{0.089}{0.911} & \MeanOnly{0.240}{0.760} & \MeanOnly{0.222}{0.208} & \MeanOnly{-0.207}{0.793} & \MeanOnly{0.387}{0.613} & 0.041 & 0.441 & 0.760 \\*
         & \ResultMethod{Consistency} & \MeanOnly{-0.313}{0.687} & \MeanOnly{-0.155}{0.845} & \SecondMeanOnly{0.226}{0.774} & \MeanOnly{0.167}{0.833} & \SecondMeanOnly{0.228}{0.772} & \MeanOnly{0.008}{0.668} & \SecondMeanOnly{0.605}{0.395} & 0.109 & 0.388 & 0.571 \\*
         & \ResultMethod{CVAE} & \MeanOnly{-0.300}{0.700} & \BestMeanOnly{0.279}{0.721} & \BestMeanOnly{0.360}{0.640} & \SecondMeanOnly{0.672}{0.328} & \BestMeanOnly{0.502}{0.498} & \MeanOnly{0.216}{0.784} & \BestMeanOnly{0.674}{0.326} & \Best{0.343} & 0.323 & 0.502 \\*
         & \ResultMethod{DDPM} & \MeanOnly{-0.555}{0.445} & \MeanOnly{-0.828}{0.172} & \MeanOnly{0.105}{0.895} & \MeanOnly{-0.330}{0.670} & \MeanOnly{-0.476}{0.524} & \MeanOnly{-0.250}{0.750} & \MeanOnly{0.349}{0.651} & -0.284 & \Best{0.315} & \BestRightEdge{0.362} \\*
         & \ResultMethod{EDM} & \MeanOnly{-0.459}{0.435} & \MeanOnly{-0.449}{0.551} & \MeanOnly{-0.136}{0.864} & \MeanOnly{-0.204}{0.796} & \MeanOnly{0.172}{0.828} & \SecondMeanOnly{0.231}{0.769} & \MeanOnly{0.028}{0.972} & -0.117 & \Second{0.320} & 0.489 \\*
         & \ResultMethod{Rectified Flow} & \SecondMeanOnly{-0.294}{0.706} & \MeanOnly{-0.616}{0.384} & \MeanOnly{-0.335}{0.665} & \MeanOnly{0.163}{0.837} & \MeanOnly{-0.437}{0.563} & \MeanOnly{-0.291}{0.709} & \MeanOnly{-0.793}{0.207} & -0.372 & 0.327 & 0.455 \\*
         & \ResultMethod{Score SDE} & \MeanOnly{-0.441}{0.559} & \MeanOnly{-0.523}{0.477} & \MeanOnly{0.109}{0.233} & \MeanOnly{0.051}{0.949} & \MeanOnly{0.199}{0.267} & \MeanOnly{-0.080}{0.920} & \MeanOnly{-0.521}{0.479} & -0.172 & 0.320 & 0.564 \\*
        \addlinespace[1pt]
         & \ResultMethod{NVCode (Ours)} & \BestMeanOnly{0.103}{0.897} & \SecondMeanOnly{0.184}{0.816} & \MeanOnly{0.082}{0.918} & \BestMeanOnly{0.816}{0.184} & \MeanOnly{0.194}{0.806} & \BestMeanOnly{0.301}{0.699} & \MeanOnly{0.518}{0.482} & \Second{0.314} & 0.368 & \SecondRightEdge{0.385} \\
\ResultBlockRule
\InterBlock{8} & \ResultMethod{cGAN} & \MeanOnly{-0.638}{0.188} & \SecondMeanOnly{0.474}{0.248} & \MeanOnly{-0.580}{0.206} & \MeanOnly{-0.325}{0.349} & \MeanOnly{-0.339}{0.336} & \MeanOnly{-0.646}{0.186} & \MeanOnly{-0.112}{0.400} & -0.309 & 0.409 & 0.687 \\*
         & \ResultMethod{Consistency} & \BestMeanOnly{0.567}{0.400} & \MeanOnly{-0.451}{0.400} & \MeanOnly{0.317}{0.400} & \BestMeanOnly{0.510}{0.400} & \MeanOnly{0.105}{0.400} & \SecondMeanOnly{0.405}{0.400} & \MeanOnly{0.143}{0.400} & 0.228 & 0.377 & 0.360 \\*
         & \ResultMethod{CVAE} & \MeanOnly{-0.097}{0.400} & \MeanOnly{0.261}{0.400} & \BestMeanOnly{0.641}{0.193} & \MeanOnly{-0.015}{0.400} & \MeanOnly{-0.050}{0.400} & \MeanOnly{0.151}{0.400} & \SecondMeanOnly{0.628}{0.197} & 0.217 & 0.305 & 0.315 \\*
         & \ResultMethod{DDPM} & \SecondMeanOnly{0.248}{0.400} & \MeanOnly{-0.527}{0.400} & \MeanOnly{-0.481}{0.400} & \MeanOnly{-0.674}{0.326} & \MeanOnly{-0.153}{0.400} & \MeanOnly{-0.435}{0.400} & \MeanOnly{-0.705}{0.295} & -0.390 & 0.307 & 0.508 \\*
         & \ResultMethod{EDM} & \MeanOnly{-0.121}{0.400} & \MeanOnly{0.382}{0.400} & \MeanOnly{0.418}{0.383} & \MeanOnly{-0.123}{0.400} & \BestMeanOnly{0.480}{0.337} & \MeanOnly{-0.098}{0.400} & \BestMeanOnly{0.720}{0.231} & \Second{0.237} & \Second{0.294} & 0.419 \\*
         & \ResultMethod{Rectified Flow} & \MeanOnly{-0.508}{0.235} & \MeanOnly{0.399}{0.293} & \MeanOnly{-0.037}{0.400} & \MeanOnly{-0.038}{0.400} & \MeanOnly{0.331}{0.347} & \BestMeanOnly{0.409}{0.286} & \MeanOnly{0.042}{0.400} & 0.086 & 0.309 & 0.463 \\*
         & \ResultMethod{Score SDE} & \MeanOnly{-0.529}{0.400} & \BestMeanOnly{0.508}{0.400} & \SecondMeanOnly{0.498}{0.400} & \SecondMeanOnly{0.088}{0.400} & \SecondMeanOnly{0.445}{0.400} & \MeanOnly{0.269}{0.400} & \MeanOnly{0.561}{0.400} & \Best{0.263} & \Best{0.273} & \BestRightEdge{0.214} \\*
        \addlinespace[1pt]
         & \ResultMethod{NVCode (Ours)} & \MeanOnly{0.127}{0.362} & \MeanOnly{0.274}{0.357} & \MeanOnly{0.394}{0.174} & \MeanOnly{-0.003}{0.168} & \MeanOnly{0.235}{0.400} & \MeanOnly{0.177}{0.400} & \MeanOnly{0.369}{0.195} & 0.225 & 0.311 & \SecondRightEdge{0.249} \\
\end{NVCodeDatasetTable}

\endgroup

\subsection{Scan Level EEG--fMRI Evidence}
\label{app:additional:scan-atlases}

\begin{figure}[p]
    \centering
    \includegraphics[width=.95\linewidth,height=.945\textheight,keepaspectratio]{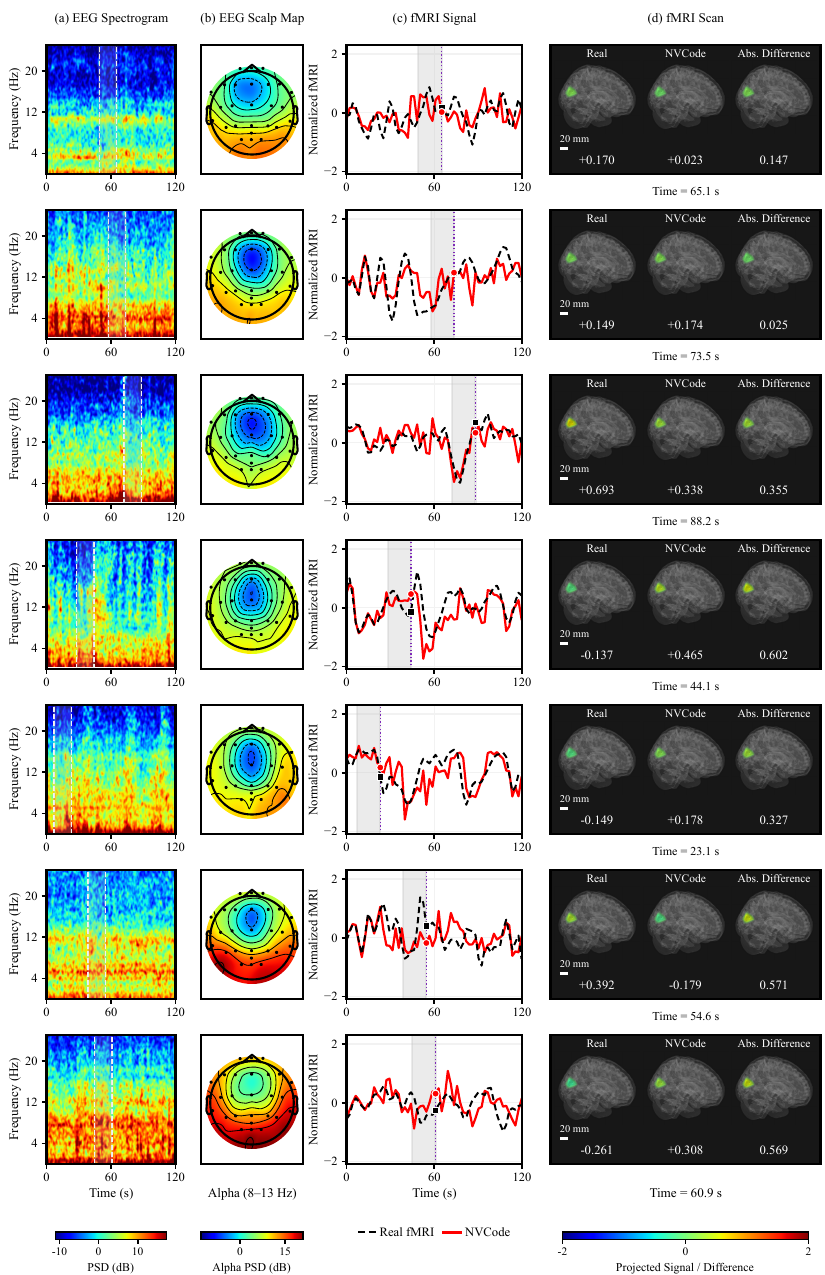}
    \captionsetup{width=.95\linewidth,justification=justified}
    \caption{\textbf{Cuneus} examples. (a) EEG spectrogram; (b) alpha scalp power; (c) real fMRI (black dashed) and NVCode (red); (d) projected real, predicted and absolute-difference amplitudes.}
    \label{fig:app-temporal-cuneus}
\end{figure}

\begin{figure}[p]
    \centering
    \includegraphics[width=.95\linewidth,height=.945\textheight,keepaspectratio]{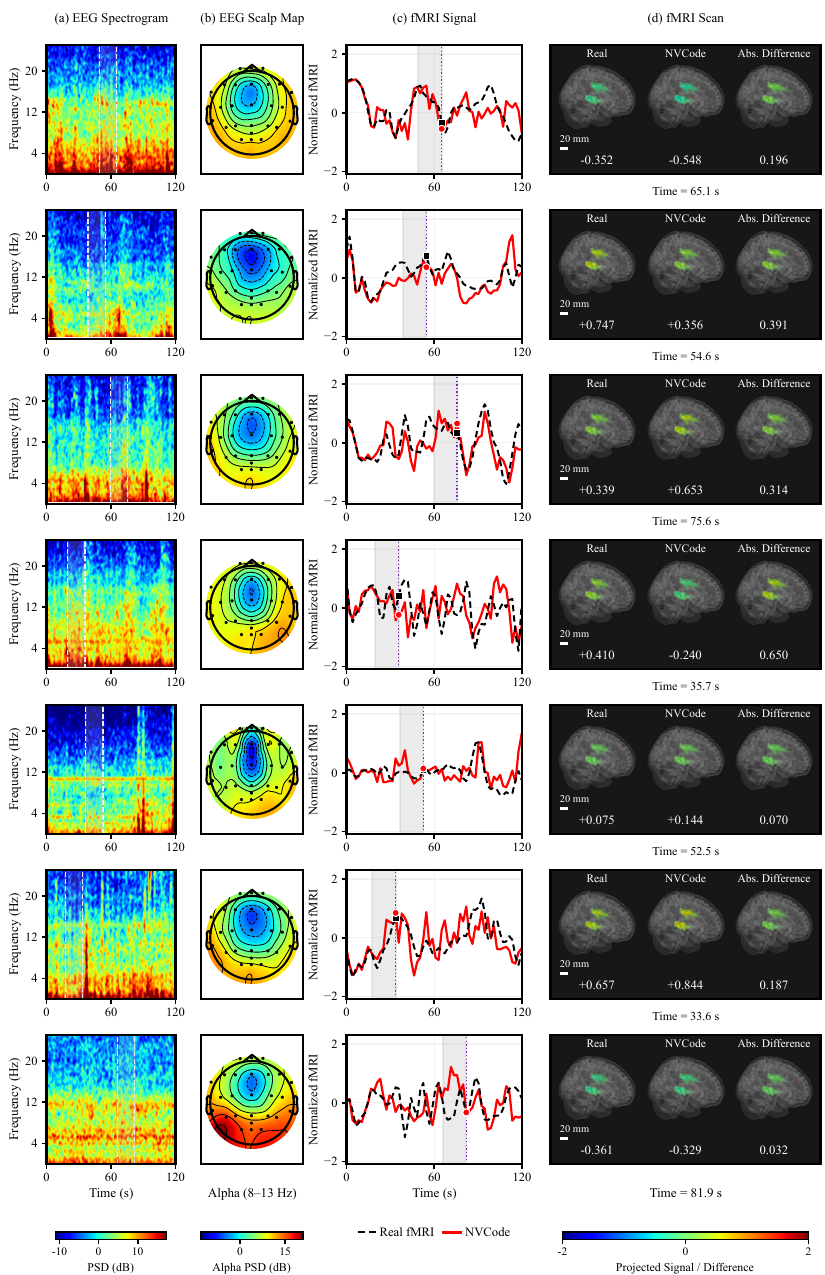}
    \captionsetup{width=.95\linewidth,justification=justified}
    \caption{\textbf{Heschl's gyrus} examples. (a) EEG spectrogram; (b) alpha scalp power; (c) real fMRI (black dashed) and NVCode (red); (d) projected real, predicted and absolute-difference amplitudes.}
    \label{fig:app-temporal-heschls-gyrus}
\end{figure}

\begin{figure}[p]
    \centering
    \includegraphics[width=.95\linewidth,height=.945\textheight,keepaspectratio]{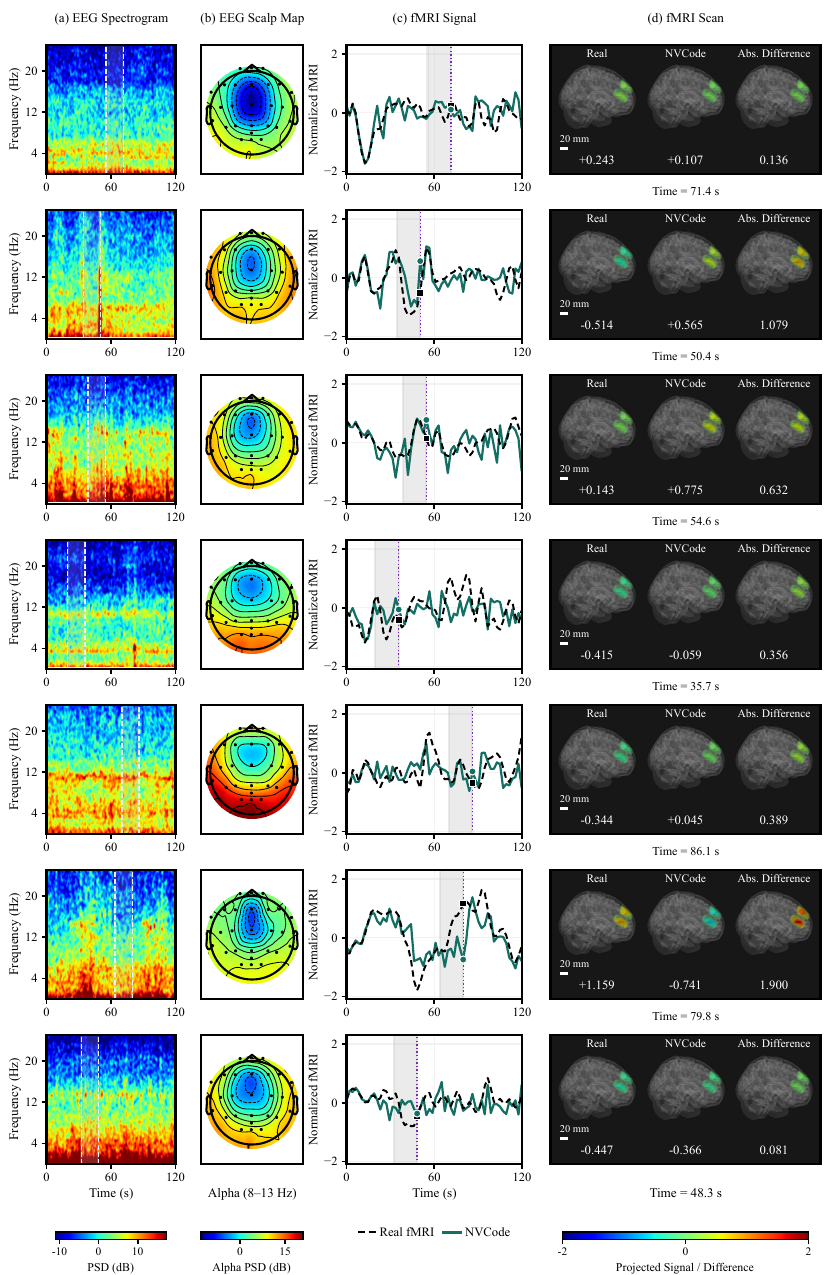}
    \captionsetup{width=.95\linewidth,justification=justified}
    \caption{\textbf{Middle frontal gyrus} examples. (a) EEG spectrogram; (b) alpha scalp power; (c) real fMRI (black dashed) and NVCode (green); (d) projected real, predicted and absolute-difference amplitudes.}
    \label{fig:app-temporal-middle-frontal-gyrus}
\end{figure}

\begin{figure}[p]
    \centering
    \includegraphics[width=.95\linewidth,height=.945\textheight,keepaspectratio]{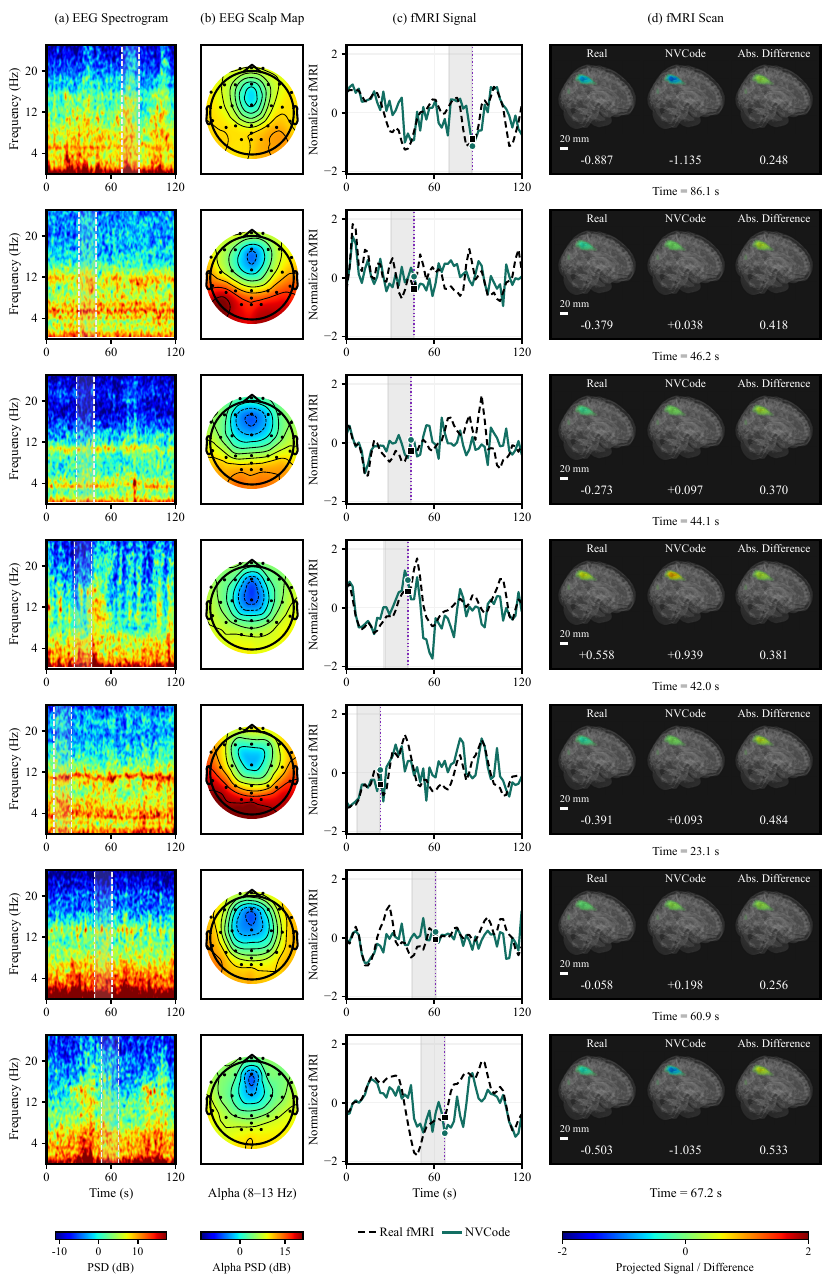}
    \captionsetup{width=.95\linewidth,justification=justified}
    \caption{\textbf{Precuneus} examples. (a) EEG spectrogram; (b) alpha scalp power; (c) real fMRI (black dashed) and NVCode (green); (d) projected real, predicted and absolute-difference amplitudes.}
    \label{fig:app-temporal-precuneus}
\end{figure}

\begin{figure}[p]
    \centering
    \includegraphics[width=.95\linewidth,height=.945\textheight,keepaspectratio]{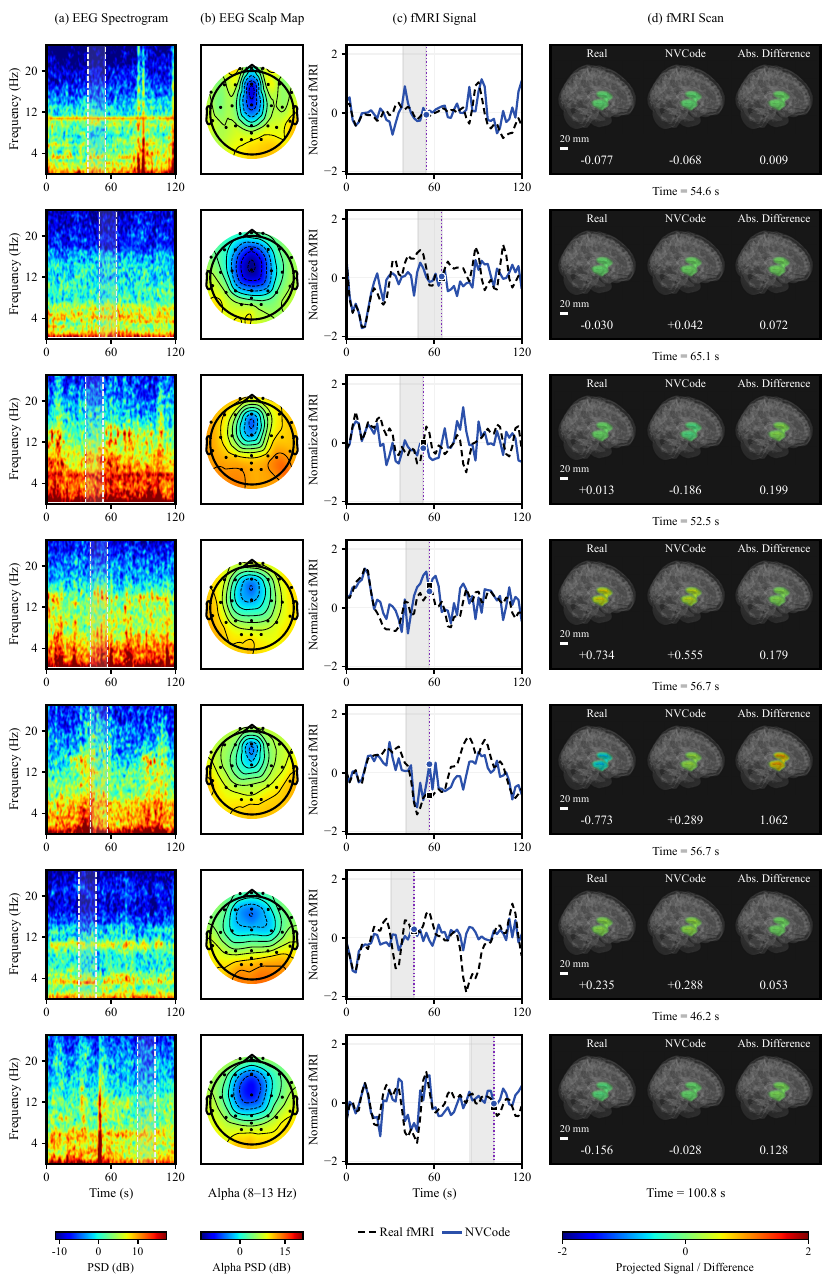}
    \captionsetup{width=.95\linewidth,justification=justified}
    \caption{\textbf{Putamen} examples. (a) EEG spectrogram; (b) alpha scalp power; (c) real fMRI (black dashed) and NVCode (blue); (d) projected real, predicted and absolute-difference amplitudes.}
    \label{fig:app-temporal-putamen}
\end{figure}

\begin{figure}[p]
    \centering
    \includegraphics[width=.95\linewidth,height=.945\textheight,keepaspectratio]{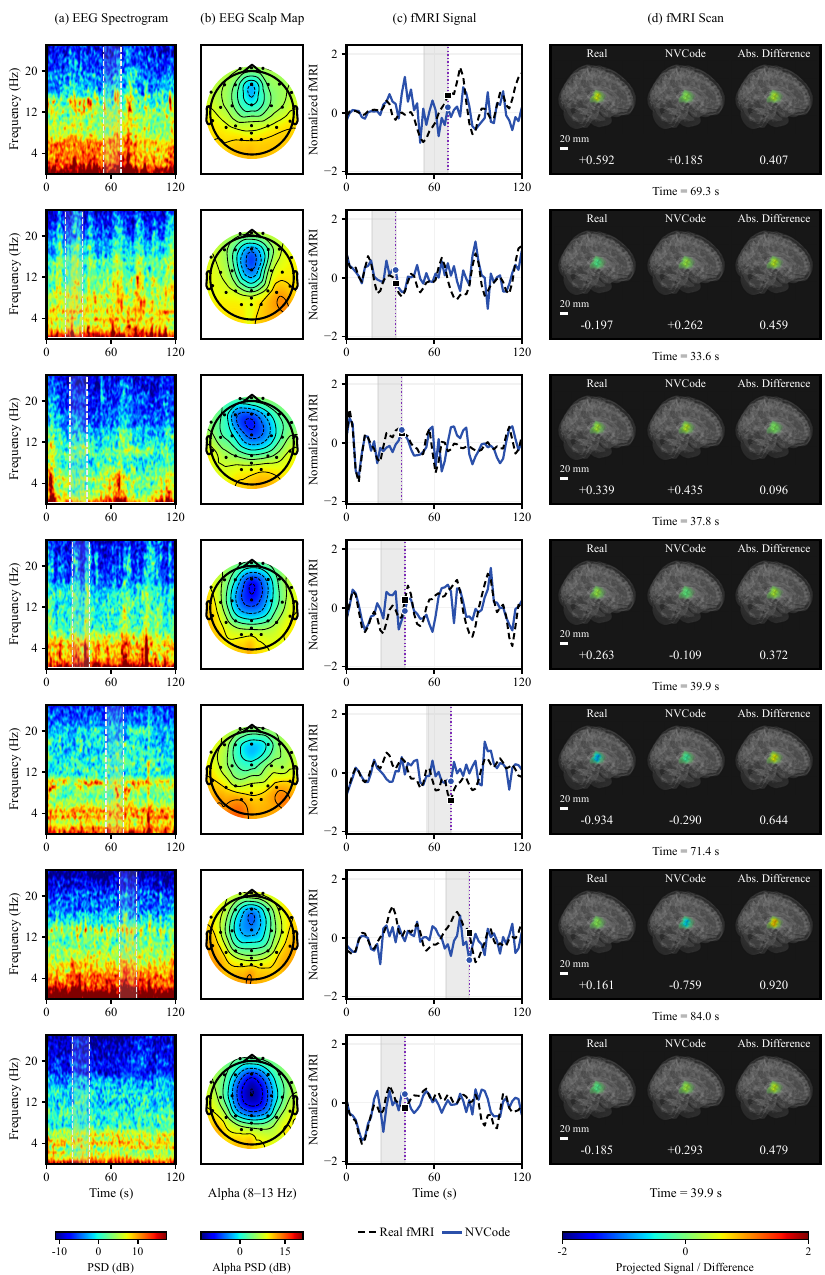}
    \captionsetup{width=.95\linewidth,justification=justified}
    \caption{\textbf{Thalamus} examples. (a) EEG spectrogram; (b) alpha scalp power; (c) real fMRI (black dashed) and NVCode (blue); (d) projected real, predicted and absolute-difference amplitudes.}
    \label{fig:app-temporal-thalamus}
\end{figure}

\begin{figure}[p]
    \centering
    \includegraphics[width=.95\linewidth,height=.945\textheight,keepaspectratio]{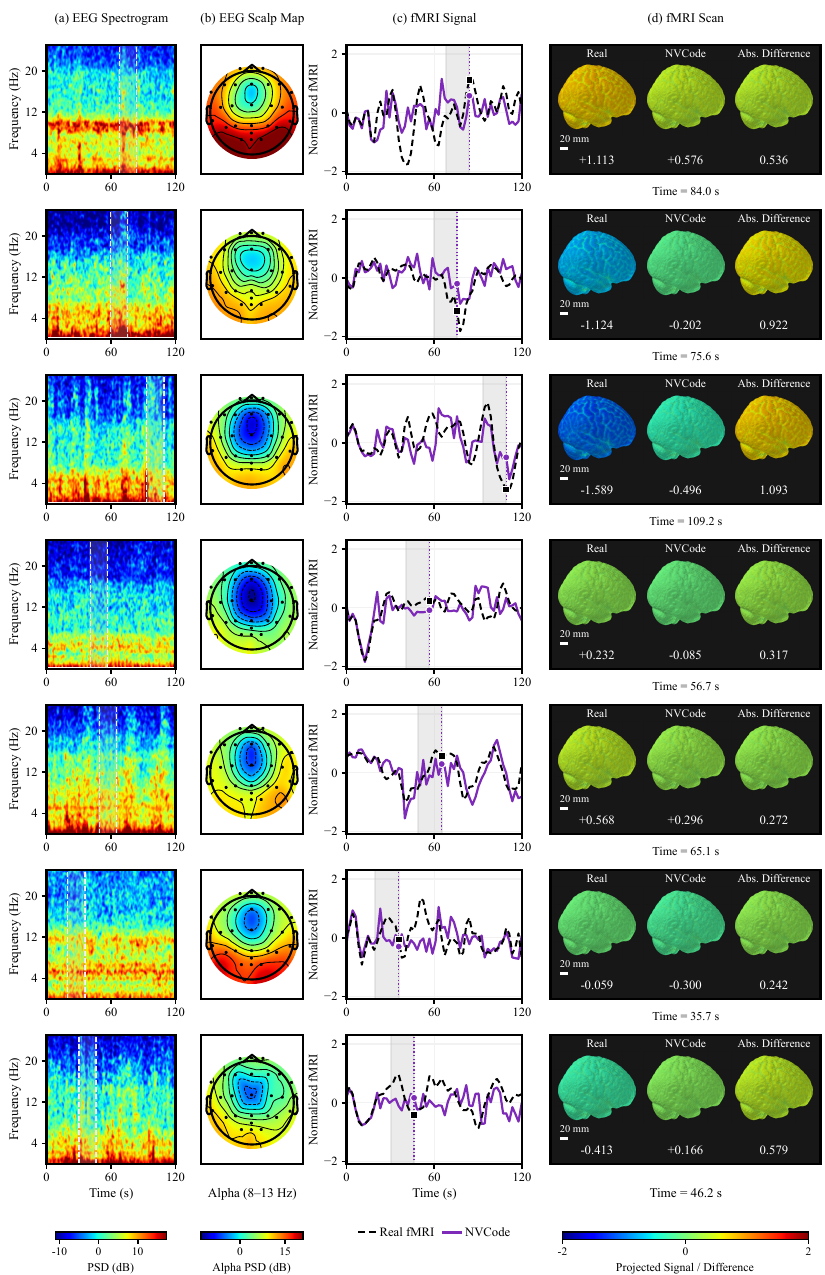}
    \captionsetup{width=.95\linewidth,justification=justified}
    \caption{\textbf{Global signal} examples. (a) EEG spectrogram; (b) alpha scalp power; (c) real fMRI (black dashed) and NVCode (purple); (d) whole-brain projections of real, predicted and absolute-difference amplitudes.}
    \label{fig:app-temporal-global-signal}
\end{figure}

Figures~\ref{fig:app-temporal-cuneus}--\ref{fig:app-temporal-global-signal} show seven scans per target. Columns (a) and (b) show the EEG spectrogram and alpha scalp power ($8$--$13$\,Hz)~\citep{buzsaki2012brain,cohen2017does}; (c) overlays the observed and predicted fMRI trajectories~\citep{vidaurre2018discovering}. The purple guide marks acquisition $t_\star$ in (d) and the end of the shaded input EEG window $[t_\star-16\,\mathrm{s},t_\star)$.

\paragraph{Snapshot amplitudes and normalization.}
Let $q_{s,n}(t)$ be the target signal after forward and backward fifth-order Butterworth filtering at $0.15$\,Hz. The display scale is
\begin{equation}
\label{eq:app-snapshot-normalization}
\begin{gathered}
\bar q_{s,n}=\frac{1}{T_s}\sum_{j=1}^{T_s}q_{s,n}(t_j),\\
y_{s,n}(t)=\frac{q_{s,n}(t)-\bar q_{s,n}}
{Q_{0.95}\!\left(\{|q_{s,n}(t_j)-\bar q_{s,n}|\}_{j=1}^{T_s}\right)+\epsilon}.
\end{gathered}
\end{equation}
Here $T_s$ counts source volumes, $Q_{0.95}$ is the empirical $95$th percentile, and $\epsilon=10^{-8}$. For denominator $b_{s,n}>0$, zero is the filtered scan mean and one display unit equals $b_{s,n}$ filtered units. Trajectories and snapshots share this affine scale within a row. Benchmark scores use the split-specific scale in Section~\ref{app:exp:data}.

\paragraph{Real, predicted, and absolute-difference values.}
At $t_\star$, the labels beneath the views give the observed amplitude, the mean of $K=50$ generated samples, and their absolute difference on this scale:
\begin{equation}
\label{eq:app-snapshot-values}
\begin{gathered}
a_{s,n}=y_{s,n}(t_\star),\quad
\widehat a_{s,n}=\frac{1}{K}\sum_{k=1}^{K}y^{(k)}_{s,n}(t_\star),\\
e_{s,n}=\bigl|a_{s,n}-\widehat a_{s,n}\bigr|.
\end{gathered}
\end{equation}
The filtered error is $\big|b_{s,n}(a_{s,n}-\widehat a_{s,n})\big|=b_{s,n}e_{s,n}$ because the shared offset cancels. The same dimensionless error can represent different changes in filtered units across scans. Draw sets with the same mean yield identical point prediction panels regardless of spread.

\paragraph{Anatomical projection and shared color scale.}
Let $\Omega_n$ be the display mask and $W_n(v)$ its nonnegative weights, from a DiFuMo component for a region or anatomical intensity for the global signal. The peak-normalized weights are
\begin{equation}
\label{eq:app-snapshot-weights}
w_n(v)=\begin{cases}
W_n(v)\big/\displaystyle\max_{u\in\Omega_n}W_n(u), & v\in\Omega_n,\\
0, & v\notin\Omega_n.
\end{cases}
\end{equation}
Apply $w_n$ to both amplitudes and their absolute difference:
\begin{equation}
\label{eq:app-snapshot-projection}
\begin{aligned}
I^{\mathrm{real}}_{s,n}(v)&=a_{s,n}w_n(v),\\
I^{\mathrm{NVCode}}_{s,n}(v)&=\widehat a_{s,n}w_n(v),\\
D_{s,n}(v)&=\bigl|I^{\mathrm{real}}_{s,n}(v)-I^{\mathrm{NVCode}}_{s,n}(v)\bigr|\\
&=e_{s,n}w_n(v).
\end{aligned}
\end{equation}
Because $w_n(v)\geq0$ and $\max_{v\in\Omega_n}w_n(v)=1$, the projected error peaks at $e_{s,n}$ before rendering. Equal amplitudes give zero everywhere; opposite signs give the sum of their magnitudes.

\paragraph{Reading the spatial views.}
Anatomy supplies the spatial pattern in the plates and Figure~\ref{fig:nvcode-pipeline}(e); NVCode supplies one scalar per target and time. Regional displays use weight levels $0.1$, $0.3$, $0.5$, $0.7$, and $0.9$ on transparent MNI152 anatomy; global surface colors average weights $2$, $4$, and $6$\,mm inward. The three views share geometry, orthographic view, shading, and opacity. Bars show $20$\,mm and colors span $[-2,2]$, with errors on the nonnegative half. Numbers give unprojected values; colors encode local weighted values. The footprint shows the mask and scale, not estimated voxelwise activation.

\paragraph{Reading the trajectories.}
The guide compares amplitudes at one time; Pearson $R$ compares aligned fluctuations across the scan. For nonconstant observed and predicted amplitude vectors $\mathbf{a}$ and $\widehat{\mathbf{a}}$, an additive shift or positive rescaling can change amplitudes while preserving correlation:
$R(\mathbf{a},\gamma\widehat{\mathbf{a}}+\beta\mathbf{1})=R(\mathbf{a},\widehat{\mathbf{a}})$ for $\gamma>0$.
Centering removes $\beta$, and $\gamma$ cancels between covariance and predicted standard deviation. A trace can track peaks and troughs but miss the guide, or match it without tracking the rest of the scan.

\clearpage

% Finish the atlas before the experiment details.
\FloatBarrier

\section{Experiment Details}
\label{app:experiment}

This appendix specifies the data construction, baseline adaptation, metric formulas, downstream utility probe and ablation protocol used in Section~\ref{sec:experiments}.

\subsection{Experiment Setup: Data Preprocessing and Machine Requirements}
\label{app:setup}
\label{sec:exp-setup}
\label{app:exp-setup}
\label{app:exp:data}
\label{app:exp:alignment}
\label{app:exp:signal}
\label{app:exp:norm}
\label{app:exp:compute}
\label{app:exp:software}

\myparagraph{Datasets and splits.}
Table~\ref{tab:datasets} summarizes eight public simultaneous EEG--fMRI datasets~\citep{huster2012methods,murta2015electrophysiological}, comprising four resting-state~\citep{biswal1995functional,fox2007spontaneous} and four task-evoked~\citep{laufs2003eeg,murta2015electrophysiological} settings. Their source cohorts report $184$ participants, while the $189$ scans retained after scan-level exclusions yield $61{,}646$ paired volumes. Each fMRI volume is paired with the immediately preceding $16$-s EEG window, so future EEG does not enter the model input. Intra-subject evaluation uses $80\%$ of volumes for training and $10\%$ for validation. Test volumes follow a $\lceil20/\mathrm{TR}\rceil$-volume buffer that separates them from validation and prevents overlap between their EEG contexts~\citep{logothetis2008we}. Inter-subject evaluation~\citep{lotte2018review} uses disjoint training and held-out subjects. Normalization statistics and checkpoint selection use only the training-subject partition.

\myparagraph{Target signals, baselines and metrics.}
Each volume defines a seven-dimensional target vector containing six DiFuMo ROI signals and one global signal~\citep{dadi2020fine,murphy2009impact}. Figure~\ref{fig:nvcode-pipeline}(b) lists these targets: Cuneus, Heschl's gyrus (H Gyrus), anterior middle frontal gyrus (MF Gyrus), anterior precuneus, putamen, thalamus, and the global signal. Every model predicts this same ordered vector from the same retained EEG--fMRI pairs and training-only normalization statistics. In Table~\ref{tab:main-encoder}, the generator architecture, sampler, target ordering, and split policy are held fixed while the EEG encoder configuration changes. In Table~\ref{tab:main-generator}, the seven frozen NVCode encoders and fitted projector supply identical conditions while the downstream generation network changes. Stochastic generators return $K{=}50$ samples for each held-out EEG epoch. Avg.~$R$ and FC-MAE evaluate the predictive mean, whereas mean marginal CRPS uses the complete empirical sample distribution. Dataset scores are first summarized over their internal splits and then macro-averaged with equal dataset weight. Appendix~\ref{app:main-exp} gives the adaptation rules, metric definitions, and aggregation procedure.

\myparagraph{Cohort and scan-count conventions.}
Table~\ref{tab:datasets} distinguishes three levels of data accounting. \emph{Cohort Size} counts participants in the original source cohorts. \emph{Scans} counts runs retained after scan-level exclusions, and \emph{Pairs} counts the volume-anchored EEG--fMRI observations from those runs before splitting. For NatView, the two retained runs provide $571$ paired volumes. NatView contributes one dataset-level score to the benchmark mean, with the same weight as any other dataset.

\myparagraph{Paired EEG to fMRI observations.}
Each observation pairs the fMRI volume acquired at $t_v$ with the preceding EEG interval $[t_v-16\,\mathrm{s}, t_v)$. The recorded onset defines the alignment without an additional hemodynamic lag shift. The repetition time is dataset dependent: $\mathrm{TR}=2.0$\,s for \XPone, \AffectiveMusic, \MultiCLARID, \GradCPT, and \ExperienceSampling, and $\mathrm{TR}=2.1$\,s for \SleepEEGfMRI, NeuroBOLT, and NatView. The fixed window supplies preceding EEG context for estimating the BOLD response at each acquisition time~\citep{logothetis2008we}.

\myparagraph{EEG preprocessing and channel harmonization.}
EEG-informed fMRI analysis combines electrophysiological preprocessing with BOLD modeling~\citep{abreu2018eeg} within broader multimodal brain imaging workflows~\citep{calhoun2016multimodal}.
All EEG segments are resampled to $200$\,Hz with the polyphase finite impulse response resampler in \texttt{mne.io.Raw.resample(..., method="polyphase")}. The default anti aliasing low pass filter is applied at the Nyquist frequency and no time shift is introduced. Each segment is cropped to $T=3200$ samples, and raw amplitudes are divided by $100$ before being fed to the encoders. Native channel counts vary across datasets, so we harmonize all recordings to the following $26$ channels of the 10--20 montage:
\begin{shadedeqstar}
\begin{aligned}
\mathcal{C}_{26}=\{&
\mathrm{FP1},\mathrm{FP2},\mathrm{F3},\mathrm{F4},\mathrm{C3},\mathrm{C4},\mathrm{P3},\mathrm{P4},\mathrm{O1},\mathrm{O2},\\
&\mathrm{F7},\mathrm{F8},\mathrm{T7},\mathrm{T8},\mathrm{P7},\mathrm{P8},
\mathrm{FPZ},\mathrm{FZ},\mathrm{CZ},\mathrm{PZ},\\
&\mathrm{POZ},\mathrm{OZ},\mathrm{FT9},\mathrm{FT10},\mathrm{TP9},\mathrm{TP10}\}.
\end{aligned}
\end{shadedeqstar}
Non EEG channels such as ECG, EOG and EMG are removed, along with carbon wire loop sensors. Missing target channels are reconstructed by spherical spline interpolation against the standard 10--20 montage using MNE Python with \texttt{standard\_1020}. Channels present in the source recording but outside the target set are discarded. Dataset provided referencing and artifact correction decisions are preserved. No dataset specific EEG augmentation is applied.

\myparagraph{fMRI signal construction.}
For each 4D fMRI scan, we extract atlas level signals with nilearn's \texttt{NiftiMapsMasker} using the 64 component DiFuMo atlas~\citep{dadi2020fine} at 2\,mm resolution. The masker is used only for soft mask spatial weighting; masker standardization and detrending are disabled so that all reported statistics are controlled by the split dependent preprocessing below. From the 64 DiFuMo components, we select six regional signals by label matching: Cuneus, Heschl's gyrus, Middle Frontal Gyrus anterior, Precuneus anterior, Putamen and Thalamus. Under \texttt{nilearn}~0.10.3, these correspond to component IDs $\{16,52,46,3,28,30\}$ in the listed order. The seventh signal is the global signal~\citep{murphy2009impact}, computed as the unweighted mean over all 64 DiFuMo components before filtering or scaling. The seven signals are stacked into $\mathbf{r}_{s,v}\in\mathbb{R}^{7}$ for scan $s$ and volume $v$.

\myparagraph{Temporal filtering.}
Target construction uses a fifth-order Butterworth low-pass filter for all seven fMRI signals. The filter is specified at each scan's sampling rate with a cutoff in the low-frequency BOLD range~\citep{cordes2001frequencies,fox2007spontaneous}:
\begin{shadedeqstar}
\begin{gathered}
f_s=1/\mathrm{TR},\\
f_c=0.15\,\mathrm{Hz}.
\end{gathered}
\end{shadedeqstar}
All methods use the same filtered fMRI targets in both evaluation protocols.

\myparagraph{Normalization and split control.}
The per-dataset reconstruction comparisons use two normalizations. The first defines the robust-normalized fMRI label scale used for reported metrics. In the intra-subject protocol, a robust scaler is fitted for each scan $s$ and signal $n$ using only the corresponding temporal training portion:
\begin{shadedeqstar}
\begin{gathered}
m_{s,n}=\mathrm{median}_{v\in\mathcal{I}^{s}_{\mathrm{train}}} r_{s,v,n},
\\
a_{s,n}=\frac{\mathrm{IQR}_{v\in\mathcal{I}^{s}_{\mathrm{train}}} r_{s,v,n}}{1.349}+\epsilon,
\\
 \epsilon=10^{-6}.
\end{gathered}
\end{shadedeqstar}
In the inter subject protocol, the same definitions are evaluated over all volumes from training subjects in dataset $d$, yielding $m_{d,n}$ and $a_{d,n}$. These parameters remain fixed and are applied unchanged to every held out subject in dataset $d$. Let $\rho=s$ for intra subject evaluation and $\rho=d(s)$ for inter subject evaluation, where $d(s)$ denotes the dataset containing scan $s$. The robust-normalized label is
\begin{shadedeqstar}
y_{s,v,n}=\frac{r_{s,v,n}-m_{\rho,n}}{a_{\rho,n}}.
\end{shadedeqstar}
This fixed transformation defines the fMRI supervision and evaluation scale. The second normalization is the split specific generator $z$ score used inside the stochastic interpolant. In the intra subject protocol, the generator mean and standard deviation are fitted on the training rows of the scan. In the inter subject protocol, they are fitted only on training subjects. Generated samples are mapped back to the robust-normalized fMRI scale before metrics are computed.

\myparagraph{Intra subject split.}
For a scan with $V$ fMRI volumes, we form four contiguous temporal portions in order: $\lfloor0.8V\rfloor$ training volumes, $\lfloor0.1V\rfloor$ validation volumes, a $\lceil20/\mathrm{TR}\rceil$ volume buffer and all remaining test volumes. The buffer keeps the test volumes and their $16$\,s EEG contexts separate from the earlier portions. We fit all trainable components on the training portion, select checkpoints on the validation portion and use the final portion for evaluation.

\myparagraph{Inter subject split.}
Inter-subject evaluation groups recordings by subject and tests transfer to held-out people within each dataset. Both normalization layers, model fitting, checkpoint selection, and early stopping use the training-subject partition. Regional representation maps and feature caches are constructed within each training fold; evaluation applies these fixed maps to held-out EEG. The benchmark therefore measures subject transfer across eight simultaneous EEG--fMRI acquisition settings.

\myparagraph{\mbox{Reproducibility controls.}}
The splits stay fixed across the three reconstruction runs, with seeds $42$, $2026$, and $114514$. At the start of each run we set \texttt{torch.manual\_seed(r)} and \texttt{numpy.random.seed(r)} for seed $r$. All methods enable \texttt{torch.use\_deterministic\_algorithms(}\allowbreak\texttt{True)} and disable cuDNN benchmarking and TF32. Training uses fp32 without mixed precision or augmentation.

\myparagraph{Machine requirements and software.}
All experiments were implemented in PyTorch and trained in parallel on four NVIDIA A100 80\,GB or H100 80\,GB GPUs. At inference time, the EEG derived context is computed once for each held out epoch and then shared across all $S\cdot K$ velocity evaluations. With $S=100$ Euler steps and $K=50$ Monte Carlo samples, inference for a typical scan with $50$ to $200$ held out epochs uses one accelerator. The full eight dataset inter subject evaluation requires independent generator fits across subject splits and approximately 20 aggregate GPU days after caching per scan EEG features. The software environment uses PyTorch~2.1, SciPy~1.11, NumPy~1.26, MNE-Python~$\geq$1.7, nilearn~0.10.3, \texttt{linear attention transformer}~0.19.1, \texttt{umap-learn}~0.5~\citep{mcinnes2018umap}, CUDA~12.1 and cuDNN~8.9.

\subsection{Main Experiment Details}
\label{app:main-exp}
\label{app:exp:baselines}
\label{app:exp:metrics}
\label{app:exp:diagnostics}

The complete-model comparison in main-paper Table~\ref{tab:end-to-end} and supplementary Table~\ref{tab:app:end-to-end} pools training partitions across the eight datasets under each protocol and evaluates the same seven outputs. The component comparisons instead fit and evaluate separate models within each dataset.

Tables~\ref{tab:main-encoder} and~\ref{tab:main-generator} compare EEG encoder configurations and conditional generation configurations, respectively. The EEG encoder outputs the feature vector $\mathbf{e}(\mathbf{x})$. The complete generator combines the conditional projector $\Pi_{\psi}$ with the generation network $F_{\theta}$, using $\epsilon$ for sampling randomness:
\begin{shadedeqstar}
G_{\psi,\theta}\!\left(\mathbf{e}(\mathbf{x}),\epsilon\right)
=F_{\theta}\!\left(\Pi_{\psi}(\mathbf{e}(\mathbf{x})),\epsilon\right).
\end{shadedeqstar}
The protocols below specify which generator parameters are trained in each comparison. Table~\ref{tab:joint-flow-ablation} compares eight configurations of target-specific encoding, joint output modeling, and stochastic interpolation.

\myparagraph{EEG encoder baselines.}
Related EEG representation work uses convolutional feature extraction~\citep{lawhern2018eegnet,roy2019deep}, recurrent sleep staging~\citep{supratak2017deepsleepnet}, and Transformer-based sleep staging~\citep{phan2022sleeptransformer}.
All encoder baselines are adapted to the same sequential EEG to fMRI prediction task. In Table~\ref{tab:main-encoder}, each baseline family uses one EEG encoder with a $1400$-dimensional output representation. A temporary fMRI readout supervises this encoder with mean squared error on the seven robust-normalized target signals. Validation MSE selects the checkpoint, after which the temporary readout is discarded and the encoder is frozen. NVCode instead fits seven target-specific encoders, each with a temporary scalar head, and concatenates their $200$-dimensional representations into the same $1400$-dimensional interface. Each row fits a separate NVCode generator with the same architecture, jointly optimizing its projector and velocity field on the frozen encoder features. The projector maps the $1400$ features to a $d_c{=}512$ context. In inter-subject evaluation, all fitting and validation selection occur within the training subjects of each split. The comparison fixes representation width, fMRI targets, split policy, normalization, generator architecture, sampler, and metrics while varying the EEG encoder configuration.

\NVAppendixText
\phantomsection\label{app:exp:encoder-capacity}%
\textbf{Encoder capacity.}\enspace We compare one \emph{Shared} encoder trained jointly on all targets with seven \emph{Regional} encoders trained on separate targets (Table~\ref{tab:encoder-capacity}). We widen the \emph{Matched} NVCode encoder to the regional parameter count. Each setup uses $1400$ frozen features and a separately fitted NVCode generator.

With similar parameter counts, regional NVCode improves Avg.~$R$ from $0.373$ to $0.423$ and lowers FC-MAE from $0.195$ to $0.135$.
\NVAppendixTable
\begingroup
\centering
\caption{\textbf{Encoder capacity by supervision.} Inter-subject scores give all eight datasets equal weight. Parameter counts exclude temporary heads and the generator.}
\label{tab:encoder-capacity}
\small
\setlength{\tabcolsep}{3pt}
\renewcommand{\arraystretch}{1.03}
\resizebox{\linewidth}{!}{%
\begin{tabular}{@{}llcccc@{}}
\toprule
\textbf{Encoder} & \textbf{Setup} & \textbf{Params (M)} & \textbf{Avg.~$R$}$\uparrow$ & \textbf{CRPS}$\downarrow$ & \textbf{FC-MAE}$\downarrow$ \\
\midrule
\multirow{3}{*}{\textbf{NVCode}} & Shared & 8.19 & 0.349 & 0.362 & 0.182 \\
 & Matched & 55.35 & \Second{0.373} & 0.317 & 0.195 \\
 & Regional & 55.38 & \Best{0.423} & 0.344 & \BestRightEdge{0.135} \\
\addlinespace[1pt]
\multirow{2}{*}{SPaRCNet} & Shared & 0.42 & 0.353 & 0.344 & \SecondRightEdge{0.163} \\
 & Regional & 1.84 & 0.342 & \Second{0.311} & 0.210 \\
\addlinespace[1pt]
\multirow{2}{*}{NeuroBOLT} & Shared & 8.19 & 0.340 & 0.410 & 0.238 \\
 & Regional & 55.38 & 0.315 & \Best{0.299} & 0.187 \\
\ResultBottomRule
\end{tabular}%
}
\par
\endgroup

\NVAppendixEnd

Widening the shared encoder raises Avg.~$R$ from $0.349$ to $0.373$, while the regional design retains the advantage seen in Table~\ref{tab:joint-flow-ablation}. The wide shared encoder attains lower marginal CRPS ($0.317$ versus $0.344$). Regional SPaRCNet and NeuroBOLT also trade lower marginal CRPS for lower Avg.~$R$ relative to their shared controls.

\myparagraph{Generative model baselines.}
The baseline families build on adversarial learning~\citep{goodfellow2014generative}, diffusion~\citep{sohl2015deep,dhariwal2021diffusion}, and score estimation~\citep{song2019generative}, with conditional diffusion also supporting image translation~\citep{saharia2022palette}.
For the cached-code comparison in Table~\ref{tab:main-generator} and its dataset breakdown, all rows use the same seven frozen NVCode encoders. We fit the projector with the NVCode velocity field, then freeze it when fitting the baseline generation networks. In this comparison, each complete generator has the form $G_j(\mathbf{e},\epsilon)=F_{\theta_j}(\Pi_{\psi^{\star}}(\mathbf{e}),\epsilon)$, with shared projector parameters $\psi^{\star}$ and separately fitted network parameters $\theta_j$. All rows use the same stored codes, z-scored labels and data splits. Each stochastic model returns $K{=}50$ samples per EEG epoch for the common evaluation metrics. The Rectified Flow control~\citep{liu2022flow} shares the Gaussian source, linear probability path, and squared velocity objective with NVCode. The two procedures compare direct fitting to observed endpoints with one round of reflow. Let
$\mathcal{D}_{\mathrm{tr}}=\{(\mathbf{c}_i,\mathbf{z}_{1,i})\}_{i=1}^{M}$,
where $\mathbf{c}_i=\mathbf{c}_{\psi}(\mathbf{x}_i)$ is the fixed EEG condition and $\mathbf{z}_{1,i}$ is the normalized joint fMRI endpoint. Define
\begin{equation}
\begin{gathered}
\gamma_t(\mathbf{a},\mathbf{b})=(1-t)\mathbf{a}+t\mathbf{b},
\\
u(\mathbf{a},\mathbf{b})=\mathbf{b}-\mathbf{a}.
\label{eq:rf-shared-path}
\end{gathered}
\end{equation}
With $\mathbf{z}_0\sim\mathcal{N}(\mathbf{0},\mathbf{I}_N)$ and $t\sim\mathcal{U}(0,1)$, the field optimum for a given projector is
\begin{equation}
\begin{gathered}
\theta^{\star}(\psi)
\in\arg\min_{\theta}
\\
\mathbb{E}
\left[
\left\|
v_{\theta}\!\left(\gamma_t(\mathbf{z}_0,\mathbf{z}_1),t,\mathbf{c}_{\psi}(\mathbf{x})\right)
-u(\mathbf{z}_0,\mathbf{z}_1)
\right\|_2^2
\right].
\label{eq:nvsi-direct-data-fit}
\end{gathered}
\end{equation}
Equation~\eqref{eq:nvsi-direct-data-fit} expresses the field optimization within Eq.~\MainPaperEqref{eq:nvsi-loss}. NVCode learns the projector and field jointly, while the reported one-reflow baseline fits its initial field with a fixed projector. With the same architecture, initialization, optimizer and budget, a single fit without reflow jointly updates both modules exactly as NVCode does.

The Rectified Flow row in Table~\ref{tab:main-generator} and Tables~\ref{tab:app:generator-all-datasets}--\ref{tab:app:generator-all-datasets:natview} denotes a specified one-reflow implementation. Its first field $v_{\theta^{(1)}}$ is fitted with Eq.~\eqref{eq:nvsi-direct-data-fit}. For each training condition $\mathbf{c}_i$, we draw and store one source $\bar{\mathbf{z}}_{0,i}$, then integrate the first field from that same source to construct its paired model-induced endpoint
\begin{equation}
\begin{gathered}
\widehat{\mathbf{z}}_{1,i}
=
\Phi_{1\leftarrow0}^{\theta^{(1)}}(\bar{\mathbf{z}}_{0,i};\mathbf{c}_i),
\\
\frac{d\mathbf{z}(t)}{dt}
=v_{\theta^{(1)}}(\mathbf{z}(t),t,\mathbf{c}_i),
\\
\mathbf{z}(0)=\bar{\mathbf{z}}_{0,i},
\label{eq:rf-generated-endpoint}
\end{gathered}
\end{equation}
where $\Phi_{1\leftarrow0}^{\theta^{(1)}}$ is the numerical flow map. We store the paired bank $\mathcal{B}_{\mathrm{RF}}=\{(\mathbf{c}_i,\bar{\mathbf{z}}_{0,i},\widehat{\mathbf{z}}_{1,i})\}_{i=1}^{M}$ without re-pairing sources and endpoints. A freshly initialized second field is fitted on straight segments from these stored pairs:
\begin{equation}
\begin{gathered}
\theta_{\mathrm{RF}}
\in\arg\min_{\theta}
\\
\mathbb{E}_{t,\mathcal{B}_{\mathrm{RF}}}
\left[
\left\|
v_{\theta}\!\left(
\gamma_t(\bar{\mathbf{z}}_0,\widehat{\mathbf{z}}_1),t,\mathbf{c}
\right)
-u(\bar{\mathbf{z}}_0,\widehat{\mathbf{z}}_1)
\right\|_2^2
\right],
\label{eq:rf-reflow-fit}
\end{gathered}
\end{equation}
Thus, NVCode directly fits observed condition--endpoint pairs once, whereas the Rectified Flow control adds a teacher-generated endpoint bank, fresh second-stage initialization, and a one-reflow objective. In all result tables, ``Rectified Flow'' denotes this one-reflow implementation.

For the NVCode--Rectified Flow comparison, both rows use the same seven frozen regional encoders, the same $1400$-dimensional concatenated representation and $512$-dimensional context, the same seven-dimensional target ordering, training-only normalization statistics, subject or temporal split, Gaussian source law, time distribution, linear interpolation, vector-valued velocity-network parameterization, optimizer settings, and validation criterion. The same total optimization budget is used: NVCode applies it to one direct fit, while Rectified Flow divides it equally between the initial fit and the one-reflow fit. Final inference is also matched at $S{=}100$ Euler steps and $K{=}50$ sources per EEG epoch, after which both methods use the same inverse normalization and evaluation code. At inference, Rectified Flow uses no held-out fMRI to construct either the test condition or the sampled trajectory.

The comparison therefore contrasts direct fitting with the specified one-reflow procedure, including its extra field initialization and model-generated endpoints, under a fixed total optimization budget.

\myparagraph{Generator-specific projectors.}
\phantomsection\label{app:exp:projector-control}
DDPM, EDM, and Score SDE use the same frozen EEG features in both settings. \emph{Frozen} fits only the generation network, using codes from the fixed NVCode projector (Table~\ref{tab:main-generator}). \emph{Own} jointly fits the network and its own $512$-dimensional projector using the generator's training objective.

Table~\ref{tab:own-projector-comparison} reports held-out-subject scores under both projector training protocols.

\myparagraph{Predictive mean Pearson correlation.}
Let $\hat Y,Y\in\mathbb{R}^{M\times N}$ stack the predictive means and observed targets for $M$ epochs in an evaluated scan or split. We average temporal Pearson correlations~\citep{pearson1895vii} over the $N$ fMRI signals:
\begin{shadedeq}
\overline{R} \;=\; \tfrac{1}{N}\sum_{n=1}^{N}
\frac{\sum_{i=1}^{M}(\hat Y_{i,n}-\bar{\hat Y}_n)(Y_{i,n}-\bar Y_n)}
     {\sqrt{\sum_{i=1}^{M}(\hat Y_{i,n}-\bar{\hat Y}_n)^2}\;\sqrt{\sum_{i=1}^{M}(Y_{i,n}-\bar Y_n)^2}},
\end{shadedeq}
where $\bar{\hat Y}_n,\bar Y_n$ are the column means. Higher is better.

\myparagraph{Mean marginal CRPS.}
The mean marginal continuous ranked probability score is computed per epoch from the $K$ Monte Carlo samples, averaged over fMRI signals and then averaged over the scan:
\begin{shadedeq}
\begin{gathered}
\overline{\mathrm{CRPS}}\;=\\
\tfrac{1}{M}\sum_{i=1}^{M}\left[\begin{aligned}
&\tfrac{1}{KN}\sum_{k=1}^{K}\|\hat{\mathbf{y}}^{(k)}_i-\mathbf{y}_i\|_1\\
&- \tfrac{1}{2K^2N}\sum_{j,k=1}^{K}\|\hat{\mathbf{y}}^{(j)}_i-\hat{\mathbf{y}}^{(k)}_i\|_1
\end{aligned}\right].
\end{gathered}
\end{shadedeq}
The bracketed term equals $\tfrac{1}{N}\sum_{n=1}^{N}\operatorname{CRPS}(\widehat F_{i,n},y_{i,n})$, where $\widehat F_{i,n}$ is the empirical distribution of $P_{i,n}=\{\hat y_{i,n}^{(k)}\}_{k=1}^{K}$. CRPS is a strictly proper scoring rule for each marginal predictive distribution~\citep{matheson1976scoring,gneiting2007strictly}. The reported average summarizes marginal probabilistic prediction quality across the seven targets. Lower is better.
Equivalently, if
$\widehat F_{i,n}=K^{-1}\sum_{k=1}^{K}\delta_{\hat y_{i,n}^{(k)}}$
is the empirical predictive distribution, then
\begin{equation}
\begin{aligned}
\operatorname{CRPS}(\widehat F_{i,n},y_{i,n})
&=\frac{1}{K}\sum_{k=1}^{K}|\hat y_{i,n}^{(k)}-y_{i,n}|\\
&\quad-\frac{1}{2K^2}\sum_{j,k=1}^{K}
|\hat y_{i,n}^{(j)}-\hat y_{i,n}^{(k)}|.
\end{aligned}
\label{eq:empirical-crps}
\end{equation}
This is the exact CRPS of the empirical predictive CDF. It follows by substituting independent draws from $\widehat F_{i,n}$ into the energy representation
$\operatorname{CRPS}(F,y)=\mathbb{E}|X-y|-\tfrac12\mathbb{E}|X-X'|$.

\myparagraph{Functional connectivity error.}
Correlation-based functional networks summarize statistical associations among brain signals~\citep{bassett2017network,deligianni2014relating,yeo2011organization}.
Let $Q_{:,n}$ denote the $n$th column of any $M\!\times\!N$ matrix and $\rho_{ab}(Q)=\mathrm{corr}(Q_{:,a},Q_{:,b})$ be the Pearson correlation~\citep{pearson1895vii} between columns $a$ and $b$. FC-MAE measures the mean absolute difference between matched off-diagonal entries in the predicted and observed $7\!\times\!7$ matrices:
\begin{shadedeq}
\mathrm{FC\text{-}MAE} \;=\; \tfrac{1}{N(N-1)}\!\sum_{a\neq b}\!\big|\rho_{ab}(\hat Y) - \rho_{ab}(Y)\big|.
\end{shadedeq}
Here $\hat Y$ contains the $K$-sample predictive mean for each EEG window. The metric matches off-diagonal entries of the predictive-mean and observed signal correlation matrices~\citep{van2010exploring,friston2011functional}. It evaluates interregional organization across EEG windows through the resulting predictive-mean trajectories. Repeated samples for each fixed EEG condition enter this metric through their mean. Lower is better.

\begin{lemma}[Affine properties of the reported metrics]
\label{lem:metric-affine-properties}
Let $\mathcal{T}(\mathbf{q})=\boldsymbol{\sigma}\odot\mathbf{q}+\boldsymbol{\mu}$ with $\sigma_n>0$ for every signal. Applying $\mathcal{T}$ to both predictions and targets leaves every per-signal Pearson correlation and the FC-MAE unchanged. For the $n$-th signal,
\begin{equation}
\operatorname{CRPS}(\mathcal{T}_n\#F,\mathcal{T}_n(y))
=
\sigma_n\,\operatorname{CRPS}(F,y).
\label{eq:crps-affine-equivariance}
\end{equation}
\end{lemma}

\begin{proof}
For any temporal vector $\mathbf{q}$,
$\mathcal{T}_n(\mathbf{q})-\overline{\mathcal{T}_n(\mathbf{q})}
=\sigma_n(\mathbf{q}-\bar{\mathbf{q}})$.
The positive scale cancels between the covariance numerator and standard-deviation denominator of Pearson correlation. The same cancellation holds independently for both columns of every functional-connectivity entry. Therefore, the complete correlation matrices and their elementwise differences are unchanged. For CRPS, both absolute differences in the energy representation acquire the common factor $\sigma_n$, which proves Eq.~\eqref{eq:crps-affine-equivariance}.
\end{proof}

Lemma~\ref{lem:metric-affine-properties} explains why inverse generator normalization preserves the temporal and cross-regional correlation diagnostics exactly. It also restores the signalwise scale on which marginal CRPS is aggregated and reported.

\myparagraph{Aggregation.}
Within each seed run, reconstruction metrics are computed per held out temporal segment in the intra-subject protocol and per held out subject split in the inter-subject protocol. If $m_{d,s}$ denotes a metric for split $s$ of dataset $d$ in that run, its dataset and benchmark summaries are
\begin{equation}
\begin{gathered}
\bar m_d=\frac{1}{|\mathcal{S}_d|}\sum_{s\in\mathcal{S}_d}m_{d,s},
\\
\bar m=\frac{1}{8}\sum_{d=1}^{8}\bar m_d.
\label{eq:macro-aggregation}
\end{gathered}
\end{equation}
This two-level macro average gives each dataset equal weight after its internal splits are summarized. Reported reconstruction means average these summaries over the three seed runs. For each dataset, method, evaluation setting, and target signal, we first aggregate correlation scores over the splits within each run. The standard deviations in Appendix~\ref{app:additional} summarize variation across three training seeds on those fixed splits. The sampled response vectors enter the predictive-mean and marginal-score calculations within each split.

\myparagraph{Sensitivity to excluding NatView.}
Reweighting the seven remaining datasets in Tables~\ref{tab:app:encoder-all-datasets}--\ref{tab:app:encoder-all-datasets:experience-sampling} and~\ref{tab:app:generator-all-datasets}--\ref{tab:app:generator-all-datasets:experience-sampling} preserved NVCode's lead in both protocols. In Intra, NVCode reached Avg.~$R=0.503$, compared with SIREN-based ($0.434$) and EDM ($0.418$), the strongest remaining encoder and generator alternatives. In Inter, it reached $0.451$, compared with SPaRCNet ($0.394$) and EDM ($0.292$). Each of the seven datasets contributes equally, using the original fitted models.

\subsection{Downstream Utility Details}
\label{app:downstream-utility}

We use subject identification to evaluate synthetic fMRI in two roles: as a test input for a classifier trained on real fMRI and as training data for a classifier tested on real fMRI.

\myparagraph{TrainReal TestSynth protocol.}
For each dataset, we train a subject identification classifier on real robust-normalized fMRI vectors from the real fMRI training temporal segments and validate it on real validation segments. The subject label space is fixed within the dataset. Classifier fitting, validation, and checkpoint selection use exclusively real fMRI data. After selection, the classifier is frozen. The real test row in Tables~\ref{tab:downstream_head} and~\ref{tab:ablation} evaluates this classifier on held out real fMRI. A synthetic row evaluates the same classifier on generated fMRI from matched held out EEG epochs. Thus, $\Delta$ measures test-input utility through the change in subject-identification accuracy. Both test conditions use the same frozen classifier fitted on the same training data.

\myparagraph{Classifier architecture and optimization.}
The classifier is a lightweight two layer network:
\begin{shadedeqstar}
\begin{gathered}
\mathrm{Linear}(7,64)\rightarrow\mathrm{LayerNorm}\rightarrow\mathrm{GELU}\\
\rightarrow\mathrm{Dropout}(0.1)\rightarrow\mathrm{Linear}(64,S_{\mathrm{subj}}),
\end{gathered}
\end{shadedeqstar}
Here $S_{\mathrm{subj}}$ is the number of subject labels in the dataset and LayerNorm follows~\citet{ba2016layer}. We use cross entropy, AdamW~\citep{kingma2014adam,loshchilov2017decoupled}, learning rate $10^{-3}$, weight decay $10^{-4}$, and batch size $256$. Early stopping monitors accuracy on the corresponding validation inputs with patience $20$ epochs.

\myparagraph{Utility score.}
For each dataset and synthetic method, we report
\begin{shadedeqstar}
\Delta=\mathrm{acc}_{\mathrm{TrainReal,TestSynth}}-\mathrm{acc}_{\mathrm{TrainReal,TestReal}}.
\end{shadedeqstar}
A higher $\Delta$ means that the frozen classifier attains higher identity accuracy on synthetic fMRI. The headline $\bar\Delta=\frac{1}{8}\sum_{d=1}^{8}\Delta_d$ weights each dataset equally. Tables report gains to three decimal places; multiplying an unrounded gain by $100$ expresses it in percentage points. This score complements reconstruction metrics by measuring classifier-accessible identity structure under the shift from real training inputs to synthetic test inputs.

\NVAppendixText
\textbf{Synthetic training.}\enspace For each dataset, we train separate classifiers on DDPM or NVCode predictions from matched training EEG and select checkpoints on synthetic validation inputs. The architecture, optimization, temporal splits, and identity labels stay fixed. All classifiers, including the real-trained reference, are tested on the same held-out real fMRI epochs.

NVCode improves accuracy on all eight datasets, compared with five for DDPM. Table~\ref{tab:tstr} reports equally weighted mean gains of $1.38$ and $0.54$ percentage points, respectively. On NatView, NVCode gains $1.9$ points while DDPM loses $1.4$ on the same real test inputs.
\NVAppendixTable
\begingroup
\centering
\caption{\textbf{Synthetic training, real fMRI tests.} Each entry is the change $\Delta_{\mathrm{TSTR}}$ from real fMRI training on the same test epochs. Mean weights all eight datasets equally.}
\label{tab:tstr}
\renewcommand\arraystretch{0.92}
\setlength{\tabcolsep}{2.8pt}
\resizebox{\linewidth}{!}{%
\small
\begin{tabular}{@{}p{.44\linewidth}*{2}{W{c}{.225\linewidth}}@{}}
\toprule
\textbf{Dataset} & \textbf{DDPM} & \textbf{NVCode} \\
\midrule
\SleepEEGfMRI & \Best{+0.014} & +0.007 \\
\MultiCLARID & +0.009 & \BestRightEdge{+0.015} \\
\GradCPT & \Best{+0.023} & +0.021 \\
NeuroBOLT & $-0.005$ & \BestRightEdge{+0.008} \\
\XPone & $-0.003$ & \BestRightEdge{+0.016} \\
\AffectiveMusic & +0.010 & \BestRightEdge{+0.014} \\
NatView & $-0.014$ & \BestRightEdge{+0.019} \\
\ExperienceSampling & +0.009 & \BestRightEdge{+0.010} \\
\midrule
\textbf{Mean} & +0.005 & \BestRightEdge{+0.014} \\
\ResultBottomRule
\end{tabular}}
\par
\endgroup

\NVAppendixEnd

With the test inputs held fixed, the gains show that a classifier can learn identity cues from synthetic NVCode outputs and use them when identifying participants from measured fMRI.

\suppressfloats[t]
\begin{table}[t]
\centering
\UseCVPRTablePalette
\captionsetup{width=.95\linewidth,justification=justified}
\caption{\textbf{Full regional aggregation provides the strongest downstream transfer.} The encoder set supplied to the projector varies. Generator capacity, sampler, training protocol, and the real-trained classifier are held constant. \RankingCaption}
\label{tab:ablation}
\renewcommand\arraystretch{1.05}
\setlength{\tabcolsep}{2.8pt}
\resizebox{.95\linewidth}{!}{%
\small
\begin{tabular}{@{}l|*{8}{c}|c@{}}
\toprule
\multirow{2}{*}{\textbf{Encoder}} & \multicolumn{8}{c|}{\textbf{Dataset accuracy / $\Delta$}} & \multirow{2}{*}{\textbf{Mean}} \\
& \SleepEEGfMRI & \MultiCLARID & \GradCPT & NeuroBOLT & \XPone & \AffectiveMusic & NatView & \ExperienceSampling & \\
\midrule
Real test & 0.861 & 0.787 & 0.746 & 0.869 & 0.874 & 0.782 & 0.747 & 0.702 & 0.796 \\
\midrule
Cuneus only & +0.004 & \Second{+0.016} & +0.018 & +0.000 & -0.011 & -0.001 & +0.007 & -0.001 & +0.004 \\
Heschl's only & -0.013 & -0.017 & +0.021 & -0.022 & +0.002 & -0.001 & -0.081 & -0.010 & -0.015 \\
Mid.\ Frontal only & +0.006 & -0.015 & -0.018 & -0.002 & -0.006 & +0.001 & -0.016 & -0.004 & -0.007 \\
Precuneus only & +0.022 & +0.008 & \Second{+0.033} & -0.003 & \Best{+0.040} & \Second{+0.021} & \Second{+0.008} & +0.014 & \SecondRightEdge{+0.018} \\
Putamen only & -0.016 & -0.015 & -0.024 & -0.019 & +0.005 & -0.024 & -0.082 & -0.036 & -0.026 \\
Thalamus only & -0.018 & -0.009 & -0.025 & -0.041 & -0.054 & -0.034 & -0.074 & +0.002 & -0.032 \\
Global Signal only & \Second{+0.024} & +0.003 & -0.006 & \Second{+0.003} & \Second{+0.036} & \Second{+0.021} & -0.001 & \Best{+0.021} & +0.013 \\
\addlinespace[1pt]
\textbf{NVCode (Ours)} & \Best{+0.028} & \Best{+0.027} & \Best{+0.036} & \Best{+0.043} & +0.019 & \Best{+0.031} & \Best{+0.045} & \Second{+0.020} & \BestRightEdge{+0.031} \\
\ResultBottomRule
\end{tabular}}
\end{table}

\subsection{Ablation Study Details}
\label{app:ablation-study}

Table~\ref{tab:ablation} compares the complete and single-encoder representation sets under TrainReal TestSynth. Main-paper Table~\ref{tab:joint-flow-ablation} shows how encoder and output designs affect reconstruction on held-out subjects.

\subsubsection{Single-signal regional-encoder ablation}
\label{app:ablation:single-signal}

All variants use the same sampler and fMRI output space. Downstream evaluation uses the same classifier and metric for every representation set.

\myparagraph{Single-signal representation variants.}
Each alternative retains the Cuneus, Heschl's, Middle Frontal, Precuneus, Putamen, Thalamus, or Global Signal representation map. The retained map enters the same type of learned projector, preserving condition dimensionality. Each representation set uses the same generator training procedure and capacity.

\myparagraph{Results and interpretation.}
Full NVCode achieves the highest mean utility lift, $\bar{\Delta}=+0.031$, compared with $+0.018$ for Precuneus only and $+0.013$ for Global Signal only. Its gains reach $+0.045$ on NatView, $+0.043$ on NeuroBOLT, and $+0.036$ on \GradCPT, with positive gains on all eight datasets. The full NVCode row in Table~\ref{tab:ablation} reproduces the downstream utility scores in main-paper Table~\ref{tab:downstream_head}.

\subsubsection{Controlled joint-output and stochastic-interpolant ablation}
\label{app:joint-flow-ablation}

\newcounter{ablationcontrol}[subsubsection]
\renewcommand{\theablationcontrol}{\thesubsubsection.\alph{ablationcontrol}}
\providecommand{\theHablationcontrol}{}
\renewcommand{\theHablationcontrol}{\theHsubsubsection.\arabic{ablationcontrol}}
\newcommand{\ablationcontrolparagraph}[2]{%
  \par\refstepcounter{ablationcontrol}\label{#1}%
  \myparagraph{\theablationcontrol\ #2}%
}

Table~\ref{tab:joint-flow-ablation} compares two EEG encoder configurations with four output designs under inter-subject evaluation. The seven controls below follow the row order in the table; the eighth row is the full NVCode reference.

\myparagraph{Shared experimental base.}
All eight configurations use the retained EEG--fMRI pairs, robust-normalized targets, subject-disjoint splits, and target ordering for the eight datasets in Table~\ref{tab:datasets}. The validation-selected EEG encoder checkpoints are reused across output designs and remain frozen. Whenever a control includes a second stage, it fits a separate two-layer projector with $d_c=512$ (Appendix~\ref{app:method:aggregator}) together with its output module. The trainable parameter count differs by at most $1\%$ between output modules designed to match capacity.

\ablationcontrolparagraph{app:ablation:joint-regression}{Deterministic Joint Regression.}
The input is the concatenated $7\times200$-dimensional representation from the seven frozen regional encoders; their temporary scalar heads are discarded. A new projector and one joint MLP that predicts all seven targets are trained together:
\begin{shadedeqstar}
\begin{gathered}
\hat{\mathbf y}=g_{\omega}\!\left(\mathbf{c}_{\psi}(\mathbf{x})\right)\in\mathbb{R}^{N},
\\
\mathcal{L}_{\mathrm{joint\text{-}reg}}
=\frac{1}{N}\left\|\hat{\mathbf y}-\mathbf y\right\|_2^2.
\end{gathered}
\end{shadedeqstar}
The MLP follows the hidden-width sequence of the NVCode velocity network without state and time inputs, with widths adjusted for parameter matching. It predicts one response vector per EEG epoch, without a velocity field or sampling.

\ablationcontrolparagraph{app:ablation:independent-interpolants}{Independent Interpolants.}
The input is the same concatenated regional representation, and the seven temporary scalar heads are again discarded. We train a new projector and seven scalar conditional velocity fields. Each target has its own independent Gaussian source and interpolant:
\begin{shadedeqstar}
\begin{gathered}
z_{0,n}\sim\mathcal{N}(0,1),
\\
z_{t,n}=(1-t)z_{0,n}+t z_{1,n},
\\
u_{t,n}=\dot z_{t,n}=z_{1,n}-z_{0,n}.
\end{gathered}
\end{shadedeqstar}
Each scalar field is trained to match $u_{t,n}$ and generates with $d s_n(t)/dt=v_{\theta_n}(s_n(t),t,\mathbf{c}_{\psi}(\mathbf{x}))$. A field receives only its own evolving coordinate together with the shared EEG condition; the seven fields therefore have no cross-coordinate state access. Their total parameter count is matched to the joint NVCode field. The control returns $K=50$ samples obtained with $S=100$ Euler steps.

\ablationcontrolparagraph{app:ablation:shared-encoder}{NVCode w/ Shared Encoder.}
The input is the $1400$-dimensional representation from one frozen encoder supervised jointly on all seven targets. Its temporary seven-output linear head is discarded. We fit a new projector and one conditional velocity field that receives the complete evolving seven-signal vector. Each EEG epoch uses $K=50$ samples with $S=100$ Euler steps per sample.

\ablationcontrolparagraph{app:ablation:regional-heads}{First-Stage Linear Heads.}
Each EEG epoch is passed through the seven validation-selected regional encoder--head pairs. The seven retained scalar heads produce one target each, and their outputs are concatenated in the fixed target order. For each EEG epoch, the retained head directly returns one point prediction without an additional projector, regressor, or velocity field.

\ablationcontrolparagraph{app:ablation:shared-independent}{Shared Indep. Interpolants.}
The input is the frozen shared encoder representation used in Sec.~\ref{app:ablation:shared-encoder}, and its temporary seven-output head is discarded. A new projector and seven scalar conditional velocity fields are trained as in Sec.~\ref{app:ablation:independent-interpolants}, so each field observes only its own evolving coordinate and the common condition. Sampling uses $S=100$ Euler steps for each of the $K=50$ draws.

\ablationcontrolparagraph{app:ablation:shared-regression}{Shared Joint Regression.}
The input is the same frozen shared encoder representation, with its temporary seven-output head discarded. A new projector and one joint seven-output MLP are trained with the loss above. This control has no velocity field or stochastic sampling and returns one point prediction per EEG epoch.

\ablationcontrolparagraph{app:ablation:shared-head}{Shared Linear Heads.}
For each EEG epoch, the retained linear head reads the $1400$ features of the validation-selected shared encoder. Its seven outputs form one point prediction for each of the seven fMRI signals.

\myparagraph{Full NVCode reference.}
Full NVCode takes the concatenated representations from the seven frozen regional encoders and discards their temporary scalar heads. It trains a new projector and one joint seven-dimensional conditional velocity field as specified in Sections~\ref{sec:method:representations}--\ref{sec:method:generation}. The field receives the complete evolving target vector. For each EEG epoch, $S=100$ Euler steps integrate the field from each of the $K=50$ independent Gaussian draws.

\myparagraph{Evaluation and aggregation.}
Stochastic variants return $K=50$ samples per held-out EEG epoch, while deterministic variants return one point prediction. Avg.~$R$ and FC-MAE are computed from the predictive mean using the definitions in Appendix~\ref{app:main-exp}. For a deterministic prediction, CRPS is evaluated as the point-mass score $\operatorname{CRPS}(\delta_{\hat y},y)=|\hat y-y|$, which places all rows on the same lower-is-better scale. We summarize the metrics for each held-out subject split and then weight the datasets equally, following Eq.~\eqref{eq:macro-aggregation}.

\myparagraph{Regional representations.}
Across the four output designs, regional encoders improve Avg.~$R$ over their shared counterparts. The gains are $0.122$ for linear heads, $0.112$ for joint regression, $0.129$ for independent interpolants, and $0.074$ for joint interpolation. FC-MAE also decreases in all four comparisons. The regional linear heads alone reach Avg.~$R=0.241$, CRPS $=0.430$, and FC-MAE $=0.213$.

\myparagraph{Output design.}
With either encoder configuration, joint interpolation yields higher Avg.~$R$ and lower FC-MAE than independent interpolation or deterministic joint regression. With the shared encoder, joint interpolation reaches $0.349$ Avg.~$R$ and $0.182$ FC-MAE, compared with $0.276$ and $0.266$ for independent interpolation. Full NVCode attains the lowest marginal CRPS ($0.344$), followed by shared joint regression ($0.357$). It also achieves the highest Avg.~$R$ and lowest FC-MAE across all eight configurations.

\subsubsection{Joint Sample Evaluation}
\label{app:ablation:joint-samples}

We score the $K=50$ seven-signal samples for each held-out EEG window on the robust-normalized fMRI scale. NVCode and Independent Interpolants use the same frozen regional encoders and test windows. The shuffled control permutes each signal's draws within the window, preserving its empirical distribution and mean.

Table~\ref{tab:joint-sample-ablation} reports the energy score, variogram score and joint coverage for the three configurations.

\myparagraph{Multivariate scores.}
For observed vector $\mathbf{y}$ and samples $\mathbf{y}^{(1:K)}$, the energy score (ES) uses Euclidean distances~\citep{gneiting2007strictly}:
\begin{shadedeqstar}
\mathrm{ES}=
\frac{1}{K}\sum_{k=1}^{K}\|\mathbf{y}^{(k)}-\mathbf{y}\|_2
-\frac{1}{2K^2}\sum_{k,\ell=1}^{K}\|\mathbf{y}^{(k)}-\mathbf{y}^{(\ell)}\|_2.
\end{shadedeqstar}
The variogram score (VS) averages the squared discrepancies in square-root absolute differences over the $21$ unordered signal pairs~\citep{scheuerer2015variogram}:
\begin{shadedeqstar}
\mathrm{VS}=\frac{1}{21}\sum_{i<j}
\left(\sqrt{|y_i-y_j|}
-\frac{1}{K}\sum_{k=1}^{K}\sqrt{|y_i^{(k)}-y_j^{(k)}|}\right)^2.
\end{shadedeqstar}

\myparagraph{Joint coverage.}
For each window, the first $31$ samples give a mean $m_i$ and sample standard deviation $s_i$ for each signal, with $s_i$ floored at $10^{-6}$. For each of the remaining $19$ draws, we compute $d(\mathbf{u})=\max_i |u_i-m_i|/s_i$. The $18$th smallest deviation sets the radius $r$. We count the observed vector as covered when $d(\mathbf{y})\le r$, i.e., all seven observed values fall within their bounds. If the observed and reference deviations are exchangeable without ties, this rule covers $18$ of $20$ possible ranks. We summarize scores within held-out subjects and weight the eight datasets equally.

\section{Method Details}
\label{app:method}

NVCode connects target-supervised EEG representation learning to joint fMRI generation through a shared neurovascular code. This appendix specifies that representation-to-generation pathway and derives its conditional velocity-matching foundation~\citep{lipman2022flow,albergo2025stochastic}. Appendix~\ref{app:ablation-study} reports how the encoder and output designs affect prediction.

\subsection{Velocity Matching Identity}
\label{app:method:nvsi-proof}

This section specializes the conditional velocity-matching identity~\citep{lipman2022flow,albergo2025stochastic} to a fixed neurovascular code map. On a probability space $(\Omega,\mathcal{F},\mathbb{P})$, let $(\mathbf{X},\mathbf{Z}_1)$ be a paired EEG input and normalized fMRI endpoint, and set $\mathbf{C}=\mathbf{c}_{\psi}(\mathbf{X})$. Draw $\mathbf{Z}_0\sim\mathcal{N}(\mathbf{0},\mathbf{I}_N)$ independently of $(\mathbf{X},\mathbf{Z}_1)$ and $T\sim\mathcal{U}(0,1)$ independently of $(\mathbf{X},\mathbf{Z}_0,\mathbf{Z}_1)$. Define
\begin{equation}
\begin{gathered}
\mathbf{Z}_T=(1-T)\mathbf{Z}_0+T\mathbf{Z}_1,
\\
\mathbf{U}=\mathbf{Z}_1-\mathbf{Z}_0.
\label{eq:proof-random-path}
\end{gathered}
\end{equation}
Assume $\mathbb{E}(\|\mathbf{Z}_0\|_2^2+\|\mathbf{Z}_1\|_2^2)<\infty$, so the conditional velocities below are square integrable.

\myparagraph{Conditional $L_2$ projection.}
Let $\mathcal{H}=\sigma(\mathbf{Z}_T,T,\mathbf{C})$ and define $\mathbf{m}=\mathbb{E}[\mathbf{U}\mid\mathcal{H}]$.

\begin{proposition}[Optimal conditional velocity]
\label{prop:optimal-velocity}
For a fixed conditioning map $\mathbf{C}=\mathbf{c}_{\psi}(\mathbf{X})$, the minimizer of the population squared velocity-matching risk over square-integrable measurable fields is
\begin{equation}
v^{\star}(\mathbf{z},t,\mathbf{c})
=\mathbb{E}\!\left[
\mathbf{Z}_1-\mathbf{Z}_0
\mid \mathbf{Z}_T=\mathbf{z},\,T=t,\,\mathbf{C}=\mathbf{c}
\right].
\label{eq:population-velocity-main}
\end{equation}
\end{proposition}

\begin{lemma}[Orthogonal risk decomposition]
\label{lem:conditional-risk-decomposition}
For every square integrable $\mathcal{H}$-measurable vector $\mathbf{V}$,
\begin{equation}
\mathbb{E}\|\mathbf{U}-\mathbf{V}\|_2^2
=
\mathbb{E}\|\mathbf{U}-\mathbf{m}\|_2^2
+
\mathbb{E}\|\mathbf{m}-\mathbf{V}\|_2^2.
\label{eq:conditional-risk-decomposition}
\end{equation}
Thus the unique minimizer equals $\mathbf{m}$ almost surely.
\end{lemma}

\begin{proof}
Expanding around $\mathbf{m}$ gives
\begin{align}
\mathbb{E}\|\mathbf{U}-\mathbf{V}\|_2^2
&=
\mathbb{E}\|\mathbf{U}-\mathbf{m}\|_2^2
+
\mathbb{E}\|\mathbf{m}-\mathbf{V}\|_2^2 \notag\\
&\quad+
2\,\mathbb{E}\!\left[(\mathbf{U}-\mathbf{m})^{\top}(\mathbf{m}-\mathbf{V})\right].
\label{eq:risk-expansion}
\end{align}
Applying the tower property to the $\mathcal{H}$-measurable $\mathbf{m}-\mathbf{V}$ gives a zero cross term:
\begin{align}
&\mathbb{E}\!\left[(\mathbf{U}-\mathbf{m})^{\top}(\mathbf{m}-\mathbf{V})\right]\notag\\
&=
\mathbb{E}\!\left[
\mathbb{E}[\mathbf{U}-\mathbf{m}\mid\mathcal{H}]^{\top}
(\mathbf{m}-\mathbf{V})
\right]=0.
\end{align}
The remaining excess-risk term is nonnegative and vanishes precisely when $\mathbf{V}=\mathbf{m}$ almost surely.
\end{proof}

With $\mathbf{V}=v(\mathbf{Z}_T,T,\mathbf{C})$, Lemma~\ref{lem:conditional-risk-decomposition} proves Proposition~\ref{prop:optimal-velocity} and yields
\begin{equation}
\mathcal{L}(v)-\mathcal{L}(v^\star)
=
\mathbb{E}\!\left[
\|v(\mathbf{Z}_T,T,\mathbf{C})-v^\star(\mathbf{Z}_T,T,\mathbf{C})\|_2^2
\right].
\label{eq:population-excess-risk}
\end{equation}
The excess risk is therefore the squared field error averaged over the conditional interpolation path.

\myparagraph{Conditional continuity equation.}
The optimal field induces the following continuity equation for the conditional density along the interpolation path.

\begin{theorem}[Conditional probability-path identity]
\label{thm:conditional-continuity}
Fix a condition $\mathbf{c}$. Suppose the conditional law of $\mathbf{Z}_t$ has a differentiable density $p_t(\mathbf{z}\mid\mathbf{c})$, and differentiation may pass through the conditional expectation. Then $v^\star$ satisfies
\begin{equation}
\partial_t p_t(\mathbf{z}\mid\mathbf{c})
+
\nabla_{\mathbf{z}}\!\cdot
\left[
p_t(\mathbf{z}\mid\mathbf{c})
v^\star(\mathbf{z},t,\mathbf{c})
\right]
=0
\label{eq:conditional-continuity}
\end{equation}
in the weak sense.
\end{theorem}

\begin{proof}
Let $\varphi\in C_c^\infty(\mathbb{R}^{N})$ be a smooth compactly supported test function. Since $\dot{\mathbf{Z}}_t=\mathbf{U}$,
\begin{align}
&\frac{d}{dt}\int \varphi(\mathbf{z})p_t(\mathbf{z}\mid\mathbf{c})\,d\mathbf{z}\notag\\
&=
\frac{d}{dt}\mathbb{E}[\varphi(\mathbf{Z}_t)\mid\mathbf{C}=\mathbf{c}] \notag\\
&=
\mathbb{E}[\nabla\varphi(\mathbf{Z}_t)^{\top}\mathbf{U}\mid\mathbf{C}=\mathbf{c}] \notag\\
&=
\mathbb{E}[\nabla\varphi(\mathbf{Z}_t)^{\top}
v^\star(\mathbf{Z}_t,t,\mathbf{c})\mid\mathbf{C}=\mathbf{c}] \notag\\
&=
\int \nabla\varphi(\mathbf{z})^{\top}
v^\star(\mathbf{z},t,\mathbf{c})
p_t(\mathbf{z}\mid\mathbf{c})\,d\mathbf{z}.
\label{eq:weak-path-identity}
\end{align}
The tower property gives the third line. Because $\varphi$ has compact support, the boundary term vanishes when integrating by parts to obtain Eq.~\eqref{eq:conditional-continuity}.
\end{proof}

\begin{figure}[!t]
\centering
\captionsetup{width=.95\linewidth,justification=justified}
\includegraphics[width=.95\linewidth]{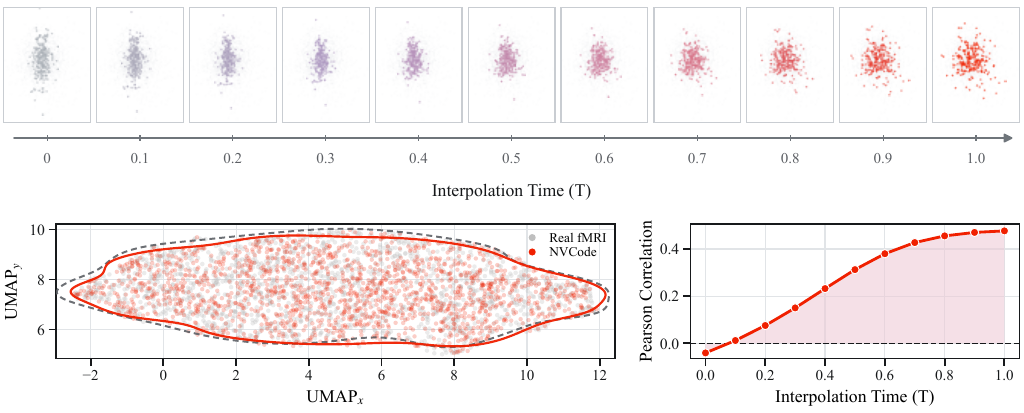}
\caption{\textbf{EEG-conditioned transport on a held-out scan.} \textbf{Top:} projected states from the Gaussian source at $t=0$ to the generated endpoint at $t=1$. \textbf{Bottom left:} real and generated endpoints overlap in UMAP space. \textbf{Bottom right:} target correlation rises along the integrated trajectory.}
\label{fig:flow-alignment}
\end{figure}

\noindent\textbf{Empirical transport diagnostic.}
Figure~\ref{fig:flow-alignment} displays eleven states along the numerically integrated trajectory of a held-out scan, from $t=0$ to $t=1$ in steps of $0.1$. Correlation rises from source to endpoint. Real and generated samples overlap in UMAP space at the endpoint.

\begin{proposition}[Conditional endpoint transport]
\label{prop:conditional-endpoint}
Assume the ODE
\begin{equation}
\begin{gathered}
\frac{d\mathbf{z}(t)}{dt}=v^\star(\mathbf{z}(t),t,\mathbf{c}),
\\
\mathbf{z}(0)\sim\mathcal{N}(\mathbf{0},\mathbf{I}_N),
\label{eq:population-probability-flow}
\end{gathered}
\end{equation}
has a unique solution map $\Phi_t^{\mathbf{c}}$, and Eq.~\eqref{eq:conditional-continuity} has a unique probability-density solution. Then
\begin{equation}
\begin{gathered}
(\Phi_t^{\mathbf{c}})_{\#}\mathcal{N}(\mathbf{0},\mathbf{I}_N)
=p_t(\cdot\mid\mathbf{c}),
\\
(\Phi_1^{\mathbf{c}})_{\#}\mathcal{N}(\mathbf{0},\mathbf{I}_N)
=\mathcal{L}(\mathbf{Z}_1\mid\mathbf{C}=\mathbf{c}).
\label{eq:conditional-pushforward}
\end{gathered}
\end{equation}
\end{proposition}

\begin{proof}
The ODE pushforward and the interpolant density from Theorem~\ref{thm:conditional-continuity} solve the same continuity equation. Independence of $\mathbf{Z}_0$ and $\mathbf{C}$ gives both paths the initial law $\mathcal{N}(\mathbf{0},\mathbf{I}_N)$. Uniqueness identifies the paths, and $\mathbf{Z}_t=\mathbf{Z}_1$ at $t=1$ gives the endpoint law.
\end{proof}

Proposition~\ref{prop:conditional-endpoint} identifies the endpoint law represented by a fixed code. If the code is conditionally sufficient, the transported endpoint law also equals the law conditional on the original EEG input:
\begin{equation}
\begin{gathered}
\mathbf{Z}_1\perp\!\!\!\perp\mathbf{X}\mid\mathbf{C},
\\
\mathcal{L}(\mathbf{Z}_1\mid\mathbf{C}=\mathbf{c}_{\psi}(\mathbf{x}))
=\mathcal{L}(\mathbf{Z}_1\mid\mathbf{X}=\mathbf{x}).
\label{eq:condition-sufficiency}
\end{gathered}
\end{equation}
These identities hold for the population-optimal field under the stated regularity and uniqueness assumptions. NVCode fits a neural approximation to this field using target-supervised codes and the velocity objective in Eq.~\MainPaperEqref{eq:nvsi-loss}.

Together, Lemma~\ref{lem:conditional-risk-decomposition}, Theorem~\ref{thm:conditional-continuity}, and Proposition~\ref{prop:conditional-endpoint} connect conditional velocity regression to transport of the joint fMRI endpoint law.

\subsection{Regional EEG Evidence Encoders}
\label{app:method:enc}

NVCode learns a target-specific EEG representation with each encoder $g_{\phi_n}$ in Eq.~\MainPaperEqref{eq:enc-loss}. Each encoder combines a spatiotemporal branch with a multi-scale spectral branch. The $N$ encoders have independent parameters. Each produces one EEG feature vector under supervision from its corresponding fMRI target.

EEG representation learning also uses masked segment reconstruction~\citep{cui2024neuro} and geometry-aware pretraining across electrode layouts~\citep{yi2023learning}. Here, the regional prediction objectives define the supervision for the temporal and spectral features specified below.

\myparagraph{Spatiotemporal branch.}
Patching provides a tokenized interface for time-series representation learning~\citep{nie2022time}.
The EEG window $\mathbf{x}\in\mathbb{R}^{C\times T}$ is divided into $P=T/w=16$ non-overlapping patches of $w=200$ samples per channel. A convolutional patcher $\mathcal{P}_{\phi_n}:\mathbb{R}^{w}\rightarrow\mathbb{R}^{d}$, channel embedding $\mathbf{a}^{\mathrm{ch}}_{c}$, and temporal embedding $\mathbf{a}^{\mathrm{time}}_{p}$ define the token at channel $c$ and patch $p$:
\begin{equation}
\begin{gathered}
\mathbf{q}^{\mathrm{ts}}_{c,p}
=
\mathcal{P}_{\phi_n}\!\left(
\mathbf{x}_{c,(p-1)w+1:pw}
\right)
+\mathbf{a}^{\mathrm{ch}}_{c}
+\mathbf{a}^{\mathrm{time}}_{p},
\\
c=1,\ldots,C,\quad p=1,\ldots,P.
\label{eq:spatiotemporal-token}
\end{gathered}
\end{equation}
The resulting sequence $\mathbf{Q}^{\mathrm{ts}}\in\mathbb{R}^{CP\times d}$ retains each patch's channel--time identity.

A learnable $[\mathrm{CLS}]$ token~\citep{devlin2019bert} is inserted to support global attention exchange during the Transformer update. The sequence is processed by a pre-norm Transformer with depth $12$, hidden dimension $d=200$, $10$ attention heads, MLP expansion ratio $4$, qkv bias enabled, stochastic depth rate $0.1$, and layer-scale initialization $0.1$. If $\mathcal{T}_{\phi_n}^{\mathrm{ts}}$ denotes this Transformer and $[\cdot]_{c,p}$ selects a contextualized EEG token, the branch output is
\begin{equation}
h^{\mathrm{ts}}_{\phi_n}(\mathbf{x})
=
\frac{1}{CP}
\sum_{c=1}^{C}\sum_{p=1}^{P}
\left[
\mathcal{T}_{\phi_n}^{\mathrm{ts}}
\bigl([\mathbf{q}_{\mathrm{CLS}};\mathbf{Q}^{\mathrm{ts}}]\bigr)
\right]_{c,p}
\in\mathbb{R}^{d}.
\label{eq:spatiotemporal-pooling}
\end{equation}
The $[\mathrm{CLS}]$ token participates in attention, while the branch output averages the $CP$ contextualized EEG tokens.

\myparagraph{Multi-scale spectral branch.}
For each channel, the spectral branch computes Short-Time Fourier Transform magnitudes at four FFT lengths:
\begin{shadedeqstar}
L_{\ell}\in\{100,200,400,800\}.
\end{shadedeqstar}
With a hop length of $L_{\ell}$, scale $\ell$ has $W_{\ell}=T/L_{\ell}$ non-overlapping windows and $F_{\ell}=L_{\ell}/2+1$ frequency bins:
\begin{shadedeqstar}
\mathbf{S}_{\ell}(\mathbf{x})\in\mathbb{R}^{C\times F_{\ell}\times W_{\ell}}.
\end{shadedeqstar}

Learned projections align every scale to the same frequency and temporal dimensions. For a channel slice $\mathbf{S}_{\ell,c}(\mathbf{x})\in\mathbb{R}^{F_{\ell}\times W_{\ell}}$, use $\mathbf{P}^{\mathrm{freq}}_{\ell}\in\mathbb{R}^{F_{\ell}\times d}$ and $\mathbf{P}^{\mathrm{time}}_{\ell}\in\mathbb{R}^{W_{\ell}\times 32}$ to obtain
\begin{shadedeqstar}
\bar{\mathbf{S}}_{\ell,c}(\mathbf{x})
=
(\mathbf{P}^{\mathrm{freq}}_{\ell})^{\top}
\mathbf{S}_{\ell,c}(\mathbf{x})
\mathbf{P}^{\mathrm{time}}_{\ell}
\in\mathbb{R}^{d\times 32}.
\end{shadedeqstar}
The four aligned tensors are summed before tokenization. For aligned temporal coordinate $u\in\{1,\ldots,32\}$, define
\begin{equation}
\mathbf{q}^{\mathrm{mss}}_{c,u}
=
\sum_{\ell=1}^{4}
\bar{\mathbf{S}}_{\ell,c}(\mathbf{x})_{:,u}
+\mathbf{b}^{\mathrm{ch}}_{c}
+\mathbf{b}^{\mathrm{time}}_{u}
\in\mathbb{R}^{d}.
\label{eq:spectral-token}
\end{equation}
A four-layer linear-attention Transformer $\mathcal{T}_{\phi_n}^{\mathrm{mss}}$ processes the stacked sequence $\mathbf{Q}^{\mathrm{mss}}\in\mathbb{R}^{32C\times d}$. We average its contextualized tokens across channels and windows:
\begin{equation}
h^{\mathrm{mss}}_{\phi_n}(\mathbf{x})
=
\frac{1}{32C}
\sum_{c=1}^{C}\sum_{u=1}^{32}
\left[
\mathcal{T}_{\phi_n}^{\mathrm{mss}}(\mathbf{Q}^{\mathrm{mss}})
\right]_{c,u}
\in\mathbb{R}^{d}.
\label{eq:spectral-pooling}
\end{equation}
Equation~\eqref{eq:spectral-token} combines the four spectral resolutions in a fixed $32C$-token sequence.

\myparagraph{Representation output.}
The final regional EEG representation combines the temporal and spectral branches:
\begin{shadedeqstar}
\mathbf{e}_n
=
g_{\phi_n}(\mathbf{x})
=
h^{\mathrm{ts}}_{\phi_n}(\mathbf{x})
+
h^{\mathrm{mss}}_{\phi_n}(\mathbf{x}).
\end{shadedeqstar}
The resulting $d=200$ feature vector carries both branches into the generator's shared conditioning module after its temporary scalar head is discarded.

\Needspace{4\baselineskip}
\subsection{Shared Conditioning Module}
\label{app:method:aggregator}

\begin{figure}[!t]
\centering
\captionsetup{width=.95\linewidth,justification=justified}
\includegraphics[width=.95\linewidth]{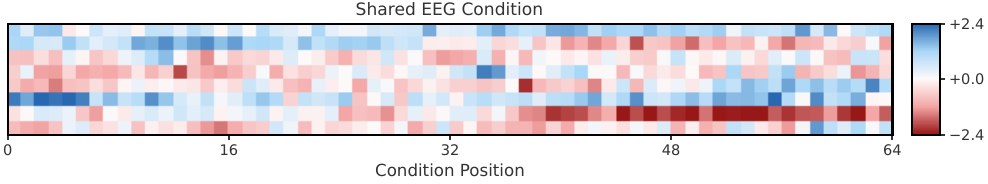}
\caption{\textbf{Shared EEG condition.} The $d_c=512$ code positions form an $8\times64$ matrix. Entries are deterministic illustrations; red, white, and blue denote negative, near-zero, and positive values, respectively.}
\label{fig:shared-condition-8x64}
\end{figure}

The shared conditioning module is the input component of the complete generator. We flatten the $N\times d$ arrangement in Eq.~\MainPaperEqref{eq:shared-conditioning} by stacking the target vectors in order, so $\mathbf{e}(\mathbf{x})\in\mathbb{R}^{Nd}$ for the matrix products below. The projector maps $Nd=7\times200=1400$ features to a $d_c=512$ neurovascular code through two layers:
\begin{shadedeqstar}
\begin{aligned}
\mathbf{h}_{\psi}(\mathbf{x})
&=
\mathrm{SiLU}\!\left(
\mathrm{LN}_1(\mathbf{W}_1\mathbf{e}(\mathbf{x})+\mathbf{b}_1)
\right),\\
\mathbf{c}_{\psi}(\mathbf{x})
&=
\mathrm{SiLU}\!\left(
\mathrm{LN}_2(\mathbf{W}_2\mathbf{h}_{\psi}(\mathbf{x})+\mathbf{b}_2)
\right).
\end{aligned}
\end{shadedeqstar}
Here $\mathbf{W}_1\in\mathbb{R}^{d_c\times Nd}$, $\mathbf{W}_2\in\mathbb{R}^{d_c\times d_c}$, and $\mathbf{b}_1,\mathbf{b}_2\in\mathbb{R}^{d_c}$. We use the default PyTorch initialization for the weights and biases of both linear layers.

The first projection makes regional encoder aggregation explicit. Partition
$\mathbf{W}_1=[\mathbf{W}_1^{(1)}\;\cdots\;\mathbf{W}_1^{(N)}]$, where
$\mathbf{W}_1^{(n)}\in\mathbb{R}^{d_c\times d}$. Then
\begin{equation}
\mathbf{W}_1\mathbf{e}(\mathbf{x})+\mathbf{b}_1
=
\sum_{n=1}^{N}\mathbf{W}_1^{(n)}\mathbf{e}_n+\mathbf{b}_1.
\label{eq:regional-condition-sum}
\end{equation}
Thus, every first-layer coordinate can aggregate target-supervised evidence from all seven encoders.

Figure~\ref{fig:shared-condition-8x64} schematically illustrates the 512-position layout. Each regional objective trains an EEG feature map; the generation objective trains how their hidden features are combined. For each EEG window, the resulting code is shared by all seven outputs throughout sampling.

Source-conditioned synthesis also appears in MRI contrast translation~\citep{dar2019image,ozbey2023unsupervised} and visual reconstruction from measured fMRI~\citep{takagi2023high,scotti2023reconstructing}. In NVCode, the source is the EEG window: $\mathbf{c}_{\psi}(\mathbf{x})$ carries its target-supervised features into the velocity network within the complete generator.

\subsection{Field Parameterization}
\label{app:method:velocity}

The velocity field combines the current seven-signal state $\mathbf{z}_t$, interpolation time $t$, and shared EEG code $\mathbf{c}_{\psi}(\mathbf{x})$. Both the state mixer and the subsequent MLP are parameterized by $\theta$:
\begin{shadedeqstar}
\begin{gathered}
\mathbf{r}_t
=
\mathbf{A}\mathbf{z}_t+\mathbf{b}_A,
\\
\mathbf{A}\in\mathbb{R}^{N\times N},
\quad
\mathbf{b}_A\in\mathbb{R}^{N}.
\end{gathered}
\end{shadedeqstar}
The state mixer exposes a learned combination of the current fMRI coordinates to each output velocity.

The time value is encoded by a sinusoidal embedding $\boldsymbol{\tau}(t)\in\mathbb{R}^{d_t}$ with $d_t=32$. For $j=0,\ldots,d_t/2-1$,
\begin{shadedeqstar}
\begin{gathered}
\tau_{2j}(t)=\sin(\omega_j t),
\\
\tau_{2j+1}(t)=\cos(\omega_j t),
\\
\omega_j=10000^{-2j/d_t}.
\end{gathered}
\end{shadedeqstar}

The velocity field is then
\begin{shadedeqstar}
v_\theta(\mathbf{z}_t,t,\mathbf{c}_{\psi}(\mathbf{x}))
=
\mathrm{MLP}_{\theta}
\!\left(
[\mathbf{c}_{\psi}(\mathbf{x});\mathbf{r}_t;\boldsymbol{\tau}(t)]
\right).
\end{shadedeqstar}
The state mixer $\mathbf{r}_t$ gives each output velocity access to the full evolving state. By contrast, each scalar field in the factorized-interpolant ablation sees its own coordinate and the common EEG code (Appendix~\ref{app:ablation-study}).

Writing $\mathbf{J}_{\mathrm{MLP},r}$ for the Jacobian of the MLP output with respect to its state-input block gives
\begin{equation}
\frac{\partial v_\theta}{\partial\mathbf{z}_t}
=
\mathbf{J}_{\mathrm{MLP},r}\mathbf{A}.
\label{eq:state-coupling-jacobian}
\end{equation}
The learned factors permit cross-coordinate responses within each joint velocity evaluation. Shared conditioning supplies the EEG evidence, while access to the evolving joint state supplies the cross-target interactions tested by the independent-interpolant control.
The MLP has widths $(d_c+N+d_t)\to h\to h/2\to h/4\to N$, with $h=512$. Its module sequence is:
\begin{enumerate}[topsep=0pt,itemsep=1pt,parsep=0pt,leftmargin=*,label=\arabic*.]
\item $\mathrm{Linear}(d_c+N+d_t,h)$, followed by LayerNorm and SiLU;
\item $\mathrm{Linear}(h,h/2)$, followed by LayerNorm and SiLU;
\item $\mathrm{Linear}(h/2,h/4)$, followed by SiLU;
\item $\mathrm{Linear}(h/4,N)$, with no activation.
\end{enumerate}
All linear layers, including the output layer, use biases and PyTorch default Kaiming uniform initialization.

\subsection{Optimization Details}
\label{app:method:optim}

NVCode first optimizes the regional representation objectives, then trains its complete generator by fitting the projector and joint field with the selected encoders fixed. The two training objectives are
\begin{align}
\min_{\phi_n,\mathbf{a}_n,b_n}\quad
&\mathcal{L}^{(n)}_{\mathrm{rep}}(\phi_n,\mathbf{a}_n,b_n),
\qquad n=1,\ldots,N,
\label{eq:stage-one-optimum}\\
\min_{\psi,\theta}\quad
&\mathcal{L}_{\mathrm{NVCode}}
\bigl(\psi,\theta;\phi_1^{\star},\ldots,\phi_N^{\star}\bigr).
\label{eq:stage-two-optimum}
\end{align}
Each regional encoder and its temporary scalar head are fitted independently with AdamW. Minimum validation MSE selects the checkpoint $\phi_n^\star$; its head is discarded and the encoder is frozen for generation.

With the selected encoders fixed, AdamW and OneCycleLR jointly optimize $\Pi_{\psi}$ and $v_{\theta}$. Training-only statistics define each endpoint $\mathbf{z}_1$, and each example receives independently sampled $\mathbf{z}_0$ and $t$. Table~\ref{tab:method-hparams} gives the split-specific epoch and stopping settings. A single validation checkpoint selects the projector and field together.

In the encoder comparison, each configuration supplies $1400$ frozen features to a separately fitted projector and velocity field of the same architecture. In the cached-code generator comparison, NVCode retains its selected projector--field pair, and each baseline fits its network using that projector's frozen outputs. Table~\ref{tab:own-projector-comparison} separately compares this setup with fitting each baseline's own projector and network jointly. Both projector settings use the same frozen EEG features, as detailed in Appendix~\ref{app:main-exp}.

\begingroup
\footnotesize
\setlength{\tabcolsep}{5pt}
\renewcommand{\arraystretch}{1.12}
\begin{table}[!t]
\centering
\captionsetup{width=.95\linewidth,justification=justified}
\caption{\textbf{NVCode hyperparameters.} Network dimensions and fitting settings are listed for the regional encoders and joint interpolant, followed by the Euler steps and the sample counts for validation and testing.}
\label{tab:method-hparams}
\begin{tabular}{@{}p{.53\linewidth}p{\dimexpr.42\linewidth-2\tabcolsep\relax}@{}}
\toprule
\textbf{Hyperparameter} & \textbf{Value} \\
\midrule
Regional representation dimension $d$ & 200 \\
Spatiotemporal patch length / patches & $w=200$ / $P=16$ \\
Representation Transformer depth / heads / MLP ratio & $12$ / $10$ / $4$ \\
Spectral branch FFT lengths & $\{100,200,400,800\}$ \\
Spectral branch temporal projection size & 32 \\
Shared context dimension $d_c$ & 512 \\
Velocity MLP hidden width $h$ / time embedding $d_t$ & 512 / 32 \\
Representation fitting optimizer & AdamW \\
Representation fitting learning rate / weight decay / batch / epochs & $5{\times}10^{-4}$ / 0.05 / 64 / 100 \\
Interpolant fitting optimizer & AdamW + OneCycleLR \\
Interpolant fitting peak learning rate / weight decay / batch & $3{\times}10^{-4}$ / $10^{-4}$ / 128 \\
Interpolant fitting max epochs / early stopping & 300 / 30 for within subject; fixed 300 for cross subject \\
Gradient norm clip / precision & 1.0 / fp32 \\
Random seeds & 42 / 2026 / 114514 \\
Euler steps $S$ / step size $\Delta t$ & 100 / $1/S$ \\
Monte Carlo samples $K$ & 50 test / 10 validation \\
\bottomrule
\end{tabular}
\end{table}
\endgroup

\subsection{Generation Details}
\label{app:method:generation}

For each evaluation EEG window, NVCode computes the shared code once and draws independent Gaussian source vectors. With $t_q=q/S$ and $\Delta t=1/S$, left-endpoint Euler integration updates all seven signals jointly:
\begin{shadedeqstar}
\begin{gathered}
\mathbf{s}_{q+1}^{(k)}
=
\mathbf{s}_{q}^{(k)}
+
\frac{1}{S}
v_{\theta^\star}
\!\left(
\mathbf{s}_{q}^{(k)},
t_q,
\mathbf{c}_{\psi^\star}(\mathbf{x}^{\star})
\right),
\\
\mathbf{s}_{0}^{(k)}\sim\mathcal{N}(\mathbf{0},\mathbf{I}_N),
\qquad q=0,\ldots,S-1.
\end{gathered}
\end{shadedeqstar}
Here $\mathbf{s}_{q}^{(k)}$ follows the learned field, while $\mathbf{z}_t$ in Eq.~\MainPaperEqref{eq:interpolation} defines its training path. We use $S=100$ steps and $K=50$ source draws per test window.

\begin{proposition}[Sampling-error decomposition]
\label{prop:euler-error}
Fix a condition $\mathbf{c}$ and write
$f_\theta(\mathbf{z},t;\mathbf{c})=v_{\theta^\star}(\mathbf{z},t,\mathbf{c})$.
Assume $f_\theta$ is $L$-Lipschitz in $\mathbf{z}$, uniformly over $t$, and the exact learned-field trajectory $\mathbf{s}_\theta(t)$ satisfies
$\|\ddot{\mathbf{s}}_\theta(t)\|_2\leq B_\theta$.
Let $\mathbf{z}^\star(t)$ be the population ODE trajectory from Eq.~\eqref{eq:population-probability-flow}, initialized from the same source sample. Then the $S$-step Euler endpoint obeys
\begin{equation}
\|\mathbf{s}_{S}-\mathbf{z}^\star(1)\|_2
\leq\frac{B_\theta}{2LS}\left(e^{L}-1\right)
+e^L\int_0^1
\bigl\|f_\theta(\mathbf{z}^\star(t),t;\mathbf{c})
-v^\star(\mathbf{z}^\star(t),t,\mathbf{c})\bigr\|_2\,dt ,
\label{eq:sampling-error-bound}
\end{equation}
with the first term interpreted as $B_\theta/(2S)$ when $L=0$.
\end{proposition}

\begin{proof}
Let $h=1/S$, $t_q=qh$, and
$e_q=\|\mathbf{s}_\theta(t_q)-\mathbf{s}_q\|_2$.
Taylor's theorem and the derivative bound give a local remainder of at most
$B_\theta h^2/2$. Lipschitz continuity then yields
\begin{equation}
\begin{gathered}
e_{q+1}\leq (1+Lh)e_q+\frac{B_\theta h^2}{2},
\\
 e_0=0.
\label{eq:euler-recursion}
\end{gathered}
\end{equation}
Summing the geometric recursion gives
\begin{equation}
e_S
\leq
\frac{B_\theta h}{2L}\left((1+Lh)^S-1\right)
\leq
\frac{B_\theta}{2LS}\left(e^L-1\right).
\label{eq:euler-global-error}
\end{equation}
Next set $d(t)=\|\mathbf{s}_\theta(t)-\mathbf{z}^\star(t)\|_2$ and
$\epsilon(t)=\|f_\theta(\mathbf{z}^\star(t),t;\mathbf{c})
-v^\star(\mathbf{z}^\star(t),t,\mathbf{c})\|_2$.
The integral forms of the two ODEs imply
\begin{equation}
d(t)\leq L\int_0^t d(s)\,ds+\int_0^t\epsilon(s)\,ds.
\end{equation}
Gr\"onwall's inequality gives
$d(1)\leq e^L\int_0^1\epsilon(t)\,dt$.
The triangle inequality combines this field-approximation term with
Eq.~\eqref{eq:euler-global-error}, proving Eq.~\eqref{eq:sampling-error-bound}.
\end{proof}

Under these assumptions, Proposition~\ref{prop:euler-error} separates the $O(S^{-1})$ Euler discretization term from the trajectory-integrated field-approximation error.

Each terminal state is mapped to the robust-normalized fMRI supervision scale by reversing the generator z-score:
\begin{shadedeqstar}
\mathbf{y}^{(k)}
=
\boldsymbol{\sigma}\odot \mathbf{s}_{S}^{(k)}+\boldsymbol{\mu}.
\end{shadedeqstar}
For one EEG window, the $K$-sample predictive mean is
\begin{shadedeqstar}
\bar{\mathbf{y}}_K=\widehat{\mathbf{y}}
=\frac{1}{K}\sum_{k=1}^{K}\mathbf{y}^{(k)}.
\end{shadedeqstar}
With independent sources, the average is unbiased for the mean of the finite-$S$ sampler. Its covariance scales inversely with the number of draws $K$:
\begin{equation}
\begin{gathered}
\mathbb{E}[\bar{\mathbf{y}}_K\mid\mathbf{x}^\star]
=
\mathbb{E}[\mathbf{y}^{(1)}\mid\mathbf{x}^\star]
=:\mathbf{m}_{\theta,S}(\mathbf{x}^\star),
\\
\operatorname{Cov}(\bar{\mathbf{y}}_K\mid\mathbf{x}^\star)
=
\frac{1}{K}\operatorname{Cov}(\mathbf{y}^{(1)}\mid\mathbf{x}^\star).
\label{eq:mc-mean-variance}
\end{gathered}
\end{equation}
Equation~\eqref{eq:mc-mean-variance} follows from linearity of expectation and independence of the source draws.

\myparagraph{Error in the predictive mean.}
For a fixed EEG window $\mathbf{x}^{\star}$ with code $\mathbf{c}$, take the population conditional mean $\mathbf{m}_{\star}(\mathbf{c})=\mathbb{E}[\boldsymbol{\sigma}\odot\mathbf{Z}_1+\boldsymbol{\mu}\mid\mathbf{C}=\mathbf{c}]$ as the reference on the supervision scale. Write $\boldsymbol{\Sigma}_{\theta,S}=\operatorname{Cov}(\mathbf{y}^{(1)}\mid\mathbf{x}^{\star})$. The conditional squared error relative to this reference is
\begin{equation}
\begin{aligned}
\mathbb{E}\!\left[\|\bar{\mathbf{y}}_K-\mathbf{m}_{\star}(\mathbf{c})\|_2^2\mid\mathbf{x}^{\star}\right]
&=\|\mathbf{m}_{\theta,S}(\mathbf{x}^{\star})-\mathbf{m}_{\star}(\mathbf{c})\|_2^2\\
&\quad+\frac{\operatorname{tr}(\boldsymbol{\Sigma}_{\theta,S})}{K}.
\end{aligned}
\label{eq:mc-conditional-mse}
\end{equation}
Expanding around $\mathbf{m}_{\theta,S}$ makes the cross term vanish by Eq.~\eqref{eq:mc-mean-variance}. At fixed $S$, increasing $K$ reduces the Monte Carlo variance without changing the finite-step sampler mean. Increasing $S$ instead reduces the Euler discretization bound in Proposition~\ref{prop:euler-error} without changing the learned field. The identity concerns the mean at one EEG window, not agreement of the full joint distributions.

For $M$ EEG windows, the sample tensor is $\mathbf{Y}^{\mathrm{samp}}\in\mathbb{R}^{K\times M\times N}$ and the stacked means form $\widehat{\mathbf{Y}}\in\mathbb{R}^{M\times N}$. Each window receives independent Gaussian sources, while its seven outputs evolve jointly. Temporal correspondence and interregional organization are evaluated across windows using $\widehat{\mathbf{Y}}$ through Pearson correlation and predictive-mean FC-MAE. Mean marginal CRPS evaluates the per-target sample distributions. These summaries separate same-window generation from the temporal structure of the predictive means.

\end{document}